%% file: main.tex
\documentclass[10pt,journal,compsoc]{IEEEtran}
\input{preamble}

\newif\ifpeerreview

\peerreviewfalse

\usepackage[nocompress]{cite}
\usepackage{url}
\usepackage{amsmath,amssymb,graphicx}

\newcommand{\jf}[1]{\textcolor{black}{#1}}

\usepackage{float}
\usepackage{xspace}
\usepackage{colortbl}
\usepackage{nicefrac}
\newcommand{\thename}{PanDORA\xspace}
\usepackage{multirow}
\usepackage{microtype}
\usepackage[percent]{overpic}
\usepackage[most]{tcolorbox}
\usepackage[export]{adjustbox}
\usepackage{cleveref}
\crefname{section}{sec.}{secs.}
\Crefname{section}{Sec.}{Secs.}
\crefname{table}{tab.}{tabs.}
\Crefname{table}{Tab.}{Tabs.}
\crefname{figure}{fig.}{figs.}
\Crefname{figure}{Fig.}{Figs.}
\crefname{equation}{eq.}{eqs.}
\Crefname{equation}{Eq.}{Eqs.}

\newcommand{\etal}{\textit{et al.}}
\usepackage{booktabs}

\usepackage[switch]{lineno}

\newcommand{\paperID}{1}

\title{\thename: Casual HDR Radiance Acquisition of Indoor Scenes for Image-based Lighting}

\author{Mohammad Reza Karimi Dastjerdi, Dominique Tanguay-Gaudreau, Frédéric Fortier-Chouinard, Yannick Hold-Geoffroy, Nima Kalantari, Jean-Fran\c{c}ois Lalonde

\texttt{\url{https://lvsn.github.io/pandora/}}
\IEEEcompsocitemizethanks{\IEEEcompsocthanksitem M. Karimi Dastjerdi, D. Tanguay-Gaudreau, F. Fortier-Chouinard, and J. Lalonde are with Université Laval. 
\IEEEcompsocthanksitem Y. Hold-Geoffroy is with Adobe Research.
\IEEEcompsocthanksitem N. Kalantari is with Texas A\&M University.}
}

\renewcommand{\footnoterule}{%
  \kern -3pt          
  \hrule width 2in    
  \kern 2.8pt         
}

\begin{document}

\IEEEtitleabstractindextext{%
\begin{abstract}
\input{sections/0_abstract}
\end{abstract}

}

\maketitle

\input{sections/1_intro}
\input{sections/2_relwork}

\input{sections/3_method}
\input{sections/4_experiments}
\input{sections/5_discussion}

\small{\noindent~\textbf{Acknowledgements} This research was supported by NSERC grants RGPIN-2020-04799 and ALLRP 557208-20, Adobe, Sentinelle Nord, and the Digital Research Alliance Canada. The authors thank Uziel Mupia Imukata for his help with data capture and Junming Chen for his early work on this project.}

\bibliographystyle{IEEEtran}
\bibliography{bibliography}

\ifpeerreview \else

\vspace*{-30pt} 
\begin{IEEEbiography}[{\includegraphics[width=1in,height=1.25in,clip,keepaspectratio]{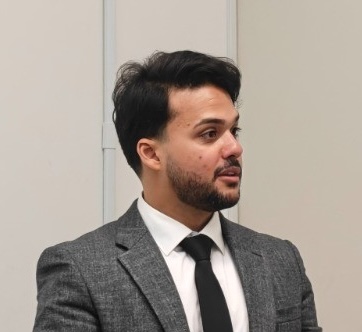}}]{Mohammad Reza Karimi Dastjerdi} received his Ph.D. from Laval University, under the supervision of Prof. Jean-François Lalonde. His work has appeared at leading venues, including ICCV, CVPR, 3DV, and BMVC, and his research contributions have been integrated into Adobe products. His current research interests include HDR imaging, neural radiance fields, generative models, and image-based lighting, with an emphasis on creative technologies and tools that empower artists.
\end{IEEEbiography}
\vfill
\vspace{-0.6in} 
\begin{IEEEbiography}[{\includegraphics[width=1in,height=1.25in,clip,keepaspectratio]{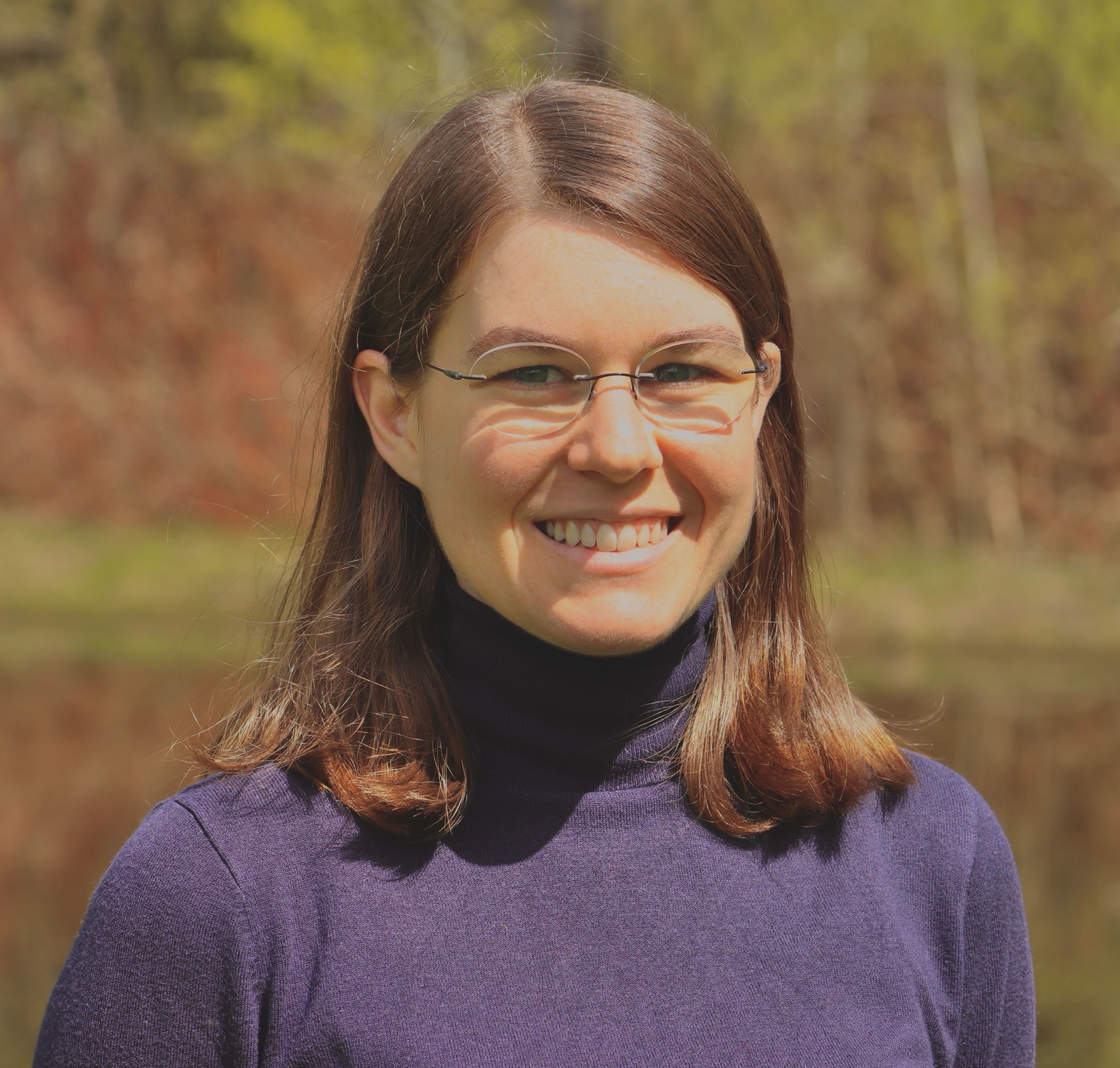}}]{Dominique Tanguay-Gaudreau} recently finished her bachelor’s in Computer Science at Laval University. She interned and later worked as a research assistant in the Computer Vision and Systems Laboratory at Laval University. She also interned with the R\&D team at Bentley Systems. She will soon start working at Bentley Systems as an associate software developer. 
\end{IEEEbiography}
\vfill
\vspace{-0.6in} 
\begin{IEEEbiography}[{\includegraphics[width=1in,height=1.25in,clip,keepaspectratio]{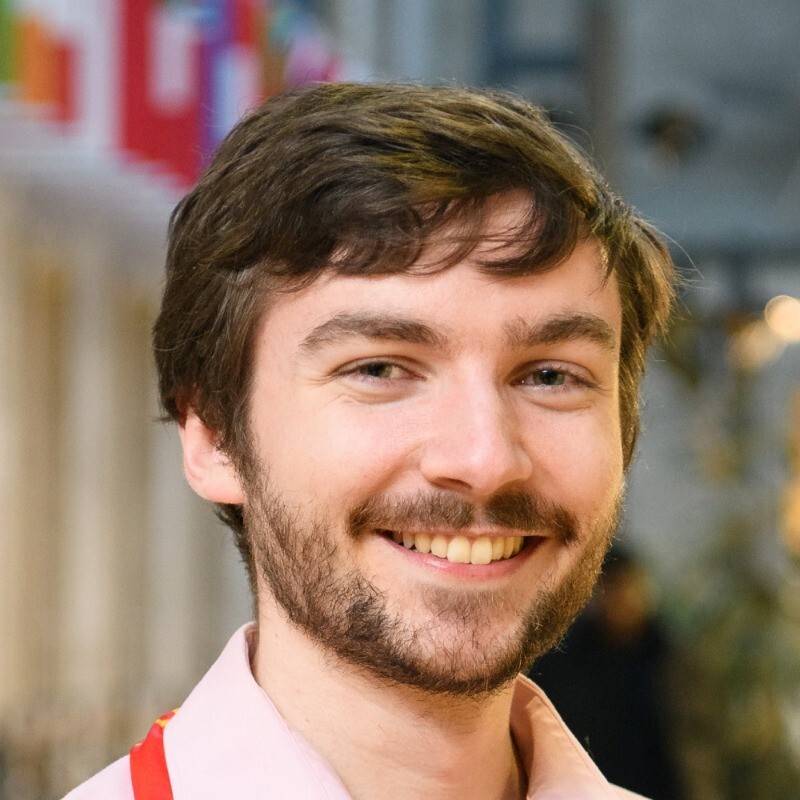}}]{Frédéric Fortier-Chouinard} is a PhD student at Laval University, working under the supervision of Prof. Jean-François Lalonde. During his PhD, he had internships at Adobe, working with Dr. Yannick Hold-Geoffroy, and at the University of California, Berkeley, working with Prof. Alexei A. Efros. His work appeared in 3DV and WACV. His research focuses on making visual generative models controllable through intuitive inputs, such as lighting, camera poses and general context. 
\end{IEEEbiography}
\vfill
\vspace{-0.6in} 
\begin{IEEEbiography}[{\includegraphics[width=1in,height=1.25in,clip,keepaspectratio]{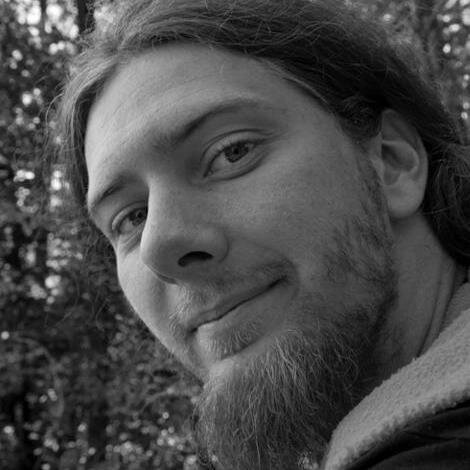}}]{Yannick Hold-Geoffroy} is a Senior Research Scientist and engineer at Adobe Research. His research interests lie at the intersection of computer vision and machine learning, with a focus on learning and using priors from natural images. He earned his Ph.D. from Laval University under the supervision of Jean-François Lalonde and Paulo Gotardo. His research interests covers lighting modeling and estimation, single-image camera and lens calibration, material and reflectance modeling, and generative image editing, appearing in venues such as CVPR, ICCV, ECCV, SIGGRAPH, and IEEE TPAMI. At Adobe, he's behind the match image technology and contributed to many generative imaging and editing technologies used by millions of users.
\end{IEEEbiography}
\vfill
\vspace{-0.6in} 
\begin{IEEEbiography}[{\includegraphics[width=1in,height=1.25in,clip,keepaspectratio]{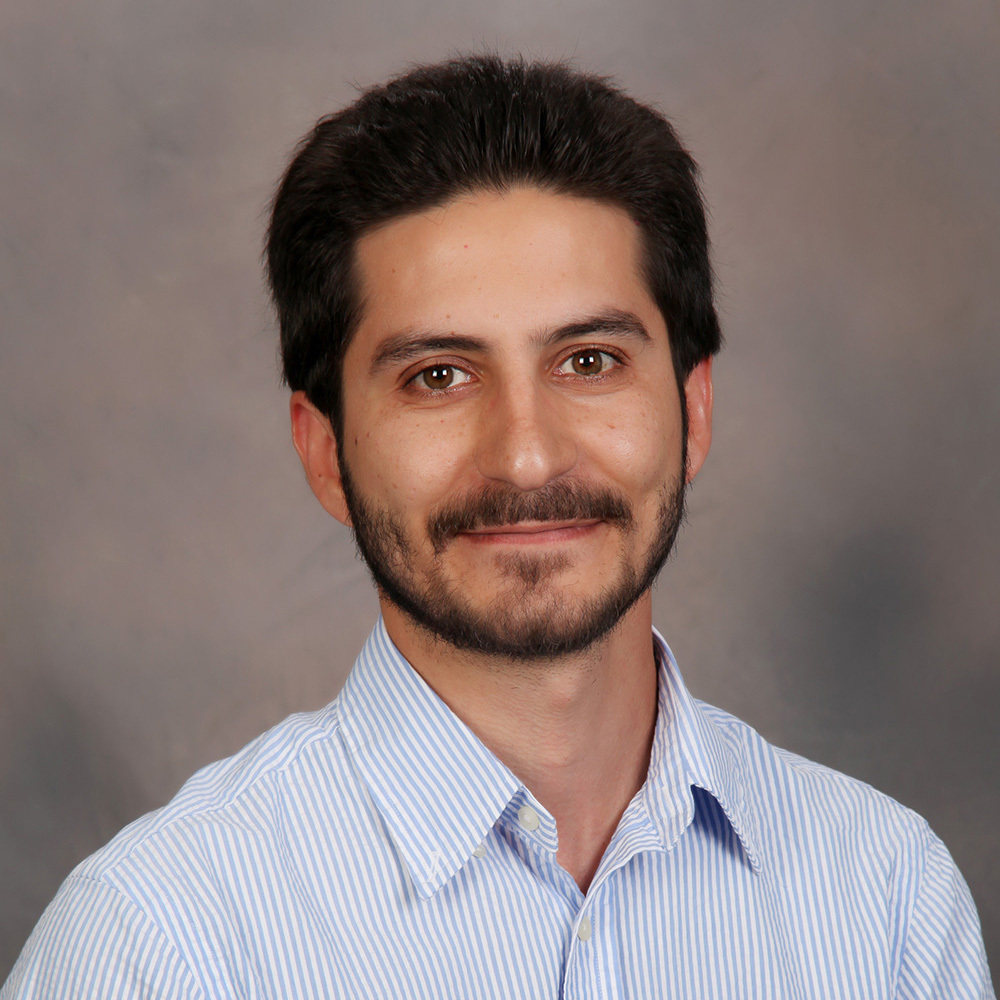}}]{Nima Kalantari} is an Associate Professor in the Department of Computer Science and Engineering at Texas A\&M University, where he is a member of the Aggie Graphics Group. Before he was a postdoctoral researcher in the Department of Computer Science and Engineering at the University of California, San Diego. He received his Ph.D. in Electrical and Computer Engineering from the University of California, Santa Barbara, in 2015. His research lies at the intersection of computer graphics, computer vision, and machine learning, with a focus on capturing and synthesizing the visual appearance of the real world. 
His work has been featured in media outlets such as Forbes, ACM TechNews, and Tech Xplore, as well as in widely viewed online videos. He is the recipient of several awards, including the NSF CAREER Award, the TEES Young Faculty Fellow Award, the Frontiers of Science Award, and a Test-of-Time Award. 
\end{IEEEbiography}
\vfill
\vspace{-0.6in} 
\begin{IEEEbiography}[{\includegraphics[width=1in,height=1.25in,clip,keepaspectratio]{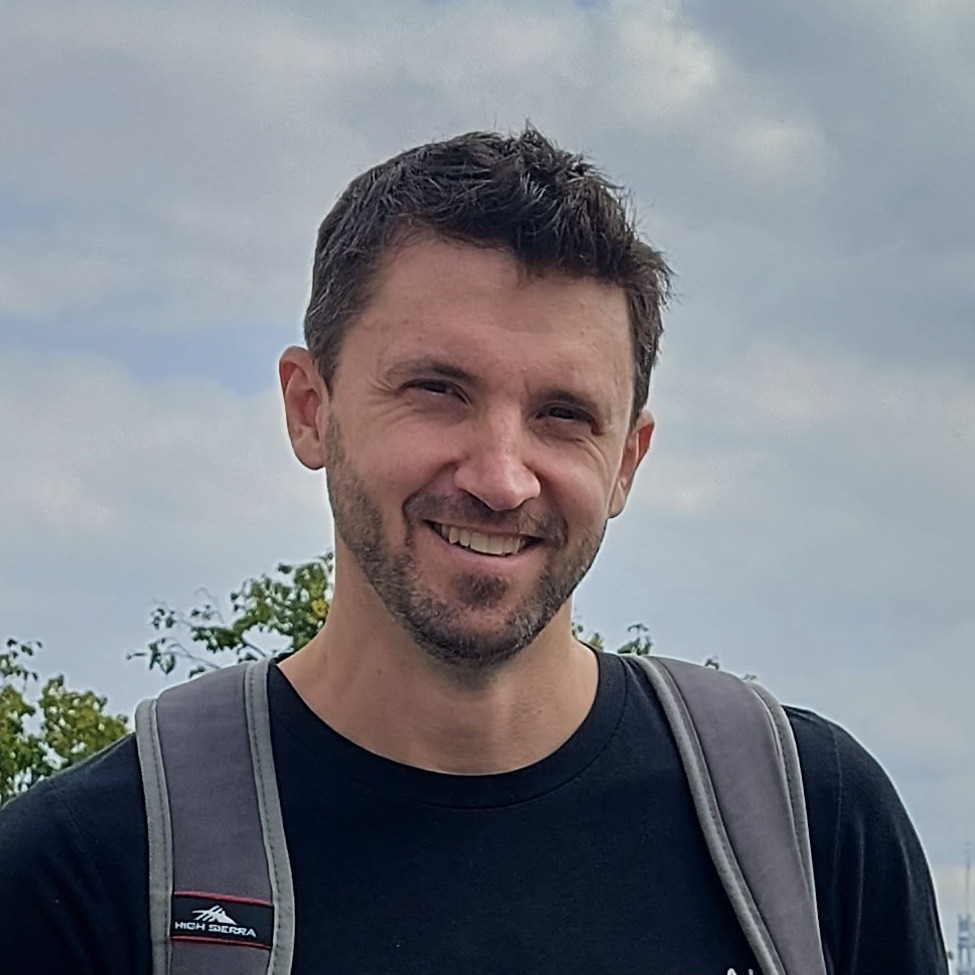}}]{Jean-Fran\c{c}ois Lalonde} is a Full Professor in the Electrical and Computer Engineering Department at Université Laval. Previously, he was a Post-Doctoral Associate at Disney Research, Pittsburgh. He received a Ph.D. in Robotics from Carnegie Mellon University in 2011. His research interests lie at the intersection of computer vision, computer graphics, and machine learning. 
He has published a large body of refereed papers in the best journals and conferences. He is actively involved in bringing research ideas to commercial products, as demonstrated by multiple patents, several technology transfers with large companies, and involvement as a scientific advisor for several high-tech startups.
\end{IEEEbiography}




\fi

\end{document}



\maketitle


\input{sec/X_suppl}

%% file: preamble.tex
\usepackage[dvipsnames]{xcolor}

\newcommand{\myparagraph}[1]{\smallskip\noindent\textbf{#1} \quad}

\newcommand{\myformat}[1]{\textsc{#1}\xspace}
\newcommand{\sceneg}{\myformat{Small office}}
\newcommand{\sceneauditoriumdark}{\myformat{Auditorium-dark}}
\newcommand{\sceneauditoriumbright}{\myformat{Auditorium}}
\newcommand{\scenebluebed}{\myformat{Blue bedroom}}

\newcommand{\sceneclasswindow}{\myformat{Classroom-windows}}
\newcommand{\sceneclassnowindow}{\myformat{Classroom-no windows}}
\newcommand{\sceneclubhouse}{\myformat{Clubhouse}}
\newcommand{\scenecoffeeroom}{\myformat{Coffee room}}

\newcommand{\scenedownstairlab}{\myformat{Basement}}
\newcommand{\scenejfoffice}{\myformat{Office}}
\newcommand{\scenelablobby}{\myformat{Lobby}}
\newcommand{\scenelvsn}{\myformat{Lab office}}

\newcommand{\scenerainbowlivingroom}{\myformat{Living room}}

\newcommand{\scenemeetingroom}{\myformat{Meeting room}}

\expandafter\def\expandafter\normalsize\expandafter{%
    \normalsize%
    \setlength\abovedisplayskip{4pt}%
    \setlength\belowdisplayskip{6pt}%
    \setlength\abovedisplayshortskip{-4pt}%
    \setlength\belowdisplayshortskip{2pt}%
}

%% file: sections/0_abstract.tex
Most novel view synthesis methods—including Neural Radiance Fields (NeRF)—struggle to capture the high dynamic range (HDR) radiance required for realistic image-based lighting (IBL). This limitation stems from a reliance on low dynamic range (LDR) imagery, which fails to capture the intensity of light sources found in indoor environments. While exposure bracketing can recover this range, it is often too slow for practical, large-scale acquisition. In this work, we introduce \thename: PANoramic Dual-Observer Radiance Acquisition, a system specifically designed for the fast and affordable capture of high-quality HDR radiance maps for IBL. Our approach utilizes two 360\textdegree{} cameras mounted on a portable monopod to simultaneously record videos at different exposures. These videos are processed by our proposed two-stage NeRF-based algorithm featuring a novel self-calibrating pipeline to estimate camera parameters. This pipeline produces non-saturated HDR radiance fields that accurately capture the radiance of a scene. When evaluated on a new dataset of real indoor environments featuring HDR ground truth lighting, PanDORA demonstrates superior fidelity in reconstructing the peak intensities necessary for downstream rendering tasks, providing a scalable and efficient solution for capturing real-world IBLs.

%% file: sections/1_intro.tex
\section{Introduction}
\label{sec:intro}
%
%
%
%
\IEEEPARstart{C}{apturing} the spatially-varying radiance of a scene is a longstanding goal with a broad range of applications, including seamlessly compositing virtual objects into real-world environments using image-based lighting (IBL). In computer graphics and vision, this goal dates back to the pioneering work of Adelson and Bergen~\cite{adelson1991plenoptic}, who defined the \emph{plenoptic function}, that is the ``reconstruction of every possible view, at every moment, from every position, at every wavelength, within the bounds of the space-time wavelength region under consideration''. Recent efforts to capture this function have made significant strides, notably with the introduction of neural radiance fields (NeRF) \cite{mildenhall2020nerf}. In this framework, the plenoptic function is implicitly modeled by a simple neural network, trained solely from 2D images (with known camera positions) to predict the color and density at each point in the 5D space (position and orientation, assuming a static scene). For IBL, this technique offers a powerful advantage over traditional environment maps: the ability to model spatially varying illumination, allowing virtual objects to react accurately to local light sources, occlusions, and complex scene geometry.

\begin{figure*}[t]
  \setlength{\hsize}{\textwidth}
  \footnotesize
  \centering
  \begin{tabular}{ccc}
  \includegraphics[height=4.8cm]{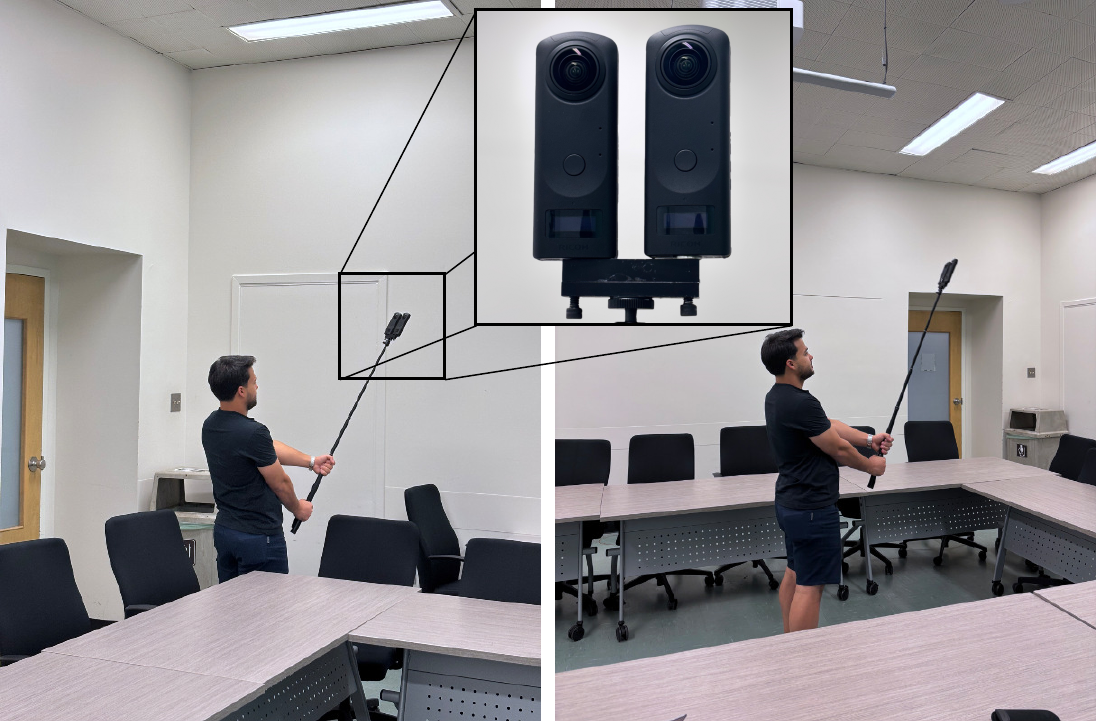} 
  & \includegraphics[height=4.8cm]{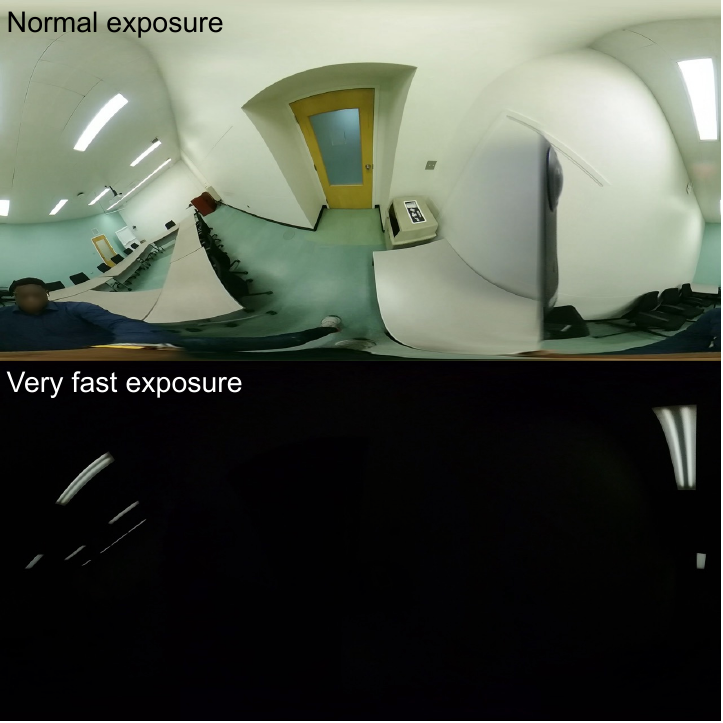}
  & \includegraphics[height=4.8cm]{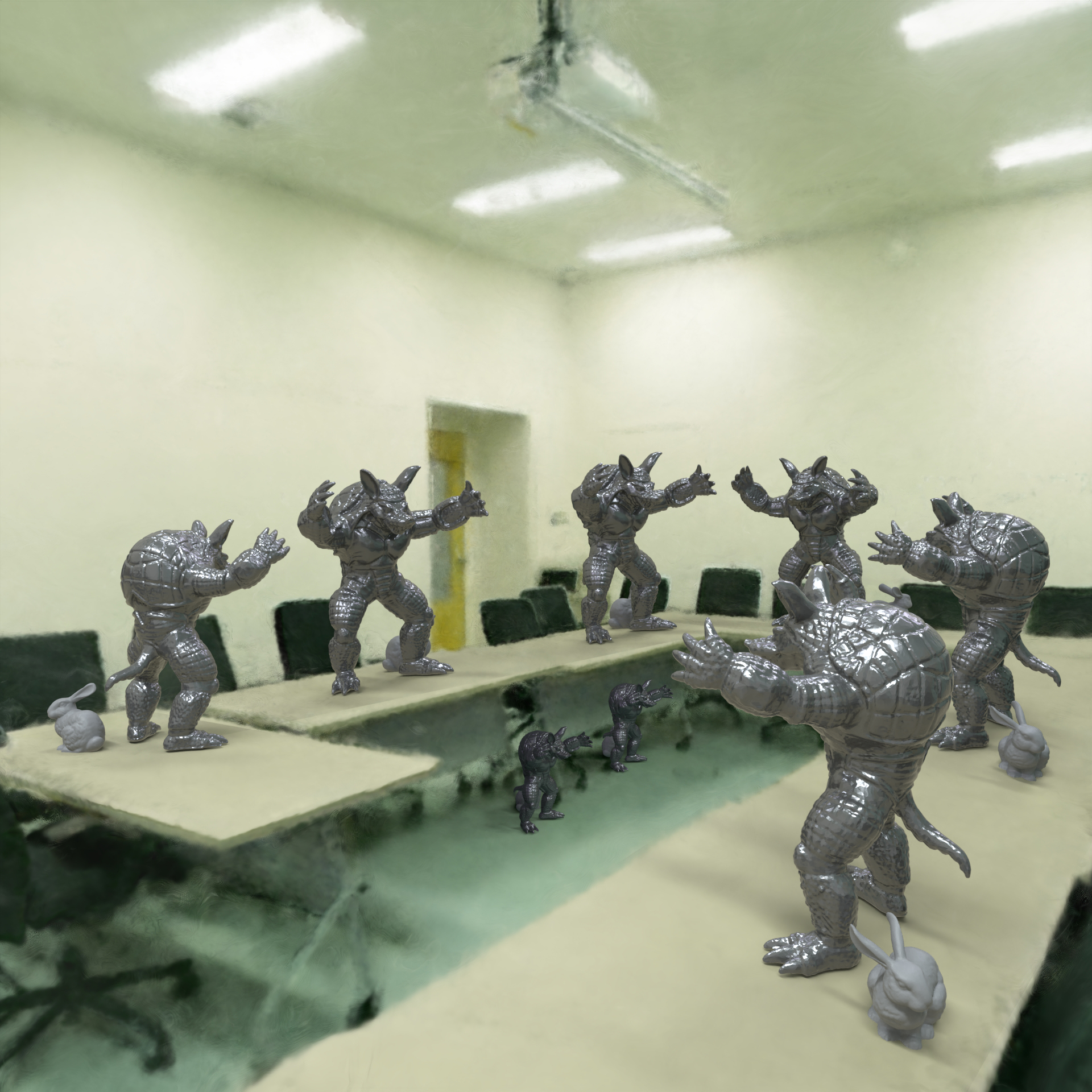} \\
  (a) A user casually scans the room with our capture device
  & (b) Frames captured at two exposures 
  & (c) Spatially-varying HDR radiance field
  \end{tabular}
  \caption{We present \thename: a novel PANoramic Dual-Observer Radiance Acquisition system for the casual capture of the HDR radiance fields in indoor scenes. Our system allows (a) a user to casually scan the scene with our capture device composed of two off-the-shelf $360^\circ$ cameras, which record (b) two videos at both well- and fast-exposure settings. After masking out the photographer, tripod, and visible camera, the resulting frames are then used to reconstruct (c) the spatially varying HDR radiance field of the scene, which can be used for image-based lighting.}
  \label{fig:teaser}
  \end{figure*}

However, for IBL to produce physically accurate shadows and specular highlights, capturing the unsaturated dynamic range of light sources is strictly required. Most existing NeRF-based techniques cannot capture the true high dynamic range (HDR) radiance of scenes since they are typically trained on photos captured with standard low dynamic range (LDR) cameras. Capturing HDR scenes generally requires multiple exposures~\cite{debevec2017recovering}, which is cumbersome for dense multi-view capture. While methods like bracketing~\cite{huang2022hdrnerf,jun2022hdr} or utilizing higher bit-depth RAW photos~\cite{mildenhall2022rawnerf} allow images to be registered by structure-from-motion (SfM) algorithms under moderate lighting, they do not scale well to the full dynamic range of complex indoor scenes. To accurately capture the unclipped intensity of a primary light source—the most critical component for IBL—the necessary exposure is so fast that the resulting image captures almost no surrounding spatial features. Consequently, these dark exposures cannot be registered by SfM algorithms, preventing the training of these methods and severely limiting their utility for realistic relighting. PanoHDR-NeRF~\cite{gera2022casual} (similarly, \cite{bolduc2025gaslight}) addresses this limitation during post-processing. The method relies on a single exposure capture and subsequently applies an inverse tonemapping neural network~\cite{yu2021lanet} specifically trained for IBL. While PanoHDR-NeRF produces acceptable rendering results with visually plausible illumination, it fundamentally lacks photometric accuracy (see \cref{sec:ablation}).

We present \thename: a PANoramic Dual-Observer Radiance Acquisition system for the casual capture of the HDR radiance field in indoor scenes, specifically designed for image-based lighting (IBL). Our proposed system, illustrated in \cref{fig:teaser}(a), comprises two $360^\circ$ cameras rigidly attached together at the end of a portable monopod. The cameras simultaneously acquire two $360^\circ$ videos: one at a regular exposure and another at a very fast exposure (typically, 1/100-th that of the regular exposure). To capture a scene, a user can simply wave the apparatus around the scene in a matter of minutes (\cref{fig:teaser}(a)). Static images are extracted from the resulting videos, pre-processed to mask the photographer and estimate the camera poses, and subsequently fed to a novel NeRF-based algorithm for HDR radiance field estimation. Specifically, we jointly learn two radiance fields, one for each exposure. This enables us to impose minimal changes to the highly-tuned NeRF approaches that work very well on LDR images. Compared to HDR baselines from previous work, our approach produces accurate lighting, preserves visual quality and retains NeRF-style ease of use.

We summarize our contributions as: 
1) a novel HDR radiance acquisition system that is affordable, easy to replicate, and allows for the casual capture of indoor environments by a novice user; 
2) a NeRF-based processing pipeline for reconstructing a HDR radiance field, enabling image-based lighting by generating HDR environment maps from arbitrary viewpoints within the captured space; 
3) a homography-based fine alignment procedure capable of matching fast- to well-exposed images; 
4) a novel dataset of 14 indoor scenes captured by our acquisition system showcasing a variety of indoor environments;
and 5) extensive experiments showing that our approach achieves state-of-the-art results over recent HDR-NeRF methods. Code and data are available on the project webpage to encourage reproducibility and further research. 

%% file: sections/2_relwork.tex
\section{Related work}
\label{sec:relwork}

\myparagraph{High dynamic range imaging} 
One significant drawback of typical cameras is their inability to capture the full dynamic range of a scene. When capturing still scenes, a common solution is bracketing---taking multiple images at different exposures and merging them into a single HDR image \cite{debevec2017recovering,robertson1999dynamic}. 
%
However, this method is not effective for dynamic scenes or cameras. Although there is research on reconstructing HDR under scene or camera motion \cite{Yan_2019_CVPR,10.1145/3072959.3073609,5559003}, these methods struggle with extreme exposure differences, preventing their use for full HDR environment map capture. Our method addresses this by aligning views using NeRF optimization and employing the standard method of \cite{debevec2017recovering}. 




%

\myparagraph{Neural radiance fields,} or NeRFs~\cite{mildenhall2020nerf}, learn an implicit volumetric representation of the scene with a simple neural network that predicts the density and view-dependent color at each 3D point. They can be trained solely from a set of 2D images, whose poses are obtained with structure-from-motion (SfM) algorithms~\cite{schoenberger2016sfm}. The original NeRF technique has been improved in many ways, notably to make its training and inference faster~\cite{fridovich2022plenoxels,muller2022instant,chen2022tensorf}, improve the visual quality with anti-aliasing~\cite{barron2021mipnerf}, handle unbounded scenes~\cite{nerf++,barron2022mipnerf360}, deal with inconsistencies through appearance embeddings~\cite{martinbrualla2021nerfwild}, and by training on omnidirectional images~\cite{gu2022omni,otonari2022non}. We leverage several such improvements by implementing our approach within Nerfstudio~\cite{nerfstudio}. 


\myparagraph{HDR radiance fields} 
Several previous works train neural radiance fields to model the high dynamic range of captured scenes. NeRF in the dark~\cite{mildenhall2022rawnerf} leverages the whole bit depth present in the RAW files of high-end cameras. HDR-NeRF~\cite{huang2022hdrnerf} proposes to train a NeRF from a series of images captured at different exposures. HDR-Plenoxels~\cite{jun2022hdr} exploits essentially the same idea in a Plenoxels~\cite{fridovich2022plenoxels} framework. These techniques, along with \cite{ruckert2022adop}, leverage differentiable tonemappers. Wu~\etal~\cite{wu2024fast} extended these methods to handle dynamic scenes. We also note that several of these ideas have been applied to the 3DGS framework~\cite{kerbl20233d}, for instance in \cite{cai2024hdr,singh2024hdrsplat,li2024chaos,liu2025gausshdr}. 
One key limitation is that exposures cannot be too slow (bright) or too fast (dark) for proper SFM registration, which restricts the dynamic range these methods can learn. In contrast, our pipeline utilizes a self-calibrating approach to robustly estimate camera poses under short exposure times. Xu~\etal~\cite{xu2023vr} developed the ``Eyeful Tower,'' a human-size rig composed of 22 Sony $\alpha$1 cameras, costing over \$USD200,000. While it produces exceptional visual quality and unsaturated HDR radiance, its daunting cost and complexity are barriers to reproducibility. 
In contrast, our method offers a good balance between accessibility and accuracy, democratizing HDR scene capture at a reasonable price (around \$USD2,000). \jf{More recently, Niemeyer~\etal~\cite{niemeyer2025nexf} proposed NExF, which uses a neural field to learn the exposure for every 3D point of the scene. While this method is more robust and performs better than HDR-NeRF~\cite{huang2022hdrnerf}, it assumes that exposure varies smoothly over the scene and that it relies on overlapping exposures, that is, regions of the scene that are correctly exposed in more than one exposure. In contrast, our method is designed to handle the challenging scenario of two fully disjoint exposures (where no scene point is properly exposed in both) and aims to capture a much higher dynamic range.}

\myparagraph{Image-based lighting} In his seminal work, Debevec~\cite{debvec1998} pioneered the use of HDR environment maps—captured via physical light probes—as distant light sources for realistic virtual object insertion. To avoid the need for specialized equipment, subsequent work estimates HDR from LDR images \cite{bolduc2025gaslight} and 360\textdegree{} panoramas \cite{hanning2021luminance,huang2026lightharmony3dharmonizingilluminationshadows,gera2022casual}. Other approaches generate omnidirectional environment maps directly from single LDR images~\cite{dastjerdi2023everlight, Phongthawee2024DiffusionLight}. Ye~\etal~\cite{Ye2025nerfrealtime} sought to model spatially-varying lighting in NeRFs, but require HDR NeRFs, which are challenging to obtain. While learning-based methods can hallucinate HDR details, they depend on network predictions that may lack photometric accuracy. In contrast, our method explicitly measures light source intensity, enabling more accurate HDR scene radiance reconstruction.


%% file: sections/3_method.tex
\section{\thename}

\begin{figure}[t!]
\centering
\includegraphics[width=0.85\linewidth]{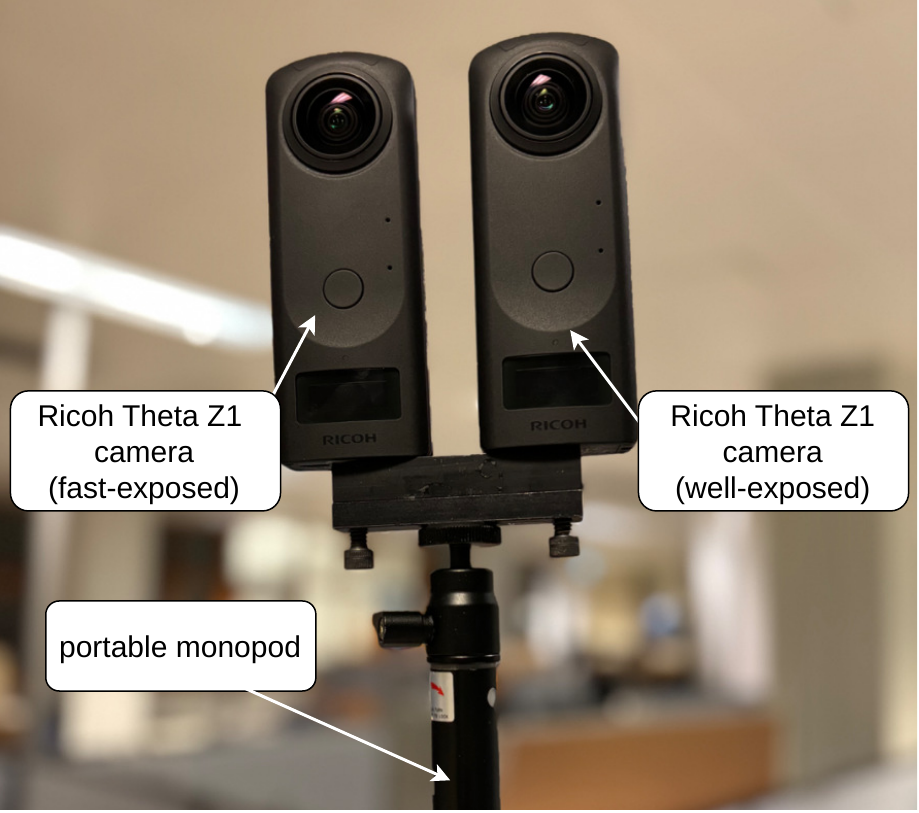}
\caption{Close-up of our proposed capture apparatus. Two Ricoh Theta Z1 panoramic cameras are rigidly attached to a portable monopod, allowing for easy manipulation. One camera captures a well-exposed video, while the other uses a much faster exposure to properly image the light sources.}
\label{fig:apparatus}
\end{figure}

%
Our capture system, illustrated in \cref{fig:teaser}(a) and \cref{fig:apparatus}, comprises two synchronized 360$^\circ{}$ cameras mounted on a portable monopod. One camera is set to record a well-exposed video while the other employs a faster exposure to accurately capture only the bright light sources. Capturing a scene can be done simply by waving the apparatus through the scene, a process that takes approximately 10 minutes and can be performed by a non-expert user. Since fusing the two captured sequences using standard HDR merging methods~\cite{debevec2017recovering} is impossible due to the baseline between the two cameras, we introduce a NeRF-based approach specifically tailored for our hardware (\cref{sec:method-architecture}). Since NeRF requires calibrated cameras and that standard structure-from-motion algorithms do not work on fast-exposed images, we propose a novel self-calibrating method (\cref{sec:camera_calibration}) which estimates the camera parameters for the fast-exposed sequence before training and subsequently refines them during training.


\subsection{NeRF architecture for HDR prediction}
\label{sec:method-architecture}
\begin{figure}[!t]
\includegraphics[width=\linewidth]{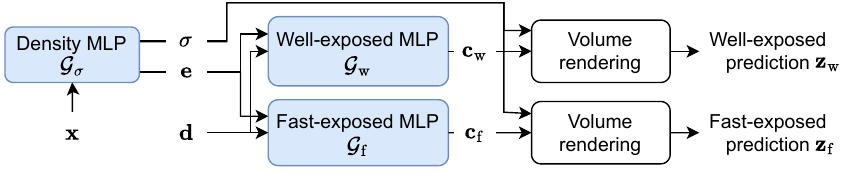}
\caption{\thename NeRF architecture, where we split the training of both well- and fast-exposed images into their own, separate MLPs. The networks are trained in two stages: in the first stage, only the well-exposed and density MLP networks are trained. In the second stage, the fast-exposed MLP is trained while the well-exposed MLP is finetuned.}
\label{fig:architecture}
\end{figure}

HDR-NeRF~\cite{huang2022hdrnerf} and HDR-Plenoxels~\cite{jun2022hdr} proposed to train a single radiance field model in HDR from multiple exposures. However, we find that significant differences in exposures lead to many ``floaters,'' which degrade reconstruction quality. To address this, we propose an alternative strategy: instead of directly training the HDR radiance field, we first learn to predict both exposures separately and then merge them to HDR. \jf{This design is also motivated by our dual-exposure capture regime, where the two exposures are disjoint in dynamic range. Methods that share a single color branch conditioned on exposure (e.g.,\ NExF \cite{niemeyer2025nexf}) rely on overlapping radiance regions to interpolate between exposures, which is not available in our setup. In contrast, our two explicit color branches learn each exposure independently, avoiding the interpolation gap and producing sharper reconstructions of both the well-exposed scene geometry and the fast-exposed light sources.} We propose the architecture illustrated in \cref{fig:architecture}, which predicts both exposures individually through their own color MLPs. Formally, the density MLP $\mathcal{G}_\sigma$ predicts the density $\sigma$ at a scene point $\mathbf{x}$. Both well- and fast-exposed MLPs ($\mathcal{G}_\mathrm{w}$ and $\mathcal{G}_\mathrm{f}$ respectively) accept as input a latent embedding $\mathbf{e}$ produced by $\mathcal{G}_\sigma$ as well as the sample view direction $\mathbf{d}$, and produce the well- and fast-exposed colors $\mathbf{c}_\mathrm{w}$ and $\mathbf{c}_\mathrm{f}$ at $\mathbf{x}$. As in the original NeRF formulation~\cite{mildenhall2020nerf}, volume rendering produces the well- and fast-exposed predictions $\mathbf{z}_\mathrm{w}$ and $\mathbf{z}_\mathrm{f}$ that can be compared to the observations to compute the NeRF photometric loss.  The networks are trained in two stages. First, the density and well-exposed MLPs $\mathcal{G}_\sigma$ and $\mathcal{G}_\mathrm{w}$ are trained on the well-exposed predictions only. Second, the fast-exposed MLP $\mathcal{G}_\mathrm{f}$ is trained, while $\mathcal{G}_\sigma$ and $\mathcal{G}_\mathrm{w}$ are fine-tuned. 
We trained our pipeline in two stages to facilitate fine alignment during the optimization process (\cref{sec:camera_calibration}).

\begin{figure}
\includegraphics[width=\linewidth]{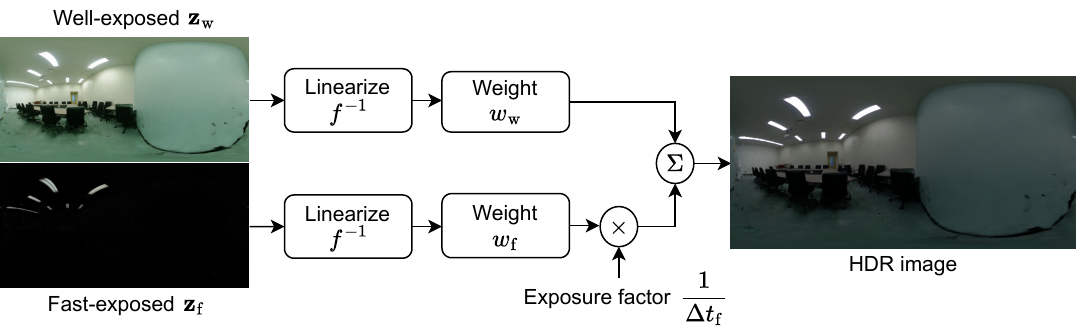}
\caption{HDR image generation by combining the well- and fast-exposed images from \thename. The predicted pixel values $\mathbf{z}_\mathrm{w}$ and $\mathbf{z}_\mathrm{f}$ are first linearized the inverse CRF $f^{-1}$, then combined by using weighting functions $w_\mathrm{w}$ and $w_\mathrm{f}$, resulting in an HDR image (right, under-exposed and tonemapped for visualization). }
\label{fig:combine}
\end{figure}

%
After training the networks, we recover HDR radiance for each ray/pixel by using a standard HDR merging method~\cite{debevec2017recovering}, as shown in \cref{fig:combine}. 
Each pixel value $\mathbf{z}$ is linearized using the inverse camera response function, obtained by fitting a $\gamma$-curve using a Macbeth color checker. Linearized images $\mathbf{p} = \mathbf{z}^\gamma$ are then weighted, re-exposed, and composed to obtain per-pixel radiance 
\begin{equation}
\mathbf{r} = \frac{w_\mathrm{w}(\mathbf{p}_\mathrm{w}) \mathbf{p}_\mathrm{w} + \frac{1}{\Delta t_\mathrm{f}} w_\mathrm{f}(\mathbf{p}_\mathrm{f}) \mathbf{p}_\mathrm{f}}{w_\mathrm{w}(\mathbf{p}_\mathrm{w}) + w_\mathrm{f}(\mathbf{p}_\mathrm{f})} \,,
\label{eq:hdr-merge}
\end{equation}
%
where $\nicefrac{1}{\Delta t_\mathrm{f}}$ is the exposure factor between the two exposures. 
The weighting functions $w_\mathrm{w}$ and $w_\mathrm{f}$ are defined as
%
\begin{equation}
w_\mathrm{w}(\mathbf{p}) \! = \! 
\begin{cases} 
1 & \text{if} \: \mathbf{p} \! < \! 0.98 \\
0 & \text{otherwise}
\end{cases} \,, 
w_\mathrm{f}(\mathbf{p}) \! = \! 
\begin{cases} 
1 & \text{if} \: \mathbf{p} \! > \! 0.1 \\
0 & \text{otherwise}
\end{cases} \,.
\end{equation}
This is effectively equivalent to the hat function in \cite{debevec2017recovering}, as a pixel is in practice never well-exposed in both exposures. 

\begin{figure}[t]
\includegraphics[width=\linewidth,trim={1.5cm 0cm 0cm 0cm},clip]{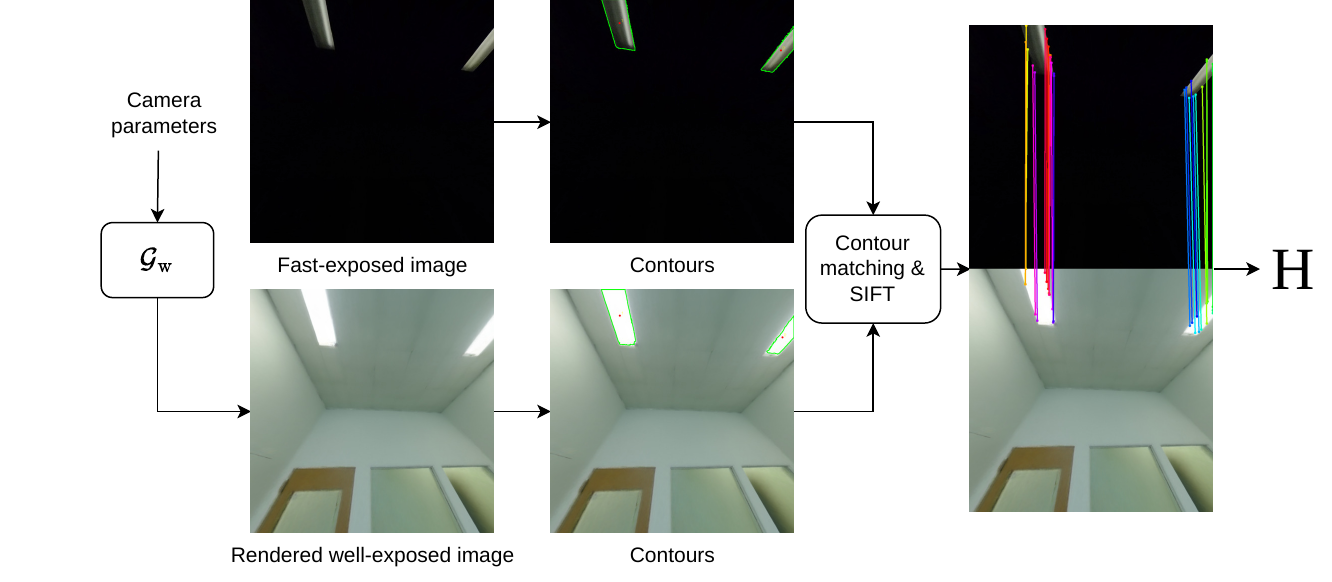}
\caption{\jf{Overview of the fine alignment algorithm. From left to right: given a fast-exposed image, we render its well-exposed counterpart at the same pose using the trained $\mathcal{G}_\text{w}$. Both images are then thresholded to extract light masks and their contours. Contours are matched across exposures via a weighted combination of Hu-moment similarity and centroid distance, after which SIFT descriptors extracted around the matched contours are used to establish correspondences for estimating a homography with RANSAC. The resulting warp aligns the fast-exposed image to its well-exposed prediction, which is then used to train $\mathcal{G}_\text{f}$.} }
\label{fig:fine_align_pipeline}
\end{figure}

\begin{figure}[t]
\centering
\scriptsize
\setlength{\tabcolsep}{1pt}
\newlength{\alignwidth}
\setlength{\alignwidth}{0.33\linewidth}
\begin{tabular}{ccc}
GT & With fine alignment & Without \\
\includegraphics[width=\alignwidth]{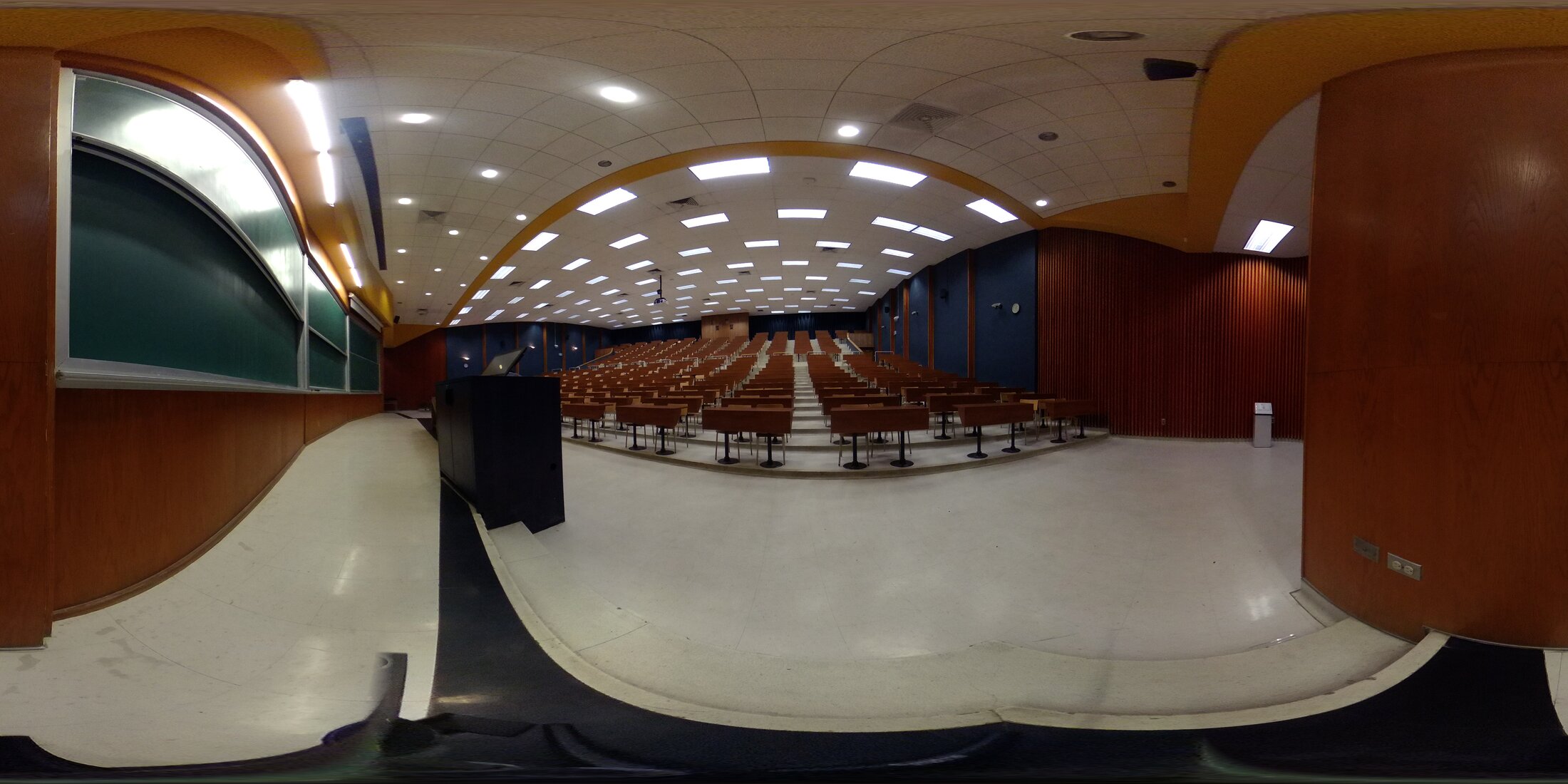} & 
\includegraphics[width=\alignwidth]{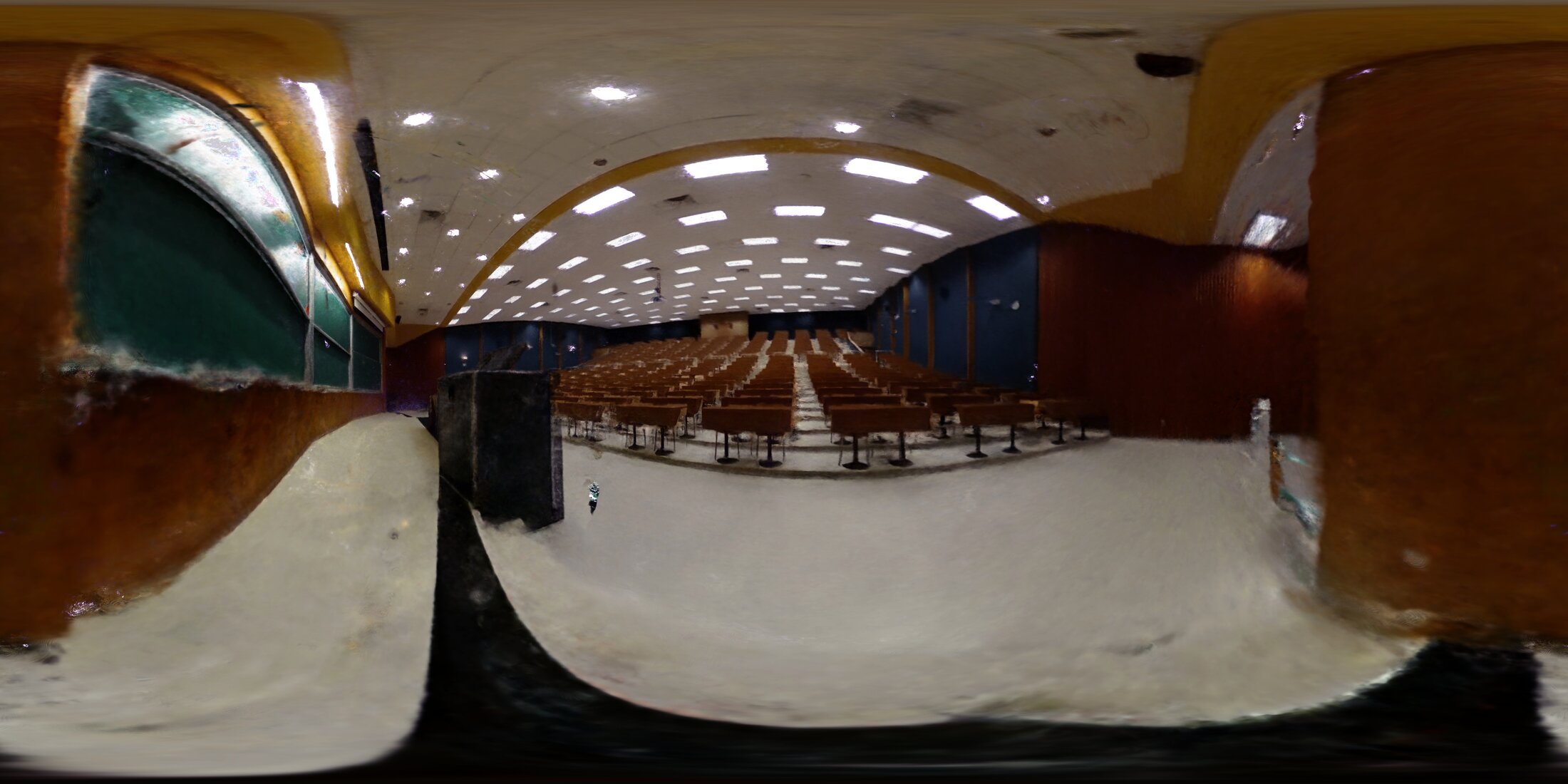} &
\includegraphics[width=\alignwidth]{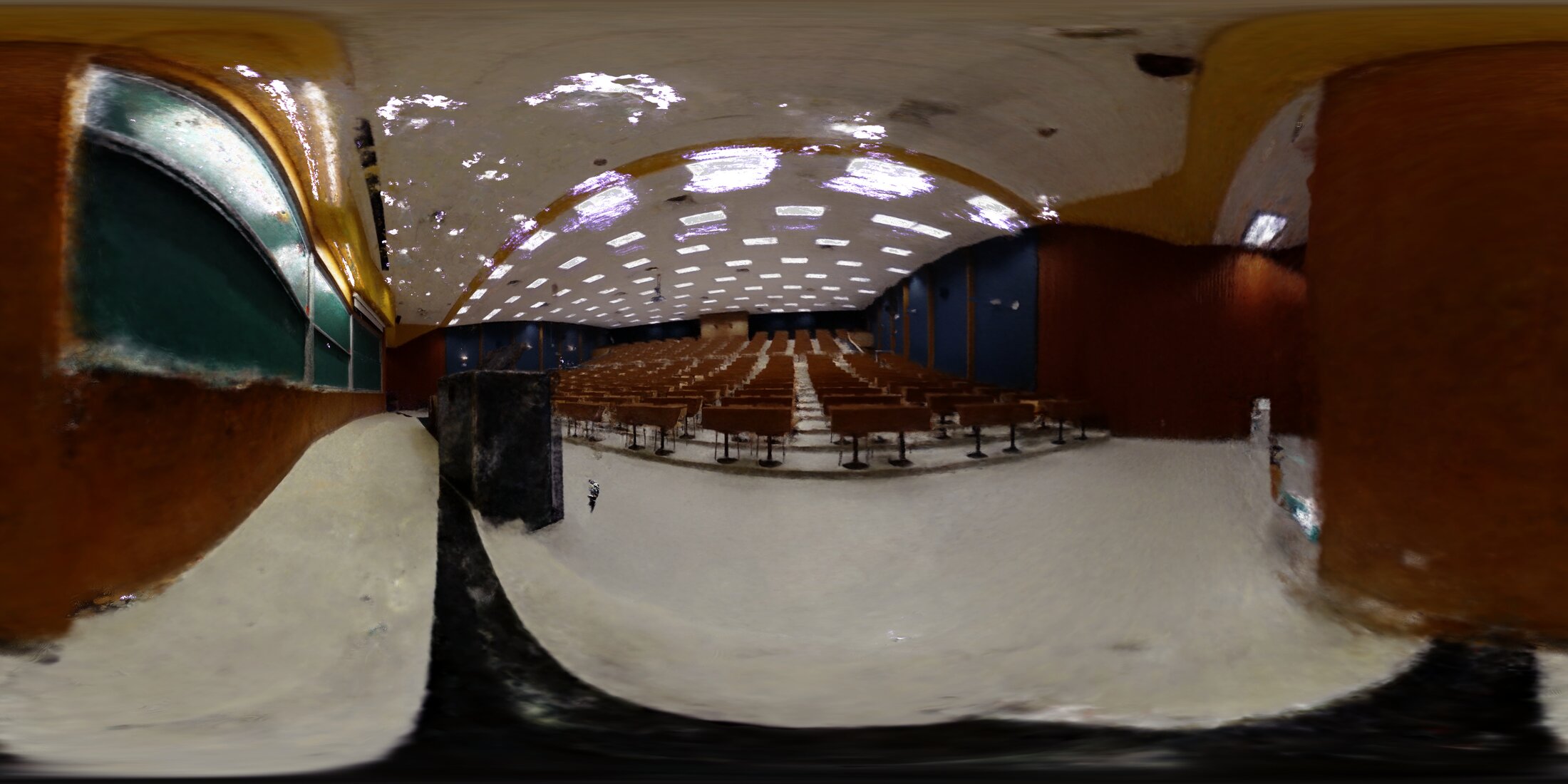} \\
\includegraphics[width=\alignwidth]{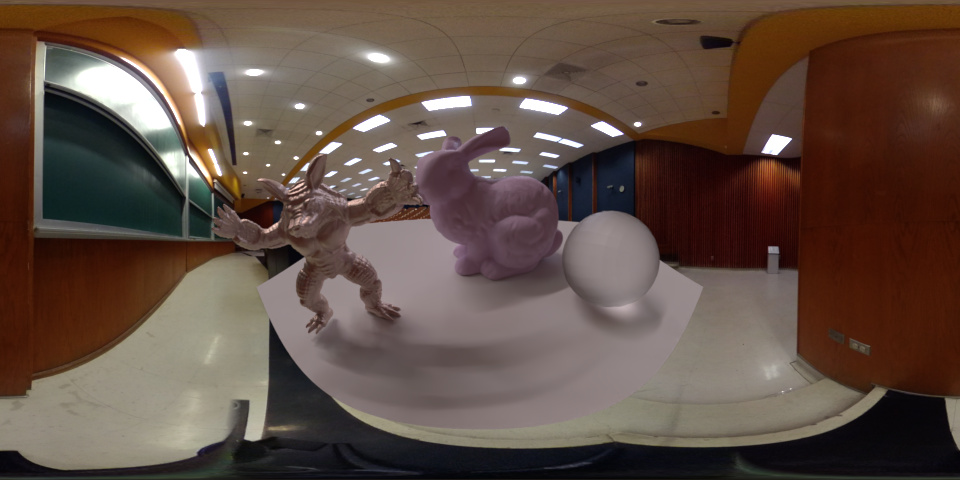} & 
\includegraphics[width=\alignwidth]{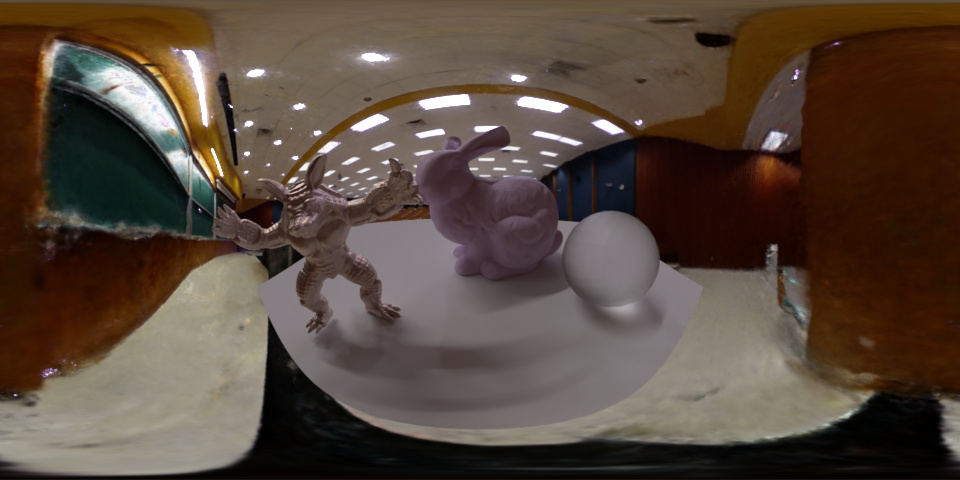} &
\includegraphics[width=\alignwidth]{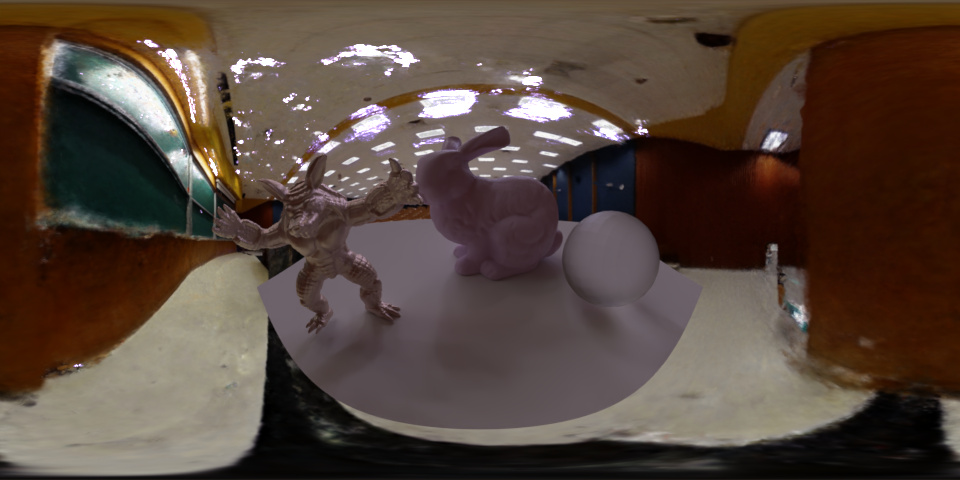} \\
\end{tabular}
\caption{Importance of the fine alignment. Relying solely on the camera pose estimation may cause slight misalignments between the well- and fast-exposed frames (third column), resulting in visual artifacts in the NeRF reconstruction. Our proposed fine alignment during training corrects for this and reduces the visual artifacts.}
\label{fig:alignment}
\end{figure}

\subsection{Self-calibrating pose estimation}
\label{sec:camera_calibration}

\myparagraph{Pre-training} To obtain each camera pose, we employ a self-calibration procedure using structure-from-motion (SfM), a common approach in NeRF-based pipelines. The fast-exposed frames are too dark (see \cref{fig:teaser}(b)) to perform registration with existing software like COLMAP~\cite{schoenberger2016sfm}. To address this, we capture two image sequences $\mathcal{S}^k = \{(I^l_i, I^r_i) \}_{i=1}^{N_k}$ (where $(l,r)$ denote the left and right cameras respectively): first, $\mathcal{S}^1$ with both cameras set to well-exposed settings, and second $\mathcal{S}^2$ with their appropriate (well- and fast-) exposures. Then, we perform SfM on $\mathcal{S}^1$ and the well-exposed frames from $\mathcal{S}^2$ to register the camera poses. We rely on the first sequence $\mathcal{S}^1$ to estimate the relative displacement between the two cameras within the SfM reconstruction coordinate system as 
\begin{align}
\Delta \mathbf{t} &= \frac{1}{N^1} \sum_{i=1}^{N^1} \left(\mathbf{t}^{1,l}_i - \mathbf{t}^{1,r}_i \right) \,. 
\end{align}
Quaternion averaging~\cite{markley2007averaging} is applied to the rotations $(\mathbf{R}^{1,l}, \mathbf{R}^{1,r})$ to obtain $\Delta \mathbf{R}$. The fast camera pose is then obtained by applying the estimated relative pose $(\Delta \mathbf{R}, \Delta \mathbf{t})$ to the well-exposed frames in the second sequence $\mathcal{S}^2$. While calibrating extrinsics with a checkerboard provides a translation $\Delta \mathbf{t}$ in metric scale, it cannot be used with SfM due to the scale ambiguity of the latter. Our method eliminates the need for this unknown scale factor.

\myparagraph{Fine alignment during training} Unfortunately, the estimated poses between the well- and fast-exposed frames are still too coarse to ensure high-quality results. Indeed, even slight misalignments can cause significant visual artifacts in the final reconstruction (see \cref{fig:alignment}, third column). We thus propose a fine alignment step for pixel-accurate alignment between the fast- and well-exposed images, performed between the two stages of NeRF training (cf.\ \cref{sec:method-architecture}). 

\jf{An overview of our proposed algorithm is shown in \cref{fig:fine_align_pipeline}.} For each fast-exposed frame, we render the well-exposed prediction at the same camera pose using the trained well-exposed MLP $\mathcal{G}_\text{w}$. Given the small camera baseline of $\sim$5cm relative to the scene distance (2--3m), \jf{which is mostly accounted for by the self-calibration procedure described above}, we assume rotation error is the dominant factor and use a homography to warp the fast-exposed image to match the well-exposed counterpart. We estimate the homography between fast- and well-exposed images using RANSAC applied to matched SIFT keypoints. However, the direct application of this method to RGB images yields poor alignment results due to the significant brightness differences between the two exposures. To address this, we compute the homography on binary images containing only the detected light masks. To obtain the light masks, we threshold the fast-exposed image at a fixed value of $\nicefrac{5}{255}$. For the well-exposed image, we apply five thresholds: $\nicefrac{230}{255}, \nicefrac{235}{255}, \nicefrac{240}{255}, \nicefrac{245}{255}$, and $\nicefrac{250}{255}$, resulting in a total of six binary masks. From these, we extract contours $K$ using the algorithm of Suzuki and Abe \cite{suzuki1985topological}.
We then compute a similarity score $\mathcal{L}_{K_{ij}}$ between each contour $K_i$ from the well-exposed masks and each contour $K_j$ from the fast-exposed mask. The similarity is defined as a weighted sum of two terms: 
\begin{equation}
\mathcal{L}_{K_{ij}} =  \lambda_M \ell_1\left(I_i, \; I_j\right) + \lambda_d \ell_2\left(c_i, c_j\right) \ ,
\end{equation}
where $c_i$ is the centroid of contour $K_i$, and $I_i = \log \left(h(K_i)\right)$ is the 7-dimensional vector of the $\log$-transformed Hu moments invariants $h(K_i)$. In our experiments, we set $\lambda_M \! = \! 0.7$ and $\lambda_d \! = \! 0.3$. 
For each contour of the fast-exposed image $K_j$, we identify the matching contour $K_i$ in the well-exposed image by selecting the one that minimizes $\mathcal{L}_{K_{ij}}$. These pairs are treated as corresponding light sources. 
This contour-based filtering reduces spurious detections from under-saturated regions in fast-exposed frames. 
Once the contours are matched, we extract SIFT descriptors from the masks surrounding each contour and match them between the well- and fast-exposed images using a brute-force nearest-neighbor algorithm. 
These SIFT correspondences are used to estimate the homography via RANSAC, which is applied to the fast-exposed image to minimize visual misalignment artifacts (see \cref{fig:alignment}, middle column). This aligned image is then used to train the fast-exposed MLP $\mathcal{G}_\text{f}$.

\subsection{Implementation details}
\label{sec:implementation-details}

The \thename apparatus (see \cref{fig:apparatus}) employs two Ricoh Theta Z1 cameras screwed on the monopod, for a total weight of around one kilogram.
Each camera captures 360\textdegree{} panoramas at a $3840 \times 1920$ resolution, and are remotely operated via WiFi. Their exposure is kept constant throughout capture. We use the auto-exposure setting to find the well-exposed setting, and configure the fast exposure to be as short as the camera allows (1/25000s), in order to capture bright light sources without saturation. The FFmpeg library is used to temporally align the videos from both cameras using audio synchronization (\jf{using a clap at the beginning and another at the end to validate that the synchronization did not drift}). After temporal alignment, static frames are extracted at 15 fps. Since we shoot 360\textdegree{} imagery, the photographer, the monopod, and the other camera are visible in the images. The photographer is masked automatically using Segment Anything~\cite{kirillov2023segany} with the prompt ``person''. Since the monopod and the other camera remain static across all frames, we manually identify them in one frame and use this mask for all frames in all videos. The masked-out pixels are ignored for camera pose estimation and during NeRF training.

\begin{figure}
    \setlength{\tabcolsep}{1pt}
    \centering
    \footnotesize
    \begin{tabular}{cc}
        \includegraphics[width=0.45\linewidth]{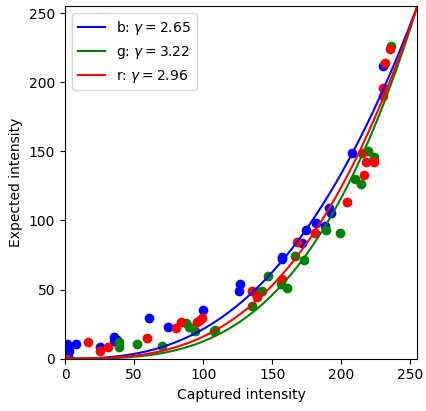} &
        \includegraphics[width=0.45\linewidth]{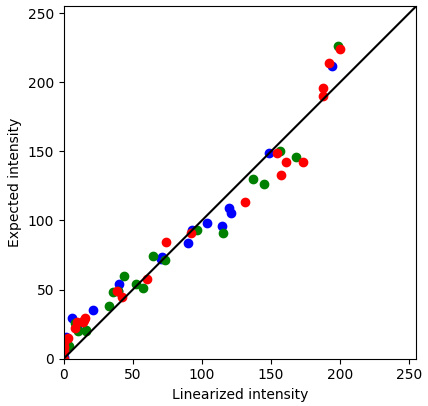} \\
        \includegraphics[width=0.45\linewidth]{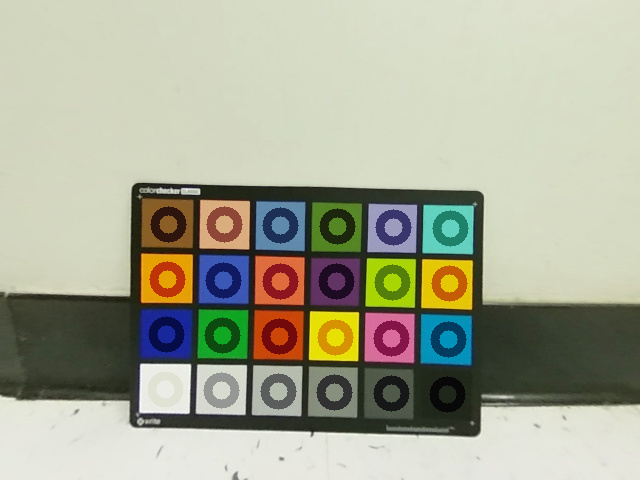} &
        \includegraphics[width=0.45\linewidth]{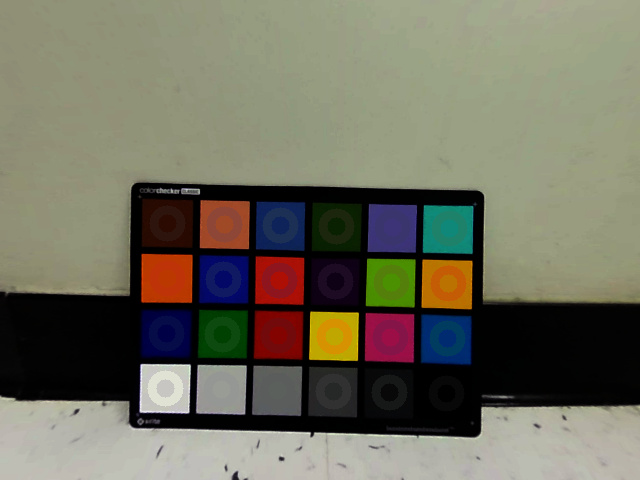} \\
        Before linearization & After linearization \\
    \end{tabular}
    \caption{\jf{
    Camera response function estimation. Using a Macbeth color checker, we fit per-channel $\gamma$ parameters on the uncorrected images (left) which are then be used to for linearization (right). On the bottom, we overlay the ground truth colors as circles to highlight the quality of the correction (bottom right).}}
    \label{fig:radiometric calibration}
\end{figure}

We employ OpenSFM~\cite{opensfm} (which natively supports 360\textdegree{} images) to perform bundle adjustment on all well-exposed cameras. 
We implement our method within the Nerfacto framework of Nerfstudio~\cite{nerfstudio}. Both phases (c.f.~\cref{sec:method-architecture}) are trained for 60k iterations, for a total of 120k training iterations. To train the NeRF, we extract eight perspective images with a 120\textdegree{} field of view from each panorama frame at $960\times 960$ pixel resolution. 
\jf{
We use a Macbeth color checker to estimate the camera response function (CRF). We use the same white balance setting (3500K) and the ``video'' capture mode, as in a regular PanDORA capture. After recording a static video, a single frame is extracted and cropped around the color checker. We automatically detect the color checker's squares using a Python implementation of the Macduff\footnote{Ryan Baumann, ``Macduff'', 2010, \texttt{\url{https://github.com/ryanfb/macduff}}.} color checker detection tool.
The uncorrected pixel values $\mathbf{z}$ are obtained by averaging the acquired pixels within each square patch of the color checker, and the ground truth colors $\mathbf{p}$ are provided by the chart manufacturer\footnote{X-Rite, ``New color specifications for ColorChecker SG and Classic Charts,'' n.d., \texttt{\url{https://www.xrite.com/service-support/new_color_specifications_for_colorchecker_sg_and_classic_charts}}.}.
We assume a CRF parametrized by $\mathbf{p} = \mathbf{z}^\gamma$ and optimize the $\gamma$ values through non-linear least squares optimization. Results of the radiometric calibration are shown in~\cref{fig:radiometric calibration}.}

\input{figures/dataset/dataset_figure}

%% file: figures/dataset/dataset_figure.tex
\newcommand{\myoverpic}[2]{
\begin{overpic}[width=0.195\linewidth]{#1}
    \put(0,0){%
      \tcbox[
        colback=white,
        colframe=white,
        arc=1pt,
        boxrule=0pt,
        left=0pt, right=0pt, top=0pt, bottom=0pt,
        enhanced
      ]
      {\tiny #2}
    }
  \end{overpic}
}

\begin{figure*}[t!]
    \centering
    \footnotesize
    \newcommand{\mywidth}{0.195\linewidth}
    \setlength{\tabcolsep}{0.5pt}
    \begin{tabular}{ccccc}
        \myoverpic{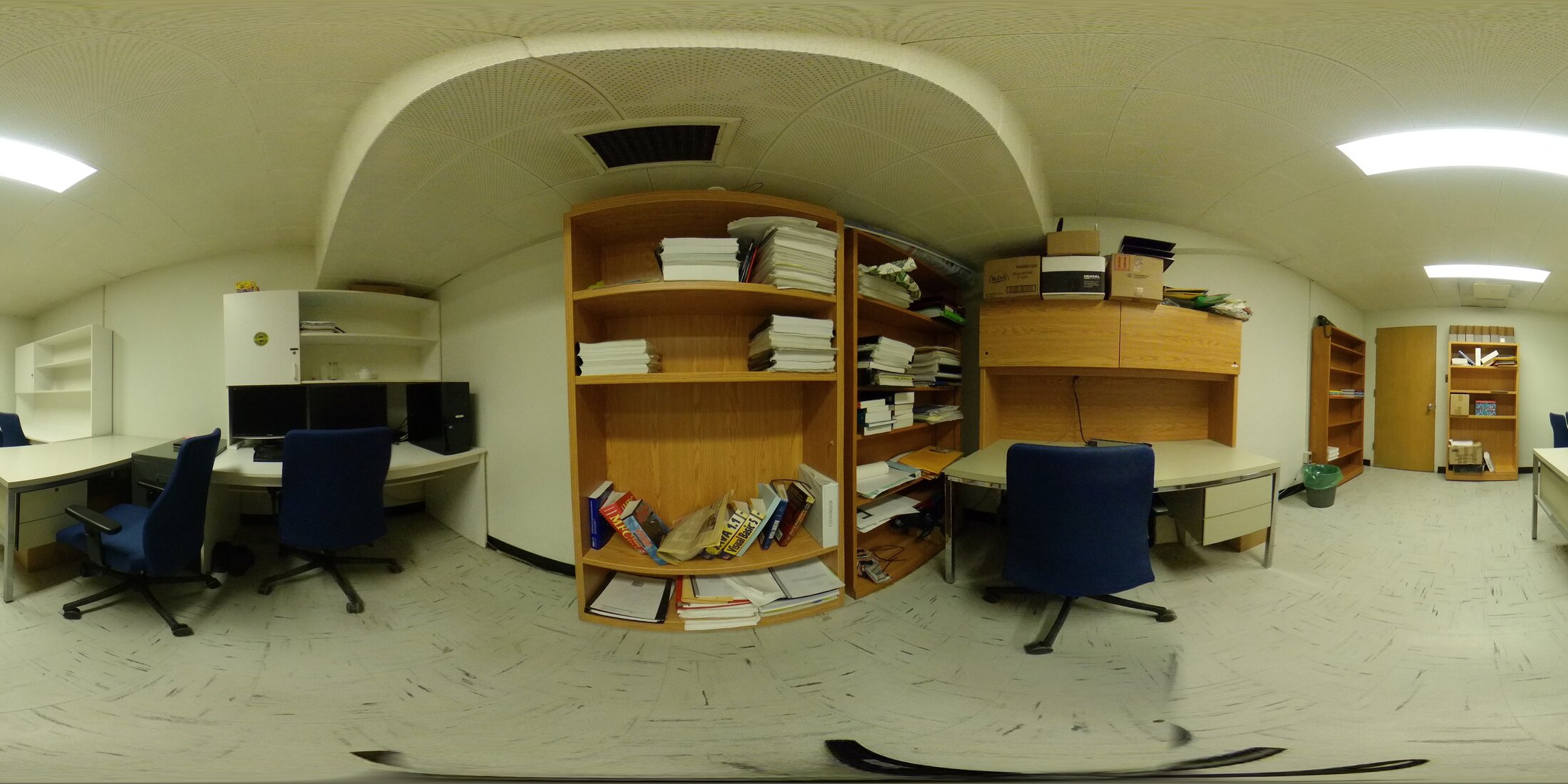}{\sceneg (125)} & 
        \myoverpic{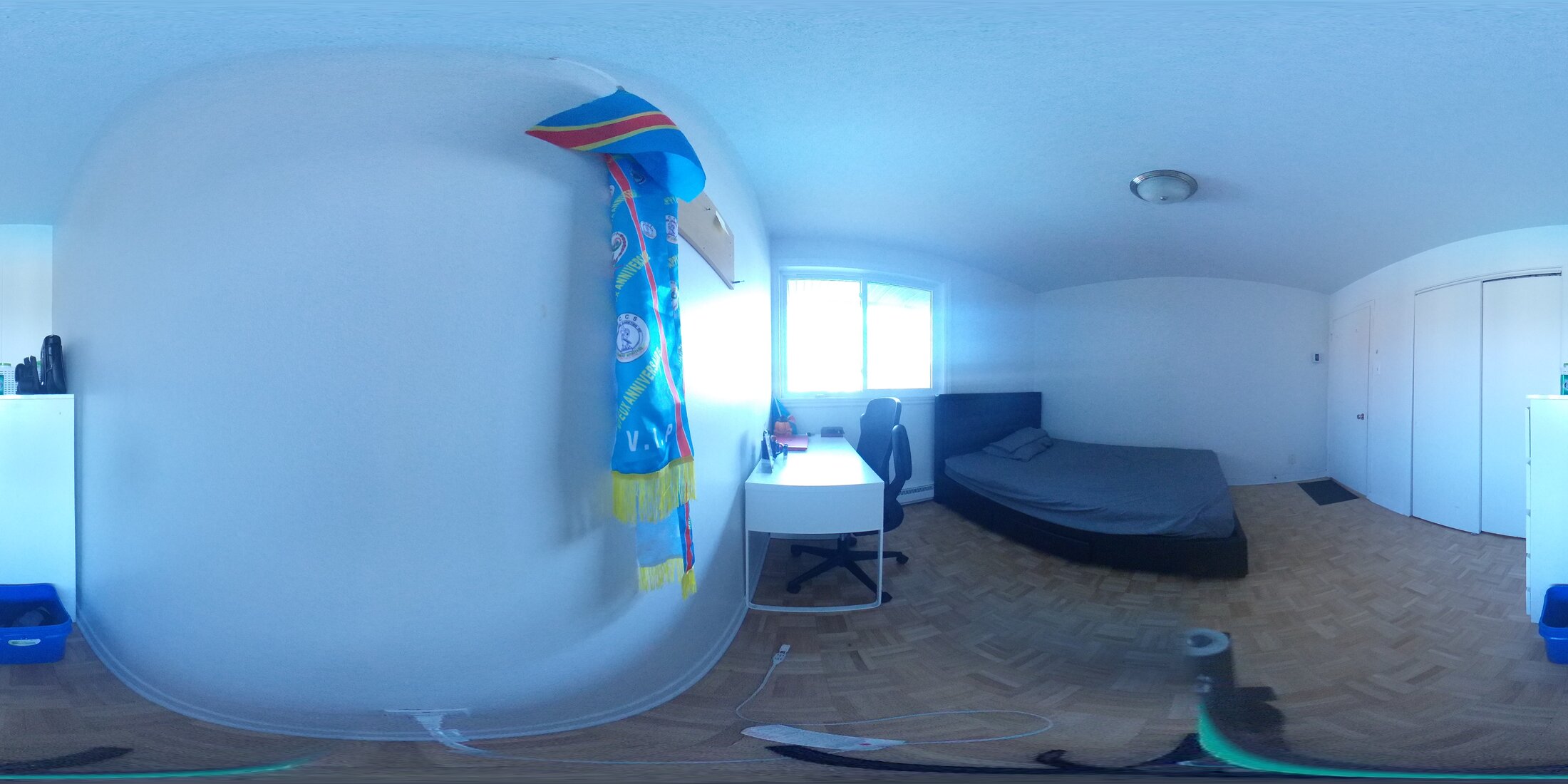}{\scenebluebed (500)} & 
        \myoverpic{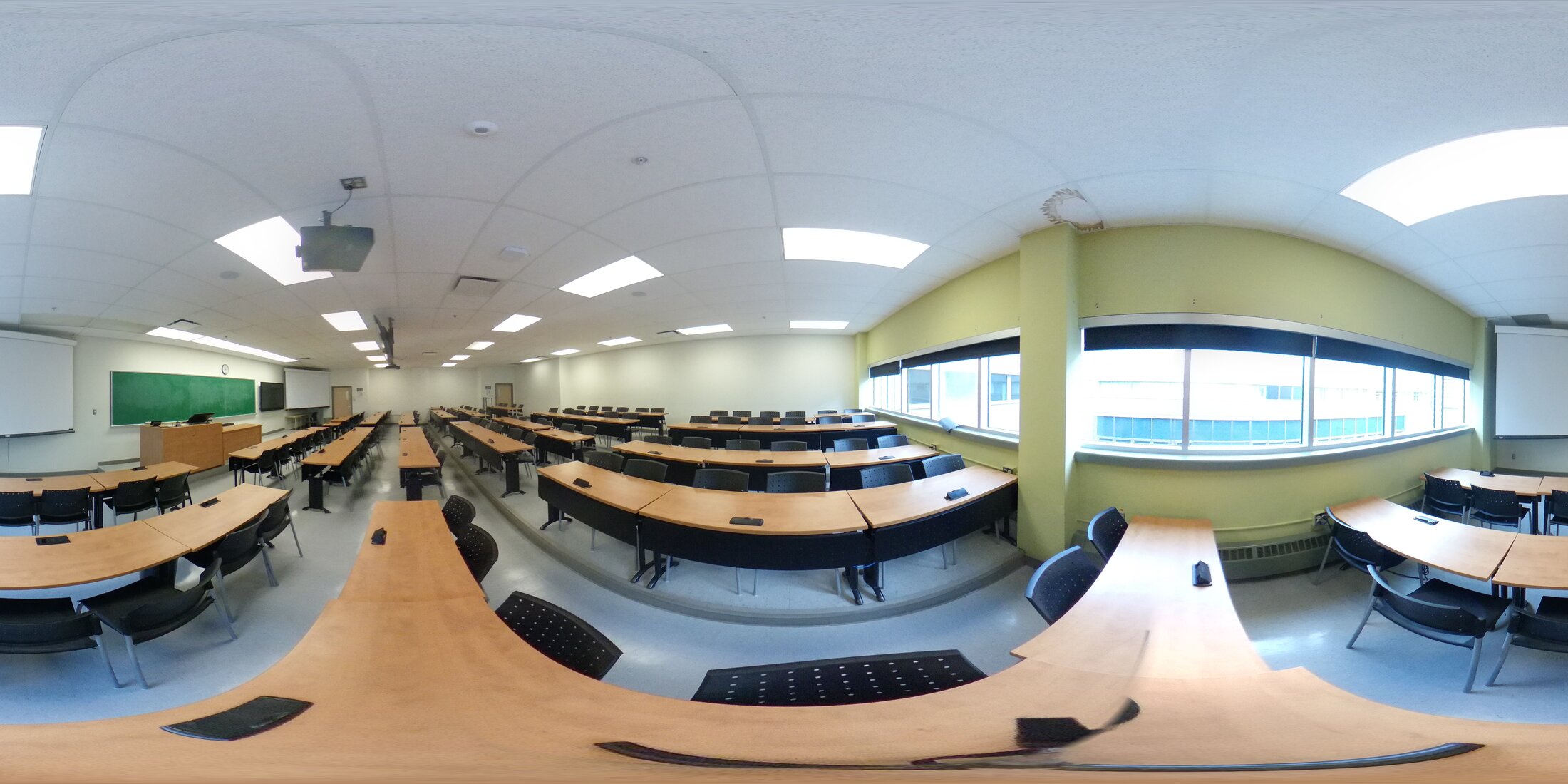}{\sceneclasswindow (78)} & 
        \myoverpic{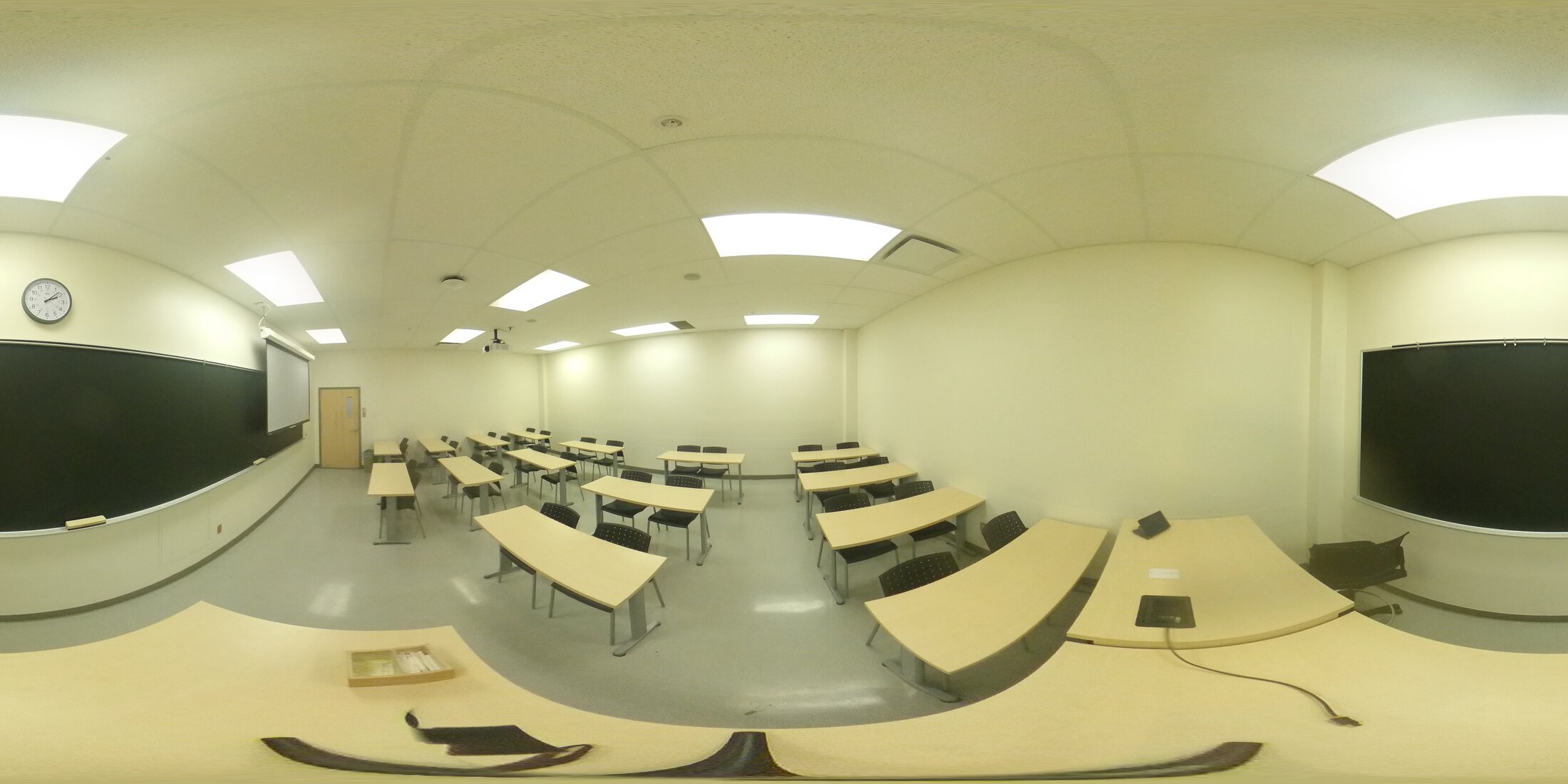}{\sceneclassnowindow (78)} & 
        \myoverpic{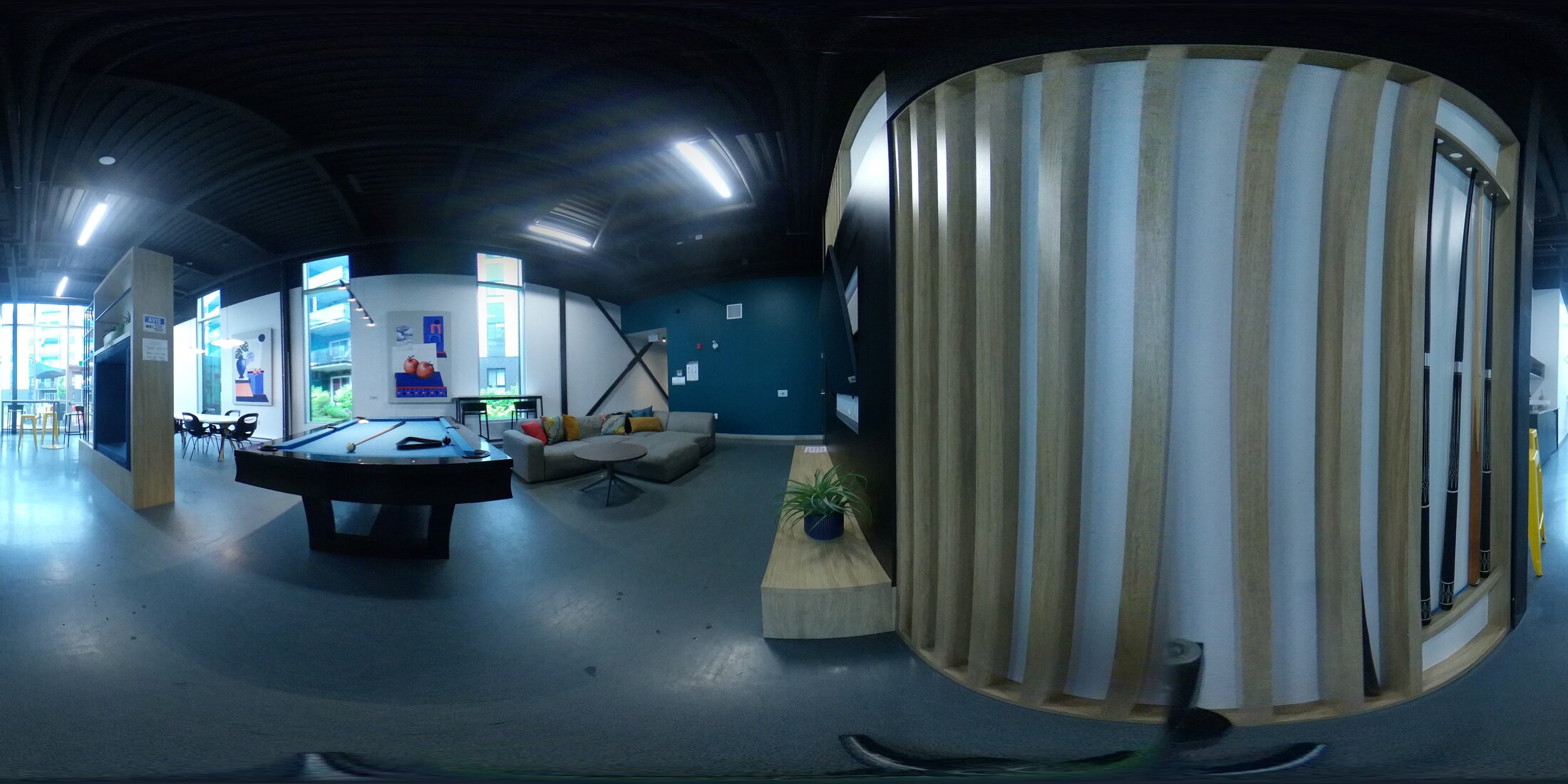}{\sceneclubhouse (200)} \\
        \myoverpic{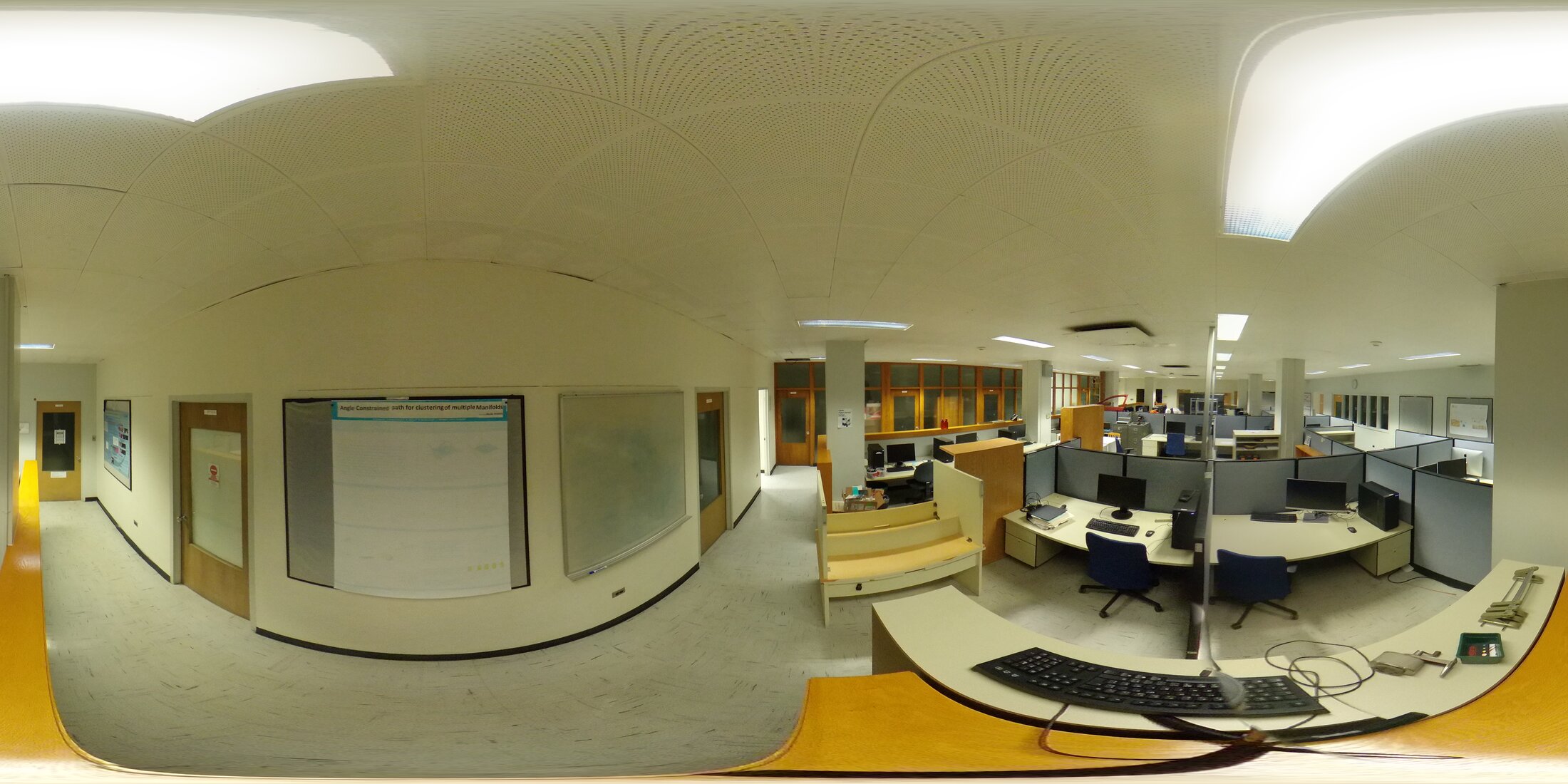}{\scenedownstairlab (500)} & 
        \myoverpic{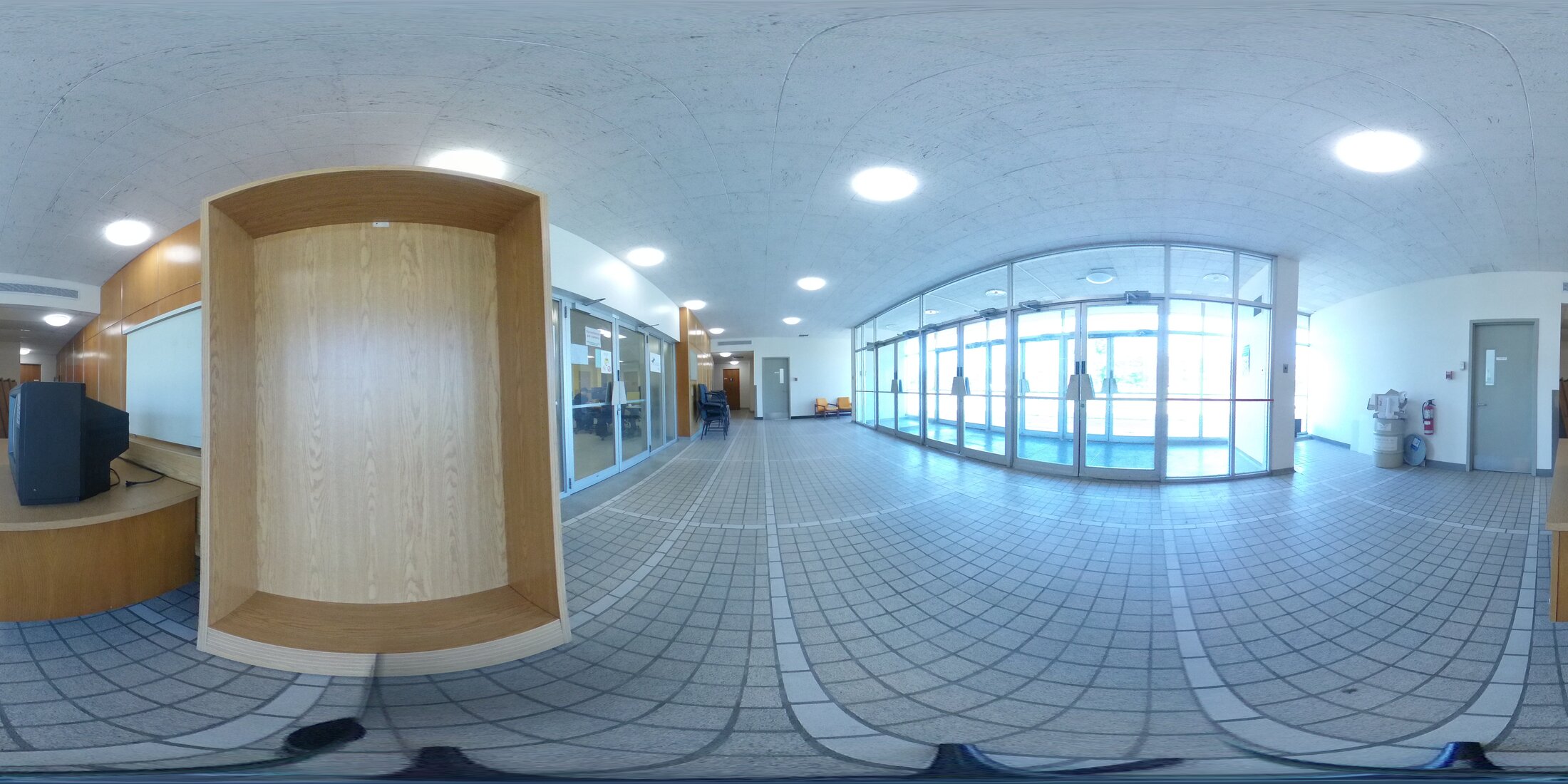}{\scenelablobby (62)} & 
        \myoverpic{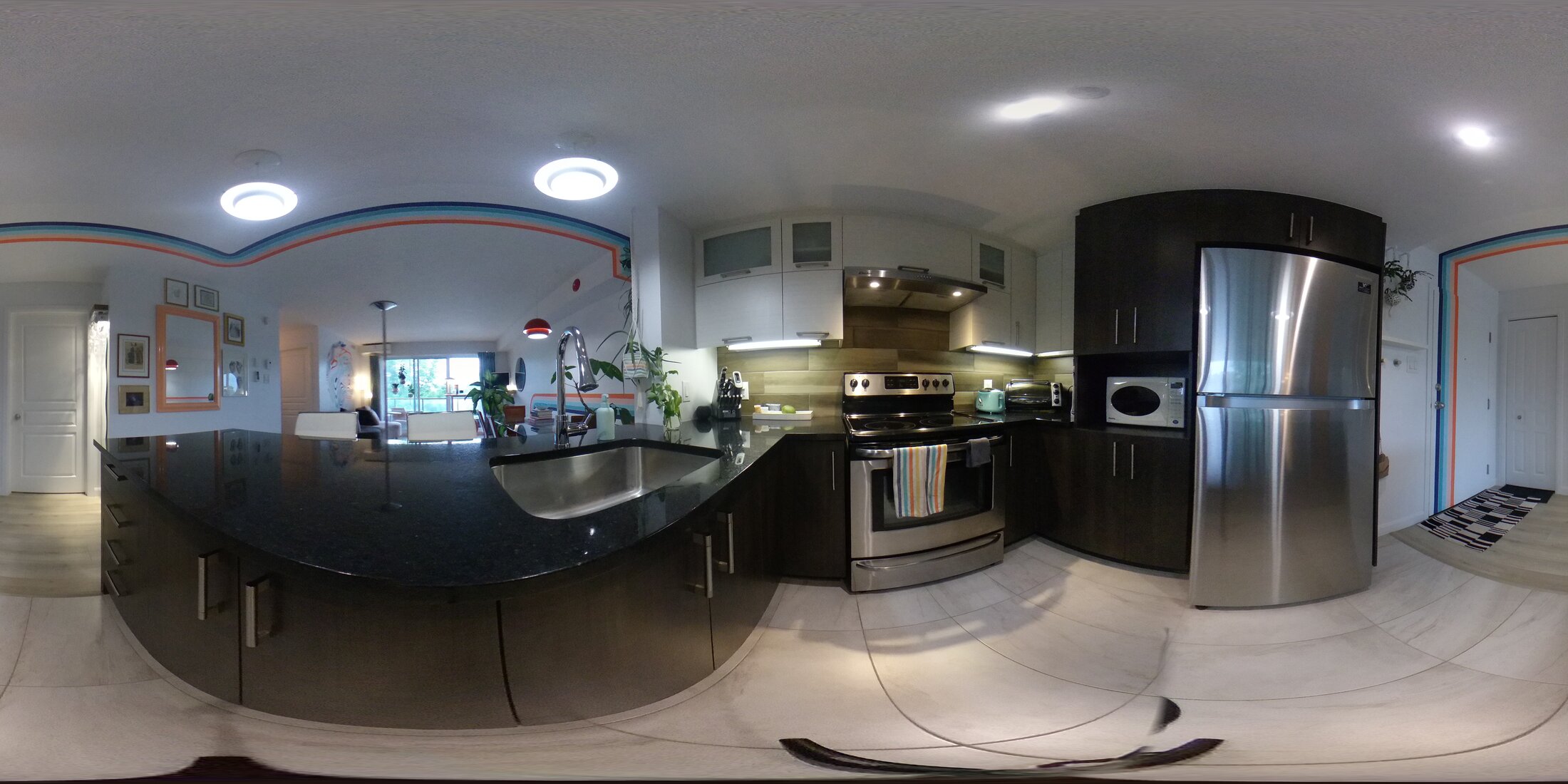}{\scenerainbowlivingroom (100)} & 
        \myoverpic{figures/dataset/audo_bright_GT14}{\sceneauditoriumbright (100)} & 
        \myoverpic{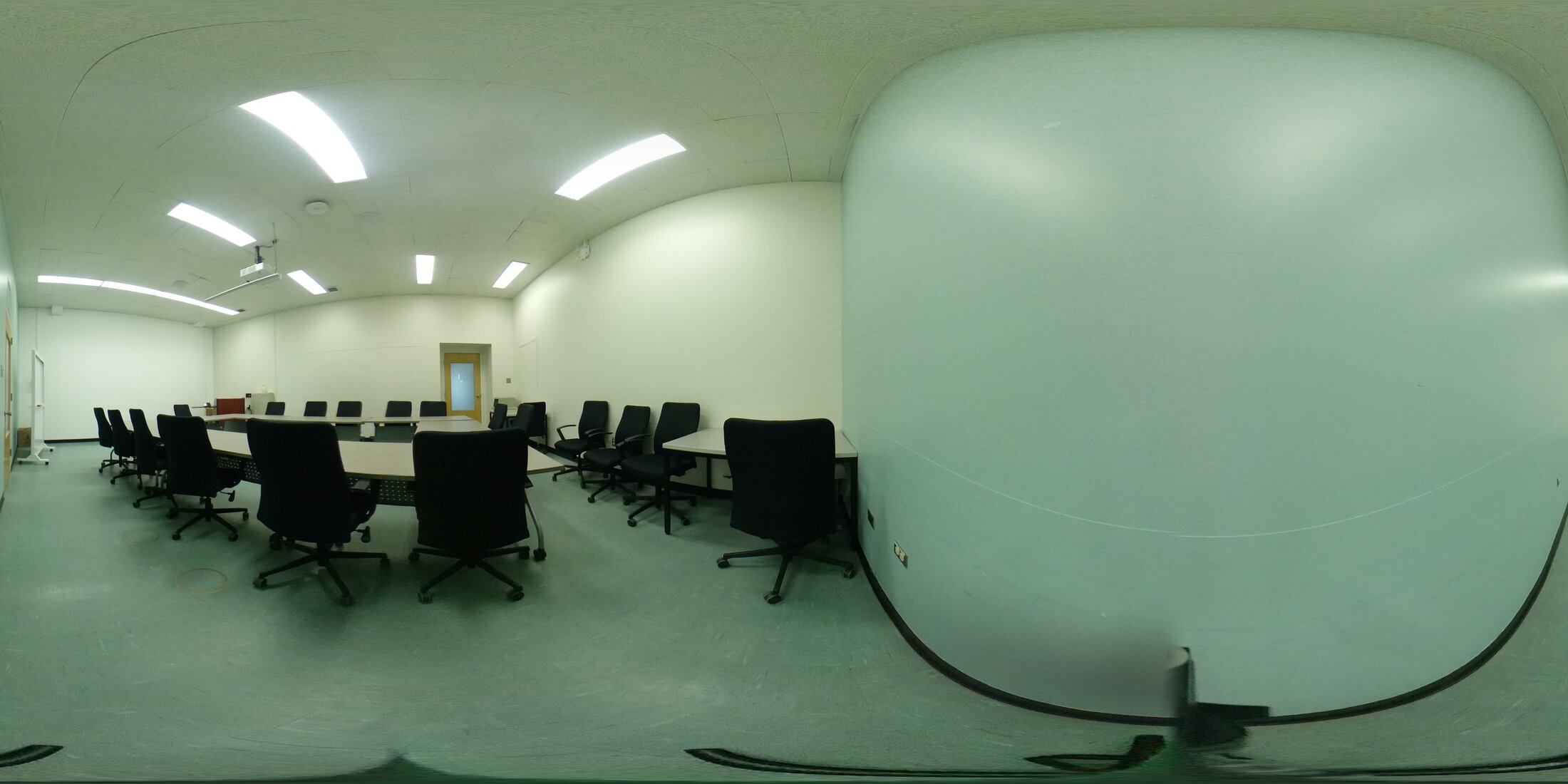}{\scenemeetingroom (100)} \\
    \end{tabular}
     \begin{tabular}{cccc}
     \myoverpic{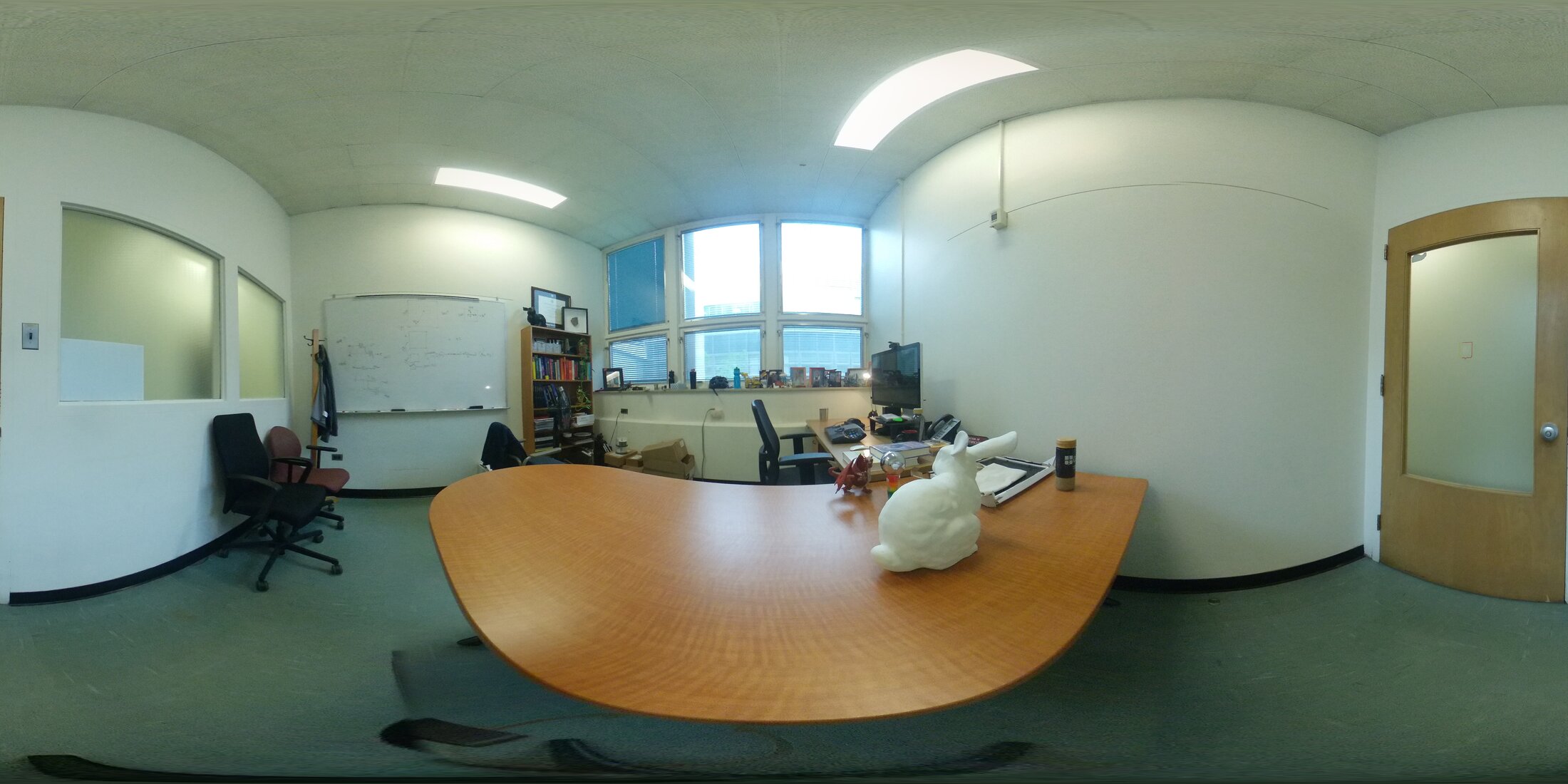} {\scenejfoffice (100)} & 
        \myoverpic{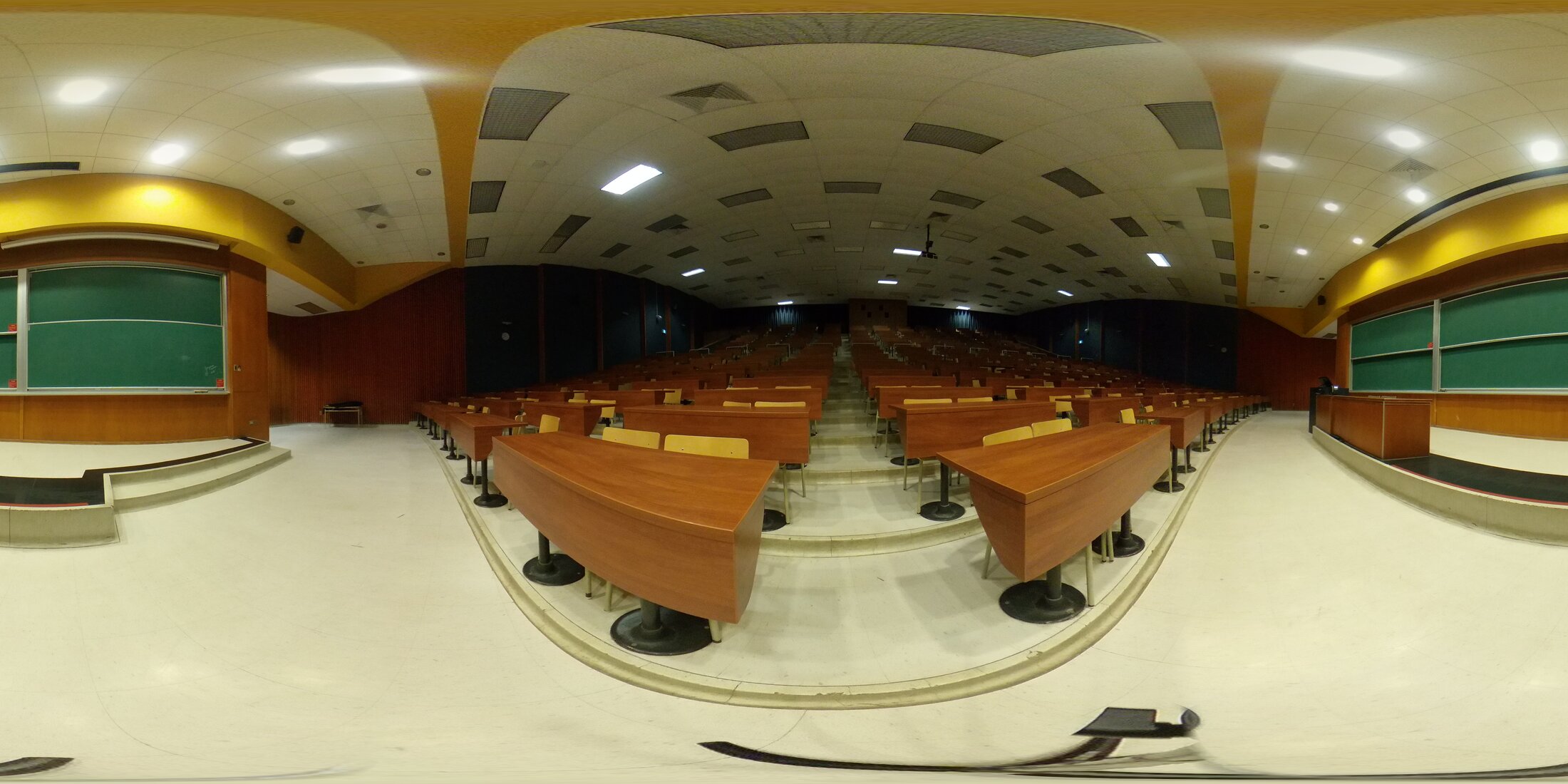} {\sceneauditoriumdark (500)} &
        \myoverpic{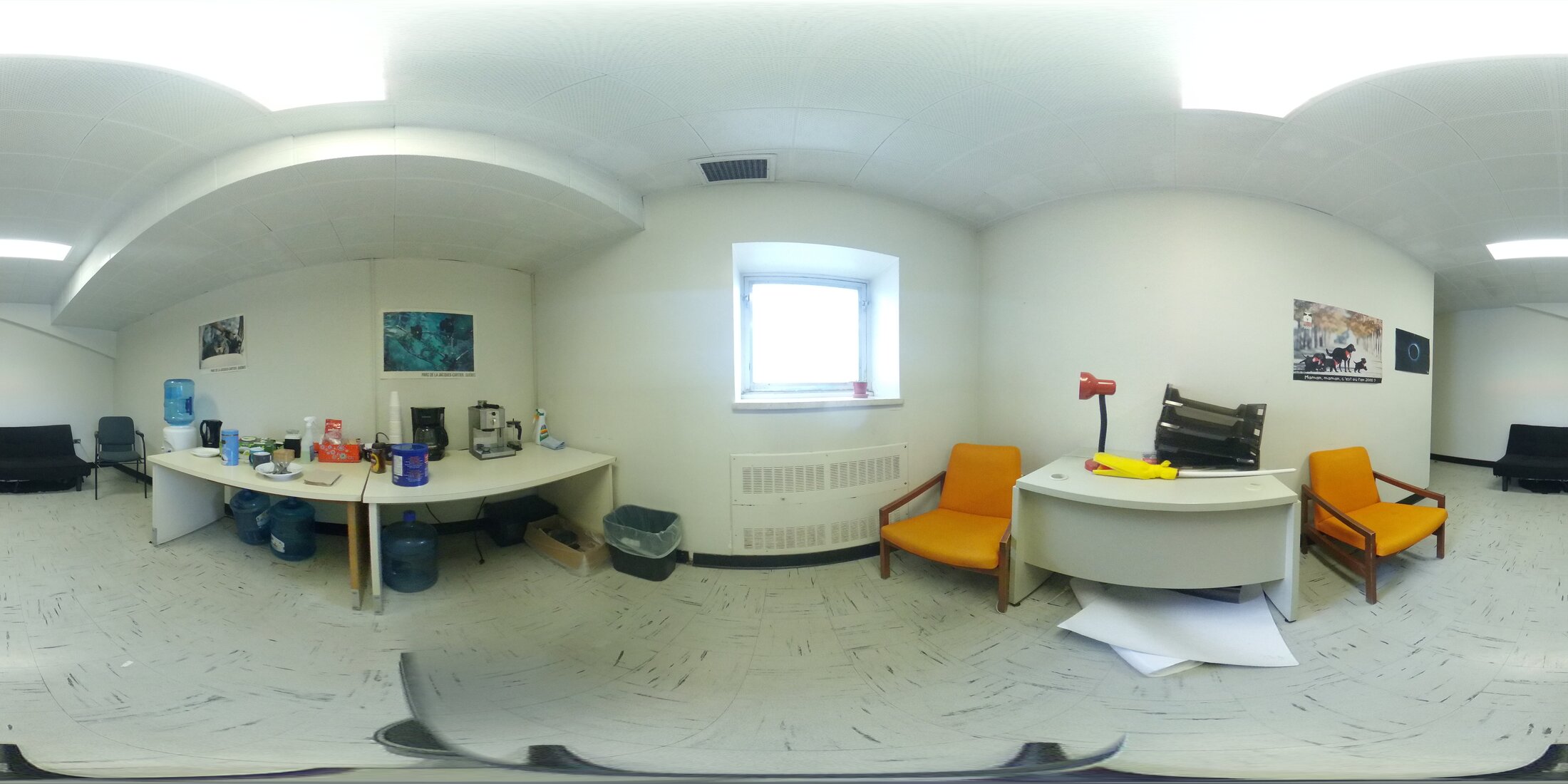} {\scenecoffeeroom (100)} &
        \myoverpic{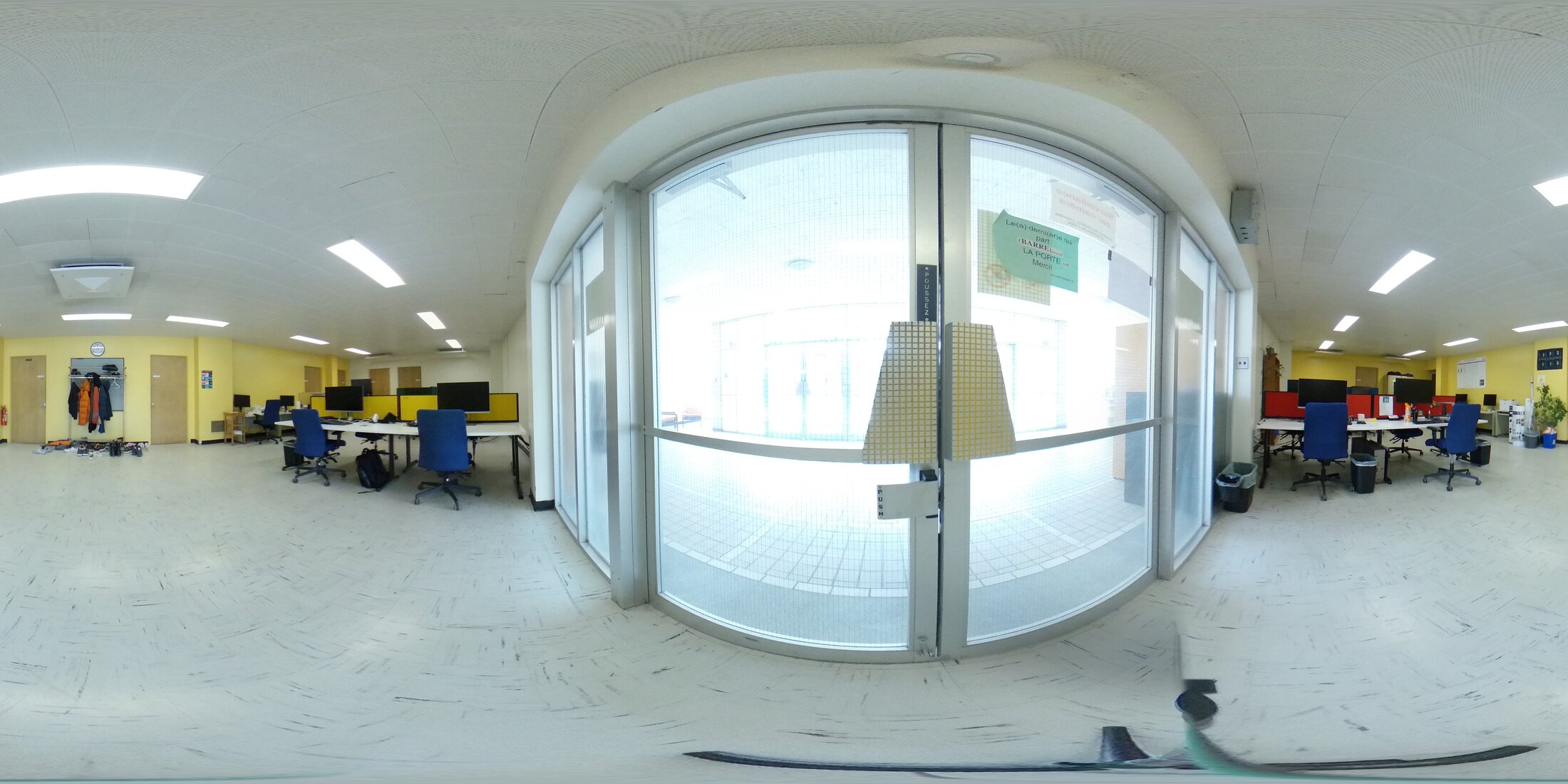} {\scenelvsn (78)} \\
     \end{tabular}
    \caption{Representative images of the 14 scenes captured in our evaluation dataset. The numbers between parentheses indicate each scene's corresponding exposure factor $1/\Delta t_\mathrm{f}$, i.e., the relative factor between the chosen well and fast exposures.}
    \label{fig:dataset}
\end{figure*}

%% file: sections/4_experiments.tex
\input{tables/quantitative}

\section{Evaluation dataset}
\label{sec:dataset}

We present a new dataset of 14 indoor scenes, representing a wide variety of interior spaces including classrooms, kitchens, living rooms, offices, and more. Each scene, previewed in \cref{fig:dataset}, was captured with our proposed \thename acquisition device, resulting in video sequences of approximately 2--3 minutes with two exposures. We set the cameras to ``manual'' with settings WB=3500K, ISO=800, f/2.1 for all the captures. Capturing while walking at a normal speed and slowly waving the capture apparatus is sufficient for a good reconstruction. \jf{Detailed dataset statistics are provided in \cref{tab:dataset-stats}.} Additionally, we captured a set of ground-truth HDR panoramas at 15 different locations for each scene (except the \scenelvsn and \sceneg that contain 18 and 12 GTs, respectively). To do so, a single Ricoh Theta Z1 camera was set to exposure bracketing mode to capture a sequence of 11 RAW (linear) exposures, which were subsequently merged using a standard HDR reconstruction algorithm~\cite{debevec2017recovering}. We register the pose of the ground truth cameras by including them in the OpenSFM reconstruction (cf. \cref{sec:camera_calibration}), while excluding them from the NeRF training. 
The dataset and code will be released publicly. Please see the supplementary material for more information about the dataset.

\begin{table}[t]
\scriptsize
\centering
\setlength{\tabcolsep}{6pt}
\caption{\jf{Per-scene statistics for all 14 real scenes in our dataset, including the number of training viewpoints~(``\#Train'') and the number of auxiliary frames used for pre-training calibration~(``\#Aux''). The table also indicates the exposure for the well-exposed camera~(``Well exp.'') and, since the fast exposure is held fixed at 1/25000\,s across all scenes, ``Exp.~ratio'' is the resulting well/fast exposure ratio.}}
\label{tab:dataset-stats}
\vspace*{-4mm} 
\begin{tabular}{lrrrr}
\toprule
Scene & \#Train & \#Aux & Well exp.\ (s) & Exp.\ ratio \\
\midrule
\scenelablobby          &  434 & 118 & 1/400 &  62$\times$ \\
\sceneclubhouse         &  708 & 172 & 1/125 & 200$\times$ \\
\scenerainbowlivingroom &  652 & 170 & 1/250 & 100$\times$ \\
\scenebluebed           &  218 &  58 & 1/50  & 500$\times$ \\
\scenemeetingroom       &  412 & 100 & 1/250 & 100$\times$ \\
\sceneauditoriumbright  &  310 &  74 & 1/250 & 100$\times$ \\
\sceneclassnowindow     &  570 & 138 & 1/320 &  78$\times$ \\
\sceneclasswindow       & 1250 & 300 & 1/320 &  78$\times$ \\
\scenedownstairlab      &  422 &  94 & 1/50  & 500$\times$ \\
\sceneg                 &  358 &  44 & 1/200 & 125$\times$ \\
\scenecoffeeroom        &  226 &  76 & 1/250 & 100$\times$ \\
\scenejfoffice          &  328 &  78 & 1/250 & 100$\times$ \\
\sceneauditoriumdark    &  258 &  68 & 1/50  & 500$\times$ \\
\scenelvsn              &  848 &  82 & 1/320 &  78$\times$ \\
\bottomrule
\end{tabular}
\end{table}

\section{Experiments}



\subsection{Baselines}
We first train the Nerfacto approach on the well-exposed panoramas alone as an LDR baseline, that approach is named ``LDR-Nerfacto'' onwards. We also select the following NeRF-based approaches to compare against: PanoHDR-NeRF~\cite{gera2022casual} and HDR-NeRF~\cite{huang2022hdrnerf}. \jf{Since there is no publicly-available implementation of NExF~\cite{niemeyer2025nexf}, HDR-NeRF~\cite{huang2022hdrnerf} is selected as a representative exposure-based baseline.} To ensure the improvements of our pipeline are not solely due to the use of a more powerful NeRF architecture, we implemented both methods in the same Nerfacto framework. 

\myparagraph{PanoHDR-NeRF~\cite{gera2022casual}} We replaced the original NeRF++ backbone~\cite{nerf++} with Nerfacto, which we dub ``PanoHDR-Nerfacto'' for clarity. Since our captured panoramas are not necessarily upright, we retrain the same LaNet inverse tonemapping network~\cite{yu2021lanet} on randomly-rotated HDR panoramas of $512 \times 256$ pixel resolution from \cite{gardner2017Learning}. We then upscale the prediction and blend it with the full-resolution LDR panoramas ($3840 \times 1920$) according to a mask defined by predicted HDR pixel values greater than 1. 

\myparagraph{HDR-NeRF~\cite{huang2022hdrnerf}} The original HDR-NeRF code did not converge on our data, likely due to challenges posed by large scenes and its assumption of cameras facing similar directions. We therefore reimplemented HDR-NeRF in the Nerfacto framework, \jf{by matching as closely as possible the original code while taking advantage of the strengths of the Nerfacto framework. Specifically, we keep the HDR radiance field MLP output in the log domain and add it to the log-exposure before feeding it to the MLPs. We swap the HDR radiance field to the Nerfacto backbone, and remove the ReLU activation function from the output layer of this backbone, as it would prevent the model from outputting necessary negative log-exposures. The three CRF MLPs from HDR-NeRF follow the original implementation (two-layer MLPs with a width of 128). We note the importance of disabling Nerfacto’s appearance embedding, which essentially makes the network invariant to different exposures. As in HDR-NeRF, we include the unit exposure loss during training.} We validated that our implementation yielded comparable results were obtained\footnote{For example, on their ``chair'' scene for ours/theirs: PSNR=32.19/32.87, SSIM=0.898/0.905, LPIPS=0.111/0.082.}. We dub this adaptation ``HDR-Nerfacto''. HDR-NeRF exploits overlapping exposures to learn the radiance (before applying their learned tonemapper), which fails in our setup due to the lack of overlap in the dynamic range between the two captured exposures. Thus, we use the tonemapper output in our evaluation, generating two sRGB outputs (regular and fast exposure) and merging them into a single HDR image. This approach empirically provides better results than the original method. See the supp. material for more information about our implementation. 

\input{figures/comparisons/comaprisons_methods}

\subsection{Comparative evaluation}
\label{sec:evaluation}

We train all methods on all scenes from our dataset from \cref{sec:dataset}, and report quantitative metrics with respect to the captured HDR ground truth in \cref{tab:quantitative-results}. 
We used three sets of metrics for the evaluation to properly assess the texture and energy of the predicted environment maps. For the texture, we used traditional metrics used in the NeRF literature such as PSNR, SSIM, and LPIPS on the tone-mapped output (\cref{tab:quantitative-results}, columns 2--4) to assess the quality of the texture of the predicted environment map. We also evaluate the quality of HDR image reconstruction using the common HDR metrics \cite{hanji2022comparison} such as PU-PSNR and PU-SSIM~\cite{mantiuk2021pu}, and HDR-VDP3~\cite{mantiuk2023hdrvdp3} in \cref{tab:quantitative-results} (columns 5--7). Finally, we evaluate the suitability of the HDR outputs to be used for image-based lighting by rendering the virtual scene from \cite{gera2022casual}, on which we compute RMSE, si-RMSE, and RGB angular error~\cite{legendre2019deeplight} directly on the HDR renderings (\cref{tab:quantitative-results}, columns 8--10), and PSNR on the tone-mapped renderings (\cref{tab:quantitative-results}, column 11). We observe that \thename outperforms HDR baselines, while remaining competitive with the ideal LDR-Nerfacto baseline on tonemapped LDR images. Per-scene results are provided in the supplementary materials.


\input{figures/fast_exposed_results}

These observations are validated qualitatively in \cref{fig:qual}. While LDR-Nerfacto produces high-quality reconstructions, its LDR output limits realism when using it to relight virtual objects. PanoHDR-NeRF provides better environment maps for image-based lighting, but often yields underexposed renderings, possibly due to under-shooting or inconsistent HDR upscaling. It is also less color accurate (as demonstrated by the RGB angular error metric in \cref{tab:quantitative-results}). In contrast, HDR-Nerfacto produces more accurate results, resulting in more realistic renderings, but generates overly blurry environment maps. Finally, \thename produces higher-quality, sharper environment maps similar to those of LDR-trained models, while preserving realistic lighting close to the ground truth. Observe how the various lighting effects on the virtual objects (shading, cast shadows, specularities, caustics, color and intensity, etc.) closely match the ground truth HDR lighting (rightmost column). \jf{These effects can be generated because our method reconstructs the very high dynamic range light sources more accurately than other methods, as shown in~\cref{fig:fast_exposed}.} We also validate this by comparing the pixel intensity histograms against the ground truth radiance of the \sceneclubhouse scene in \cref{fig:histograms}. While \cite{gera2022casual} and \cite{huang2022hdrnerf} underestimate dynamic range, \thename aligns most closely with the ground truth radiance of the scene. 

\input{figures/histograms/histogram}

\begin{figure} [t!]
    \centering
    \newlength{\oiwidth}
    \setlength{\oiwidth}{0.4\linewidth}
    \setlength{\tabcolsep}{1pt}
    \begin{tabular}{cc}
        \includegraphics[width=\oiwidth]{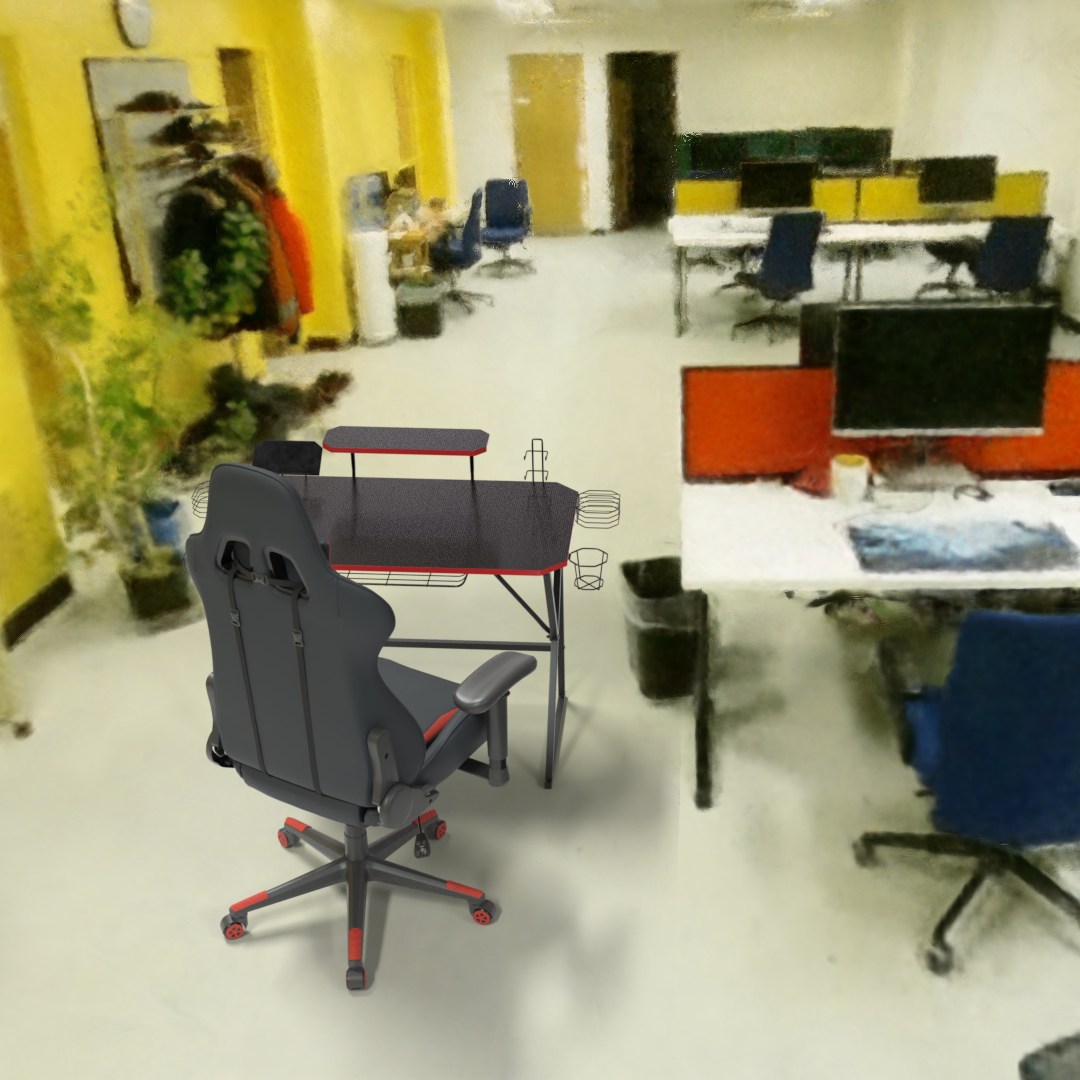} & 
        \includegraphics[width=\oiwidth]{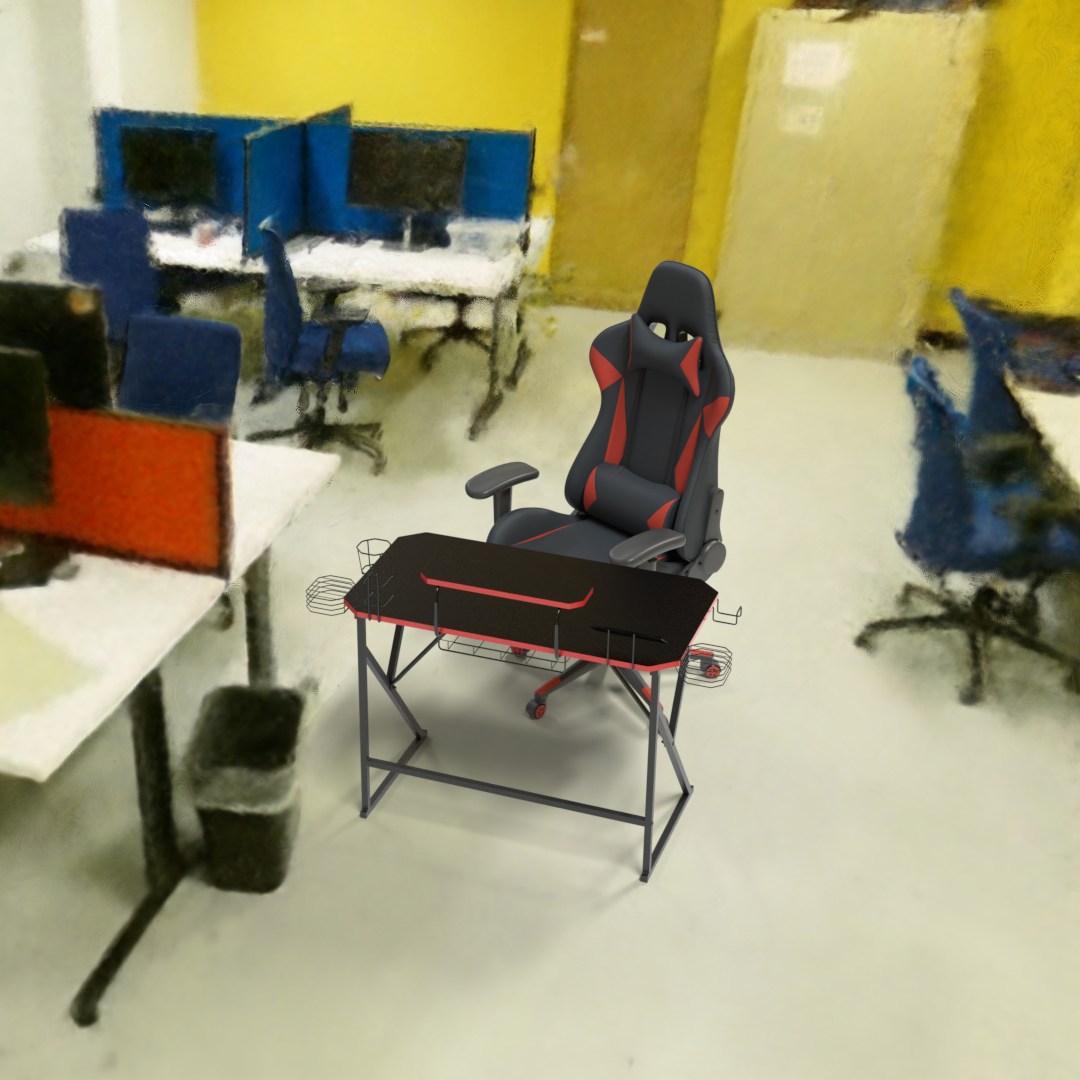} 
    \end{tabular}
    \caption{Relighting virtual objects (red-black chair and desk) from two different viewpoints in a scene. The objects and a planar shadow catcher are composited onto the background, which is also rendered by our method. Notice how the lighting produces both realistic shading on the objects and shadows on the ground. }
    \label{fig:object-insertion}
\end{figure}

\input{tables/ablation}

\subsection{Spatially-varying lighting}
\label{sec:spatially_varying}

\jf{Our method produces spatially-varying HDR radiance fields that enable the insertion of virtual objects, coherent relighting, and viewing from multiple viewpoints, as shown in \cref{fig:teaser}(c) and \cref{fig:object-insertion}. In \cref{fig:teaser}(c), multiple objects are rendered at different positions in the scene. Notice how the armadillos under the tables are significantly darker than the ones on the tables. \Cref{fig:object-insertion} shows virtual objects rendered from two different viewpoints. To further illustrate the effectiveness of our approach at learning lighting in a spatially-varying manner, we select the three scenes with the greatest spatial diversity in lighting---\sceneclubhouse, \scenecoffeeroom, and \scenerainbowlivingroom---and present the results in \cref{fig:spatially_varying}. Notice how our approach successfully reconstructs the lighting across the full dynamic range of each scene, from the brightest regions near the windows to the darkest corners.}

\input{figures/spatially_varying_results/spatially_varying}

\subsection{\jf{Additional analyses}}
\label{sec:ablation}

\myparagraph{Comparison of capture strategies} Our aim is to simplify spatially-varying HDR capture for indoor scenes. To do so, we propose that capturing two simultaneous exposures (well- and fast-exposed) is a good compromise between ease of capture and HDR accuracy for IBL. Here, we compare this strategy with other options. To do so, we conducted a comparative analysis across ten scenes, where exposures from the GT HDR panoramas were selected to avoid any bias which may arise from the underlying NeRF model. Results are reported in \cref{tab:ablation-two-exposure}. First, selecting a single exposure---either only the exposure corresponding to the well-exposed setting, ``well-exposed only'', or lifting that exposure to HDR using an inverse tonemapping network ``well+LDR2HDR''---yields underperforming metrics. This justifies the need to shoot a second exposure. We next compare two options, where we combine the well-exposed setting with: the fastest exposure (``well+fast'') and a mid-range exposure (halfway between the fast- and well-exposed frames, ``well+mid''). As expected, including a second exposure improves metrics, but selecting the fastest exposure is what results in a reconstruction closer to the full HDR ground truth. 
Finally, capturing more than two simultaneous exposures (e.g., with a third camera) would increase complexity (synchronization, calibration, etc.) while reducing coverage (since cameras occlude each other).

\begin{figure}[t!]
    \centering
    \setlength{\tabcolsep}{1pt}
    \begin{tabular}{c}
        \includegraphics[width=0.8\linewidth]{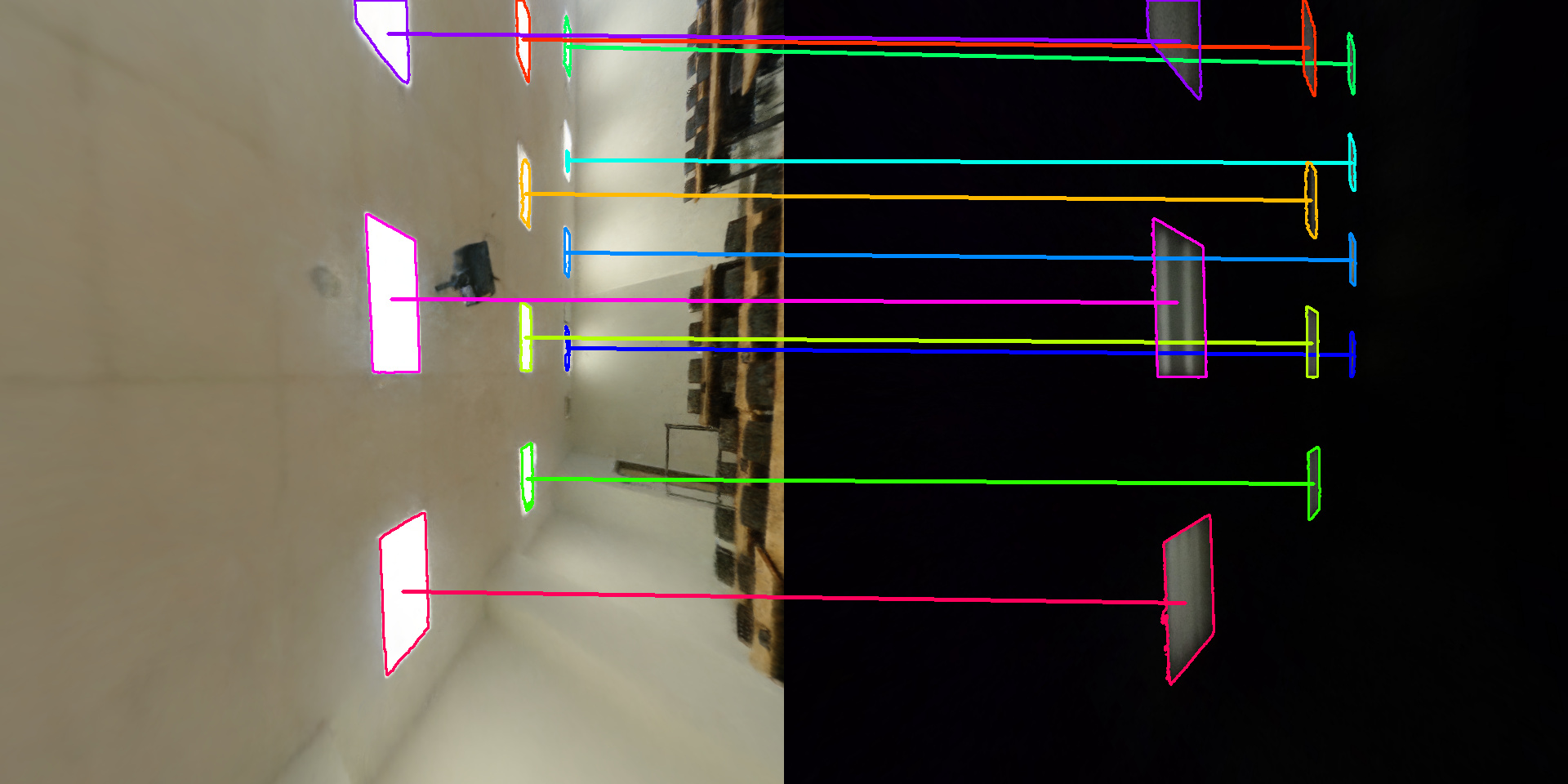} \\
        \includegraphics[width=0.8\linewidth]{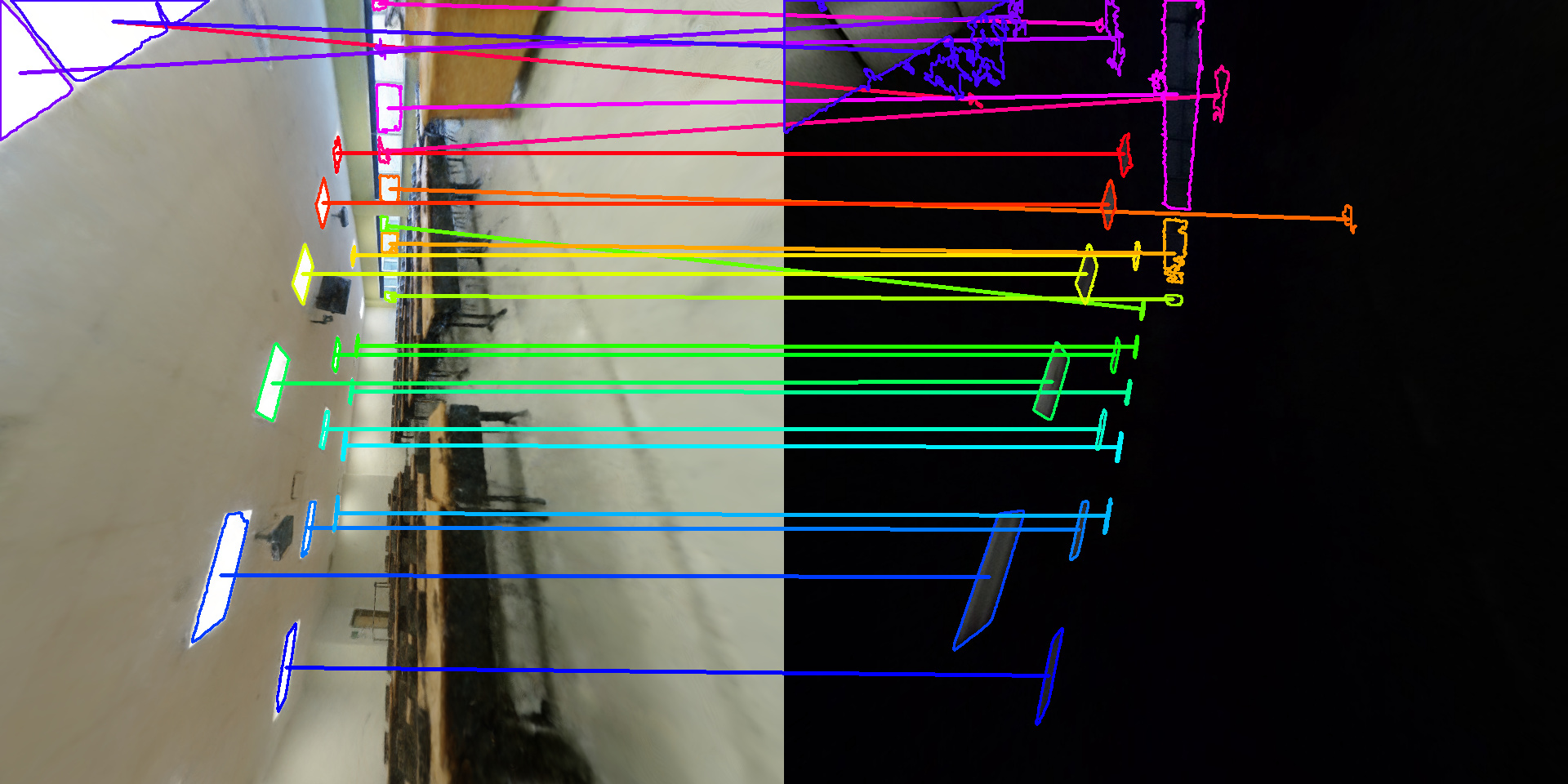}
    \end{tabular}
    \caption{\jf{Success (top) and failure (bottom) of the fine alignment. In fast-exposed images, windows under-saturate into noisy, featureless dark regions, while well-exposed images preserve the detailed window structure (panes, frames), causing contour matching to fail (bottom). Example shown from the \sceneclasswindow scene.}}
    \label{fig:fine_alignment_challenge}
\end{figure}

\myparagraph{Impact of fine alignment during training} 
\jf{Both quantitative (\cref{tab:ablation-pandora}) and qualitative (\cref{fig:alignment}) results confirm the benefit of our fine alignment on both LDR and HDR metrics, particularly in challenging scenes such as \sceneauditoriumbright, which contains numerous small, bright light sources. Although the fine alignment demonstrates robustness in scenes with artificial lighting, its performance degrades in environments where light sources span the intensity range between the two camera exposures. This limitation is present in the windows in \sceneclasswindow, as illustrated in \cref{fig:fine_alignment_challenge}. In fast-exposed images, these windows appear as noisy, featureless dark regions due to under-saturation, whereas well-exposed images retain detailed window structures, including panes and frames. This occurs because the true intensity of the windows exceeds the dynamic range of the well-exposed camera and falls below that of the fast-exposed camera. This is a challenge for the contour-matching step: a single saturated blob in the fast exposure has no shape correspondence to the multiple sub-contours in the well-exposed image. As a result, matching is often more robust without visible windows (\cref{fig:fine_alignment_challenge}, first row), but degrades noticeably with them (second row). In these cases, one can detect when estimated homographies result in too large a displacement and simply ignore them.}

\input{tables/gsplats_vs_ours}

\subsection{3D Gaussian Splatting}

3D Gaussian Splatting~\cite{kerbl20233d} (3DGS) has revolutionized novel view synthesis by significantly accelerating rendering speeds. Although our method is based on NeRF, our design principles are equally applicable to 3DGS. To demonstrate this, we adapt our NeRF-based approach to 3DGS by attaching two sets of Spherical Harmonics (SH) coefficients to each 3D Gaussian primitive, one for each of the well- and fast-exposures. During the initial 60k iterations, optimization is confined to the well-exposed sequence. For the subsequent 60k iterations, we freeze the geometry and well-exposed SH coefficients, focusing optimization exclusively on the fast-exposure SH coefficients. As in \cref{sec:implementation-details}, the 3DGS is trained on perspective crops. At inference time, both well- and fast-exposure SH coefficients are evaluated, and resulting RGB values are merged into HDR using \cref{eq:hdr-merge}. 360\textdegree{} panoramas for IBL are obtained with \cite{li2025omnigs}. \jf{
We implement this 3DGS variant within the same Nerfstudio framework as our NeRF-based pipeline.} Comparative results between NeRF and 3DGS across ten scenes are detailed in \cref{fig:gsplats,tab:nerf-vs-gsplats-results}. As shown in \cref{fig:gsplats}, NeRF shows more robust performance in HDR radiance field reconstruction, especially for the fast-exposure part of the dynamic range. This is also reflected quantitatively in \cref{tab:nerf-vs-gsplats-results}. While NeRF achieves higher fidelity, it is slow and unsuitable for real-time applications, as rendering a $1920 \times 3840$ equirectangular image takes $\sim$13s on a TITAN RTX, vs. 100ms for 3DGS, making it a better choice for applications such as augmented reality.


%% file: tables/quantitative.tex
\begin{table*}
\small
\centering
\setlength{\tabcolsep}{2pt}
\caption{Quantitative results on our dataset of 14 real scenes. Metrics are shown in 4 groups (left to right): LDR panoramas, HDR panoramas, HDR and LDR renders (``LDR r.''). For ``renders'', we use the HDR panoramas to render a virtual scene (see \cref{fig:qual}), the metrics are computed on the result. Results are color coded by \colorbox{red!25}{best} and \colorbox{orange!25}{second-}best.}
\vspace*{-4mm} 
\label{tab:quantitative-results}
\begin{tabular}{lccccccccccccc}
\toprule
& \multicolumn{3}{c}{LDR panos}
& & \multicolumn{3}{c}{HDR panos}
& & \multicolumn{3}{c}{HDR render}
& & LDR r. \\
Method
& PSNR$_\uparrow$
& SSIM$_\uparrow$
& LPIPS$_\downarrow$
& \,
& PU-PSNR$_\uparrow$ 
& HDR-VDP$_\uparrow$
& PU-SSIM$_\uparrow$
& \,
& si-RMSE$_\downarrow$
& RMSE$_\downarrow$
& RGB ang.$_\downarrow$ 
& \,
& PSNR$_\uparrow$ \\
\midrule
LDR-Nerfacto	
&   \cellcolor{red!25} 20.02	& \cellcolor{red!25}0.66	& \cellcolor{red!25}0.40	
& & 26.81	&  6.25	& \cellcolor{red!25}0.90	
& & 0.28	& 0.40	& 5.40	
& & 27.97 \\
PanoHDR-Nerfacto	
& 19.51	& 0.62	& \cellcolor{orange!25}0.44	
& & \cellcolor{orange!25} 28.43	& \cellcolor{red!25} 6.64	& \cellcolor{red!25}0.90	
& & \cellcolor{orange!25}0.18	& \cellcolor{orange!25}0.31	& 6.25	
& & \cellcolor{orange!25}29.18 \\
HDR-Nerfacto	
&   19.27	& 0.60	& 0.50	
& & 27.59	& 6.34	& \cellcolor{orange!25}0.89	
& & 0.23 & 0.36	& \cellcolor{orange!25}4.07	
& & 28.26 \\
\thename (ours)	
&   \cellcolor{orange!25}19.83	& \cellcolor{orange!25}0.65	& \cellcolor{red!25}0.40	
& & \cellcolor{red!25}29.08	& \cellcolor{red!25}6.64	& \cellcolor{red!25}0.90	
& & \cellcolor{red!25}0.17	& \cellcolor{red!25}0.26	& \cellcolor{red!25}3.59	
& & \cellcolor{red!25}29.42 \\
\bottomrule
\end{tabular}

\end{table*}

%% file: figures/comparisons/comaprisons_methods.tex
\begin{figure*}[!t]
    \centering
    \footnotesize
    \newlength{\mywidth}
    \setlength{\tabcolsep}{1pt}
    \setlength{\mywidth}{0.2\linewidth}
    \begin{tabular}{ccccc}
    LDR-Nerfacto & PanoHDR-Nerfacto & HDR-Nerfacto & \thename (ours) & GT   \\
     \includegraphics[width=\mywidth]{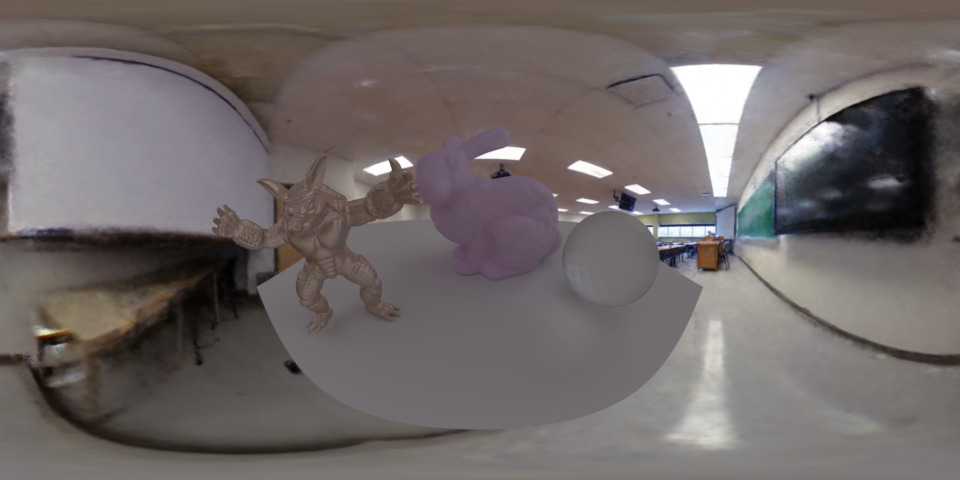} &
     \includegraphics[width=\mywidth]{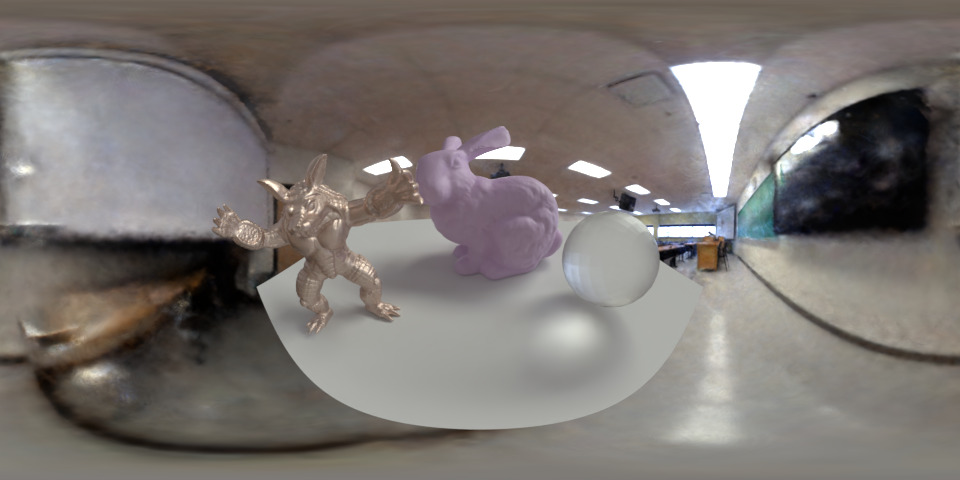} &
     \includegraphics[width=\mywidth]{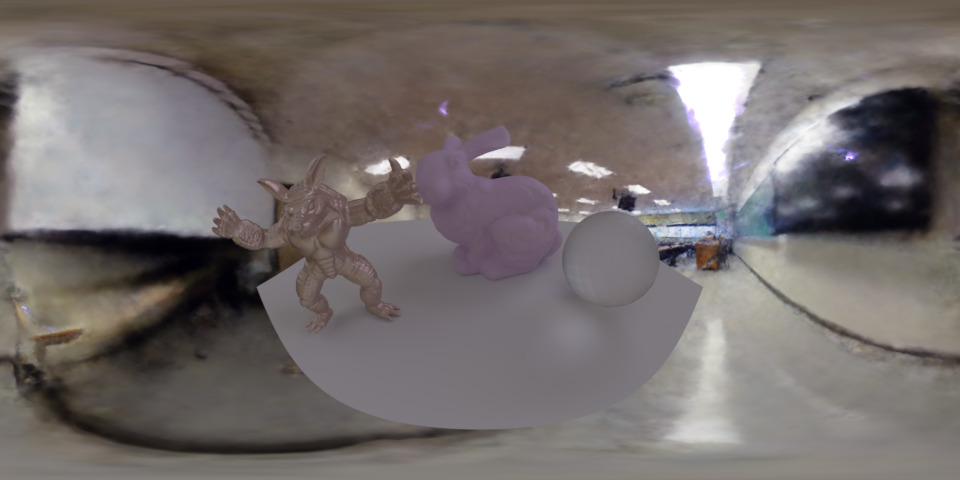} &
     \includegraphics[width=\mywidth]{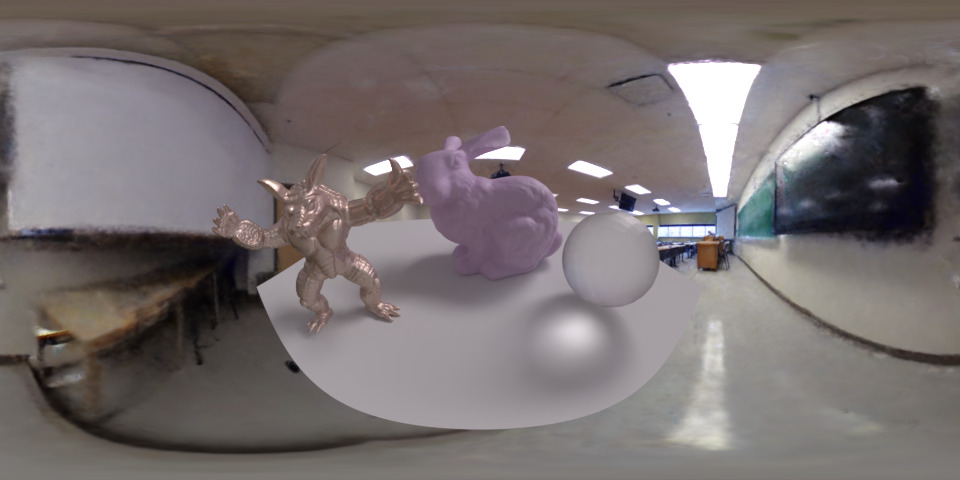} & 
     \includegraphics[width=\mywidth]{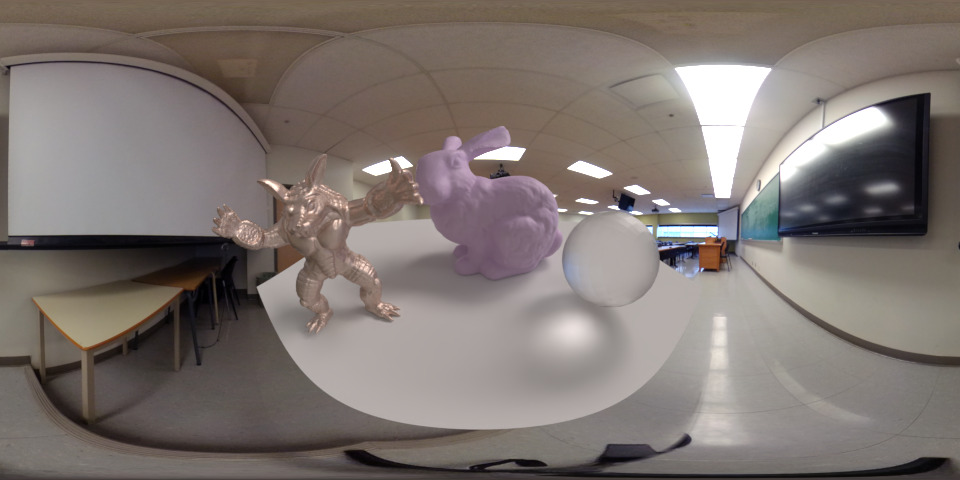} \\
     
     \includegraphics[width=\mywidth]{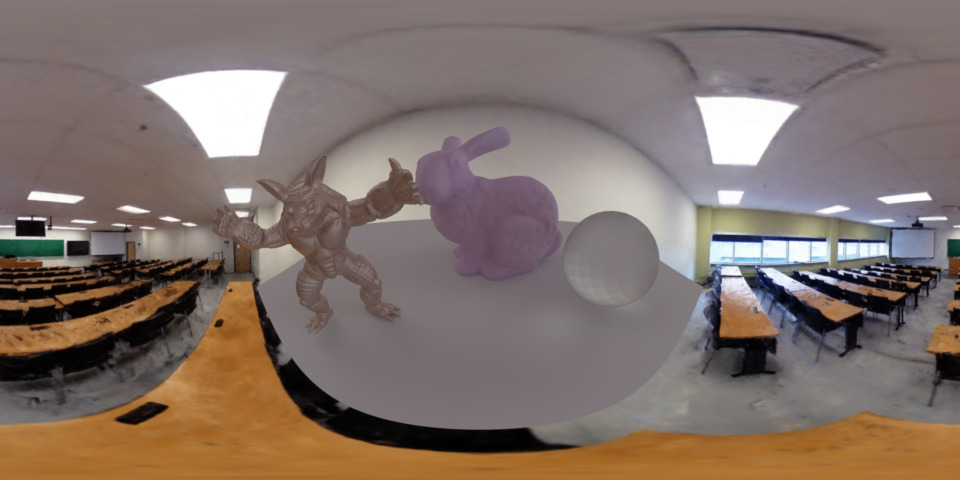} &
     \includegraphics[width=\mywidth]{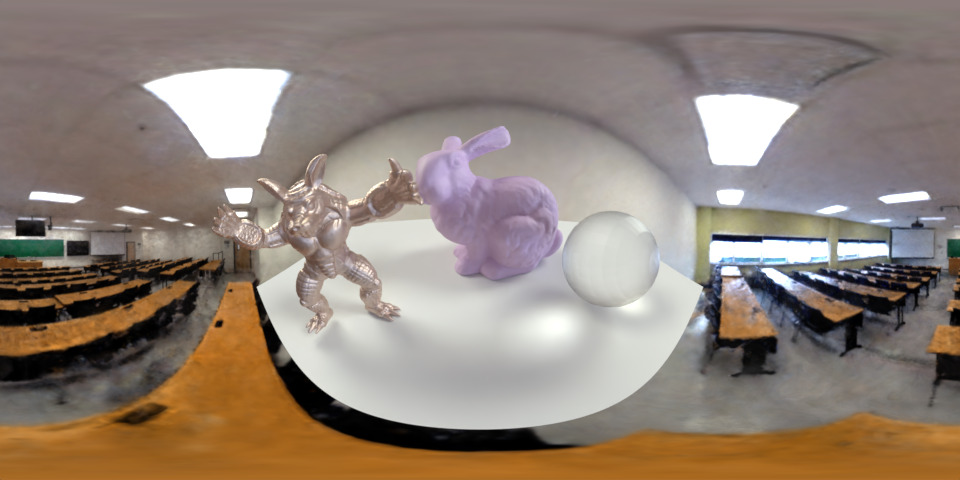} &
     \includegraphics[width=\mywidth]{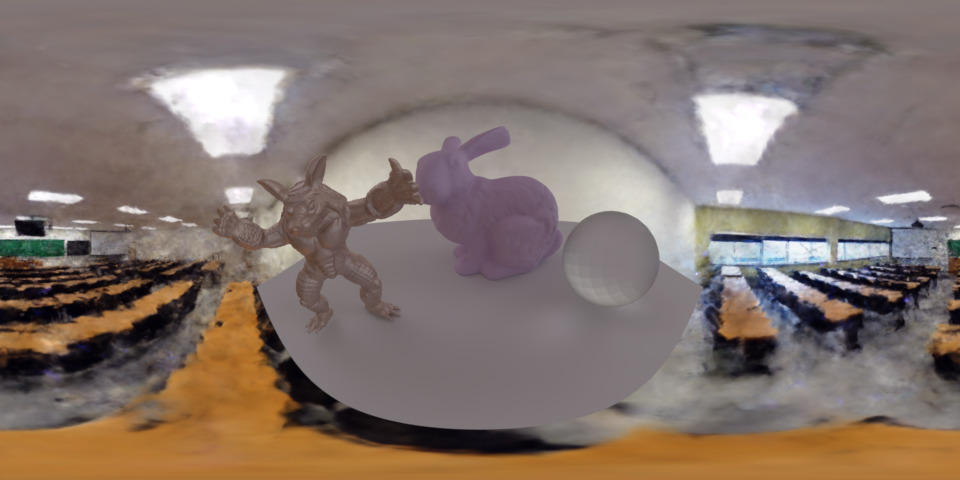} &
     \includegraphics[width=\mywidth]{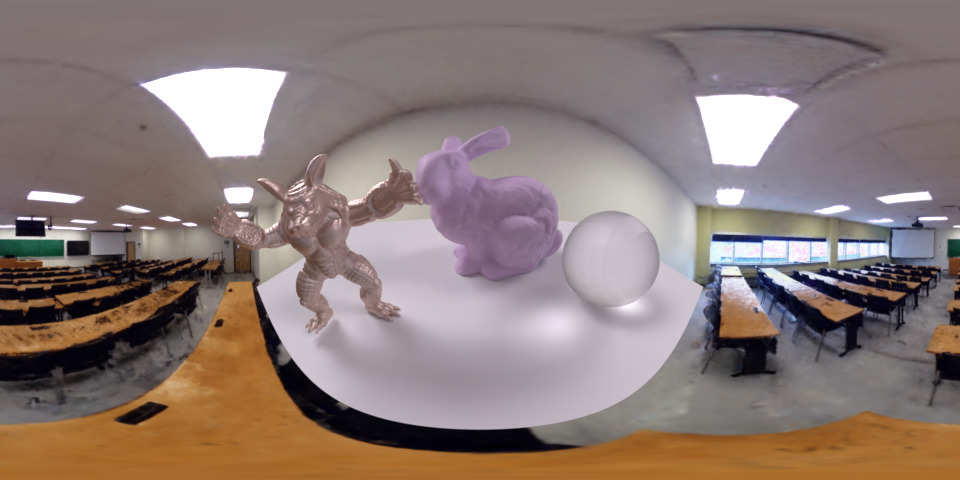} & 
     \includegraphics[width=\mywidth]{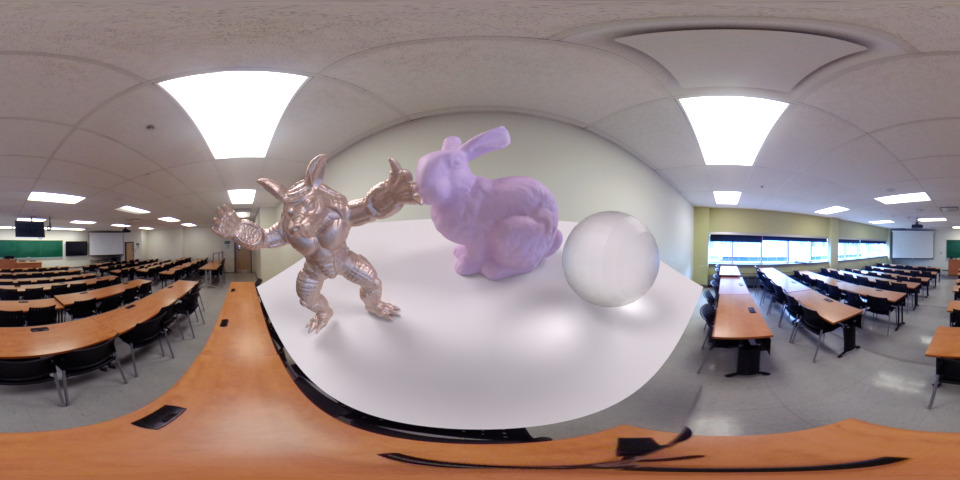} \\

     \includegraphics[width=\mywidth]{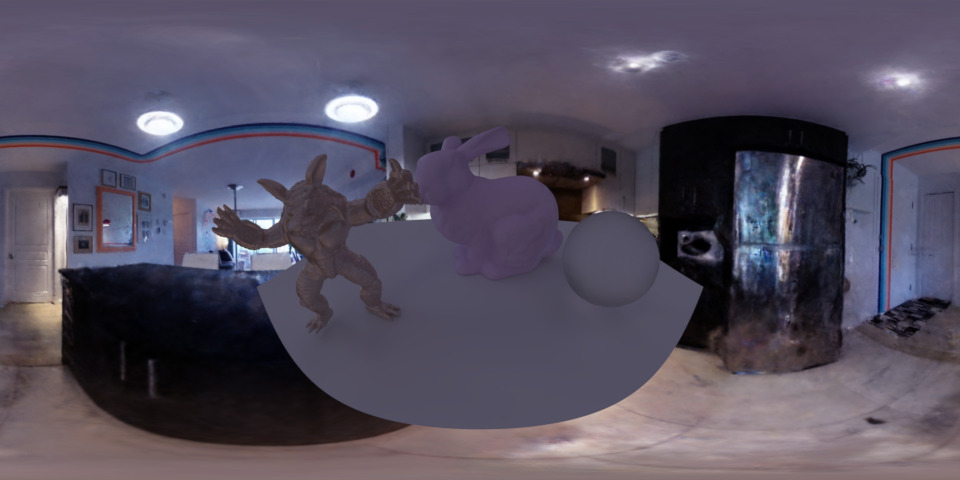} &
     \includegraphics[width=\mywidth]{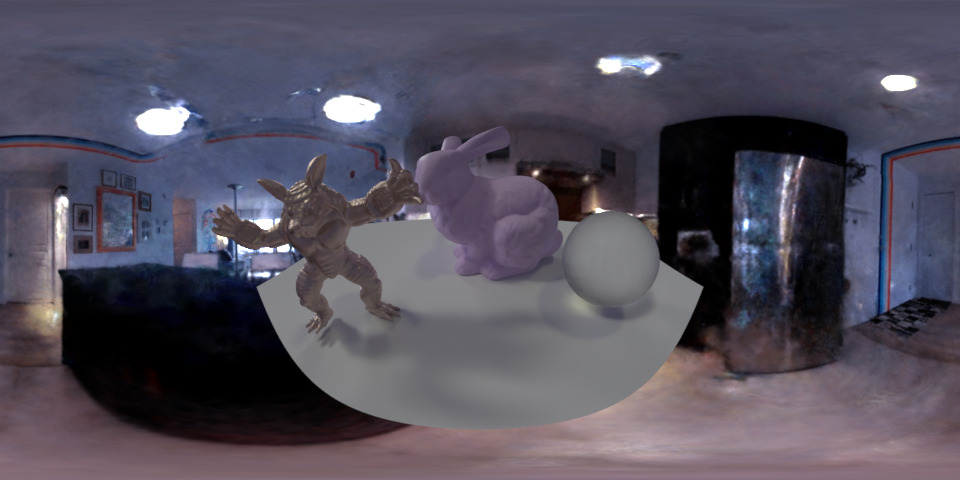} &
     \includegraphics[width=\mywidth]{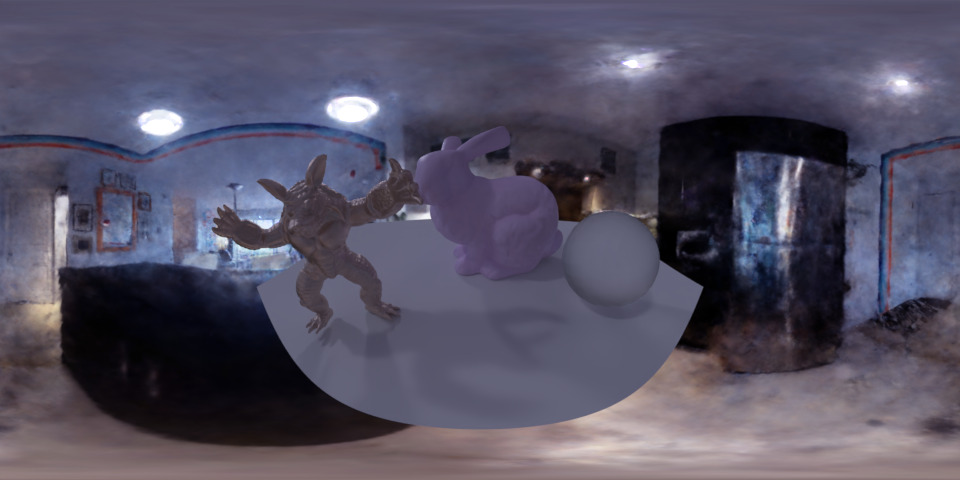} &
     \includegraphics[width=\mywidth]{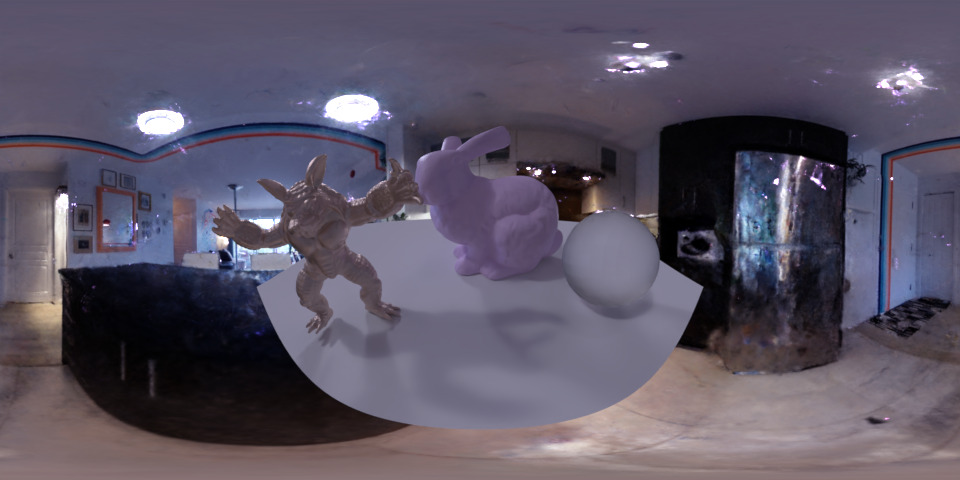} &
     \includegraphics[width=\mywidth]{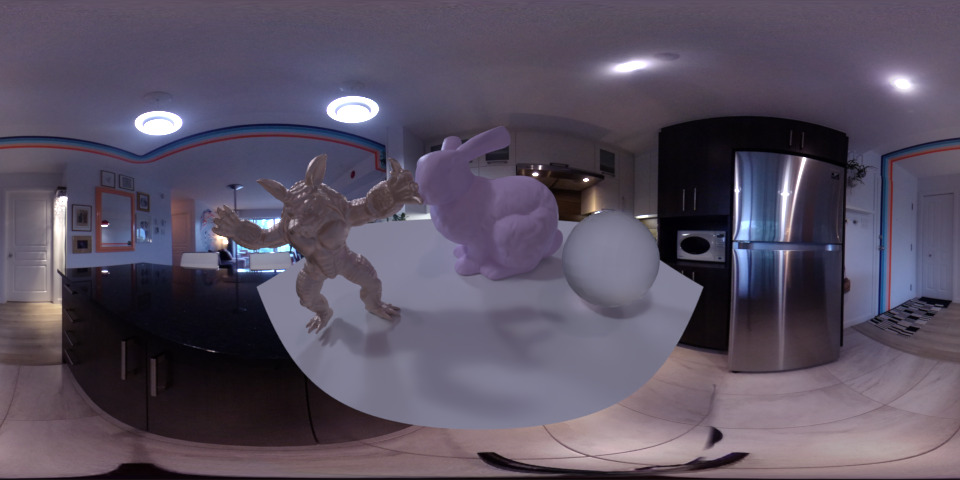} \\

     \includegraphics[width=\mywidth]{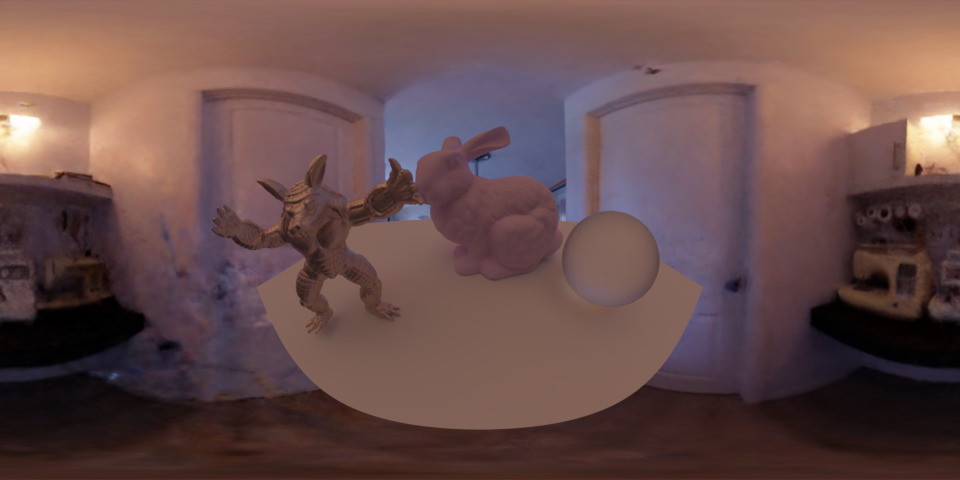} &
     \includegraphics[width=\mywidth]{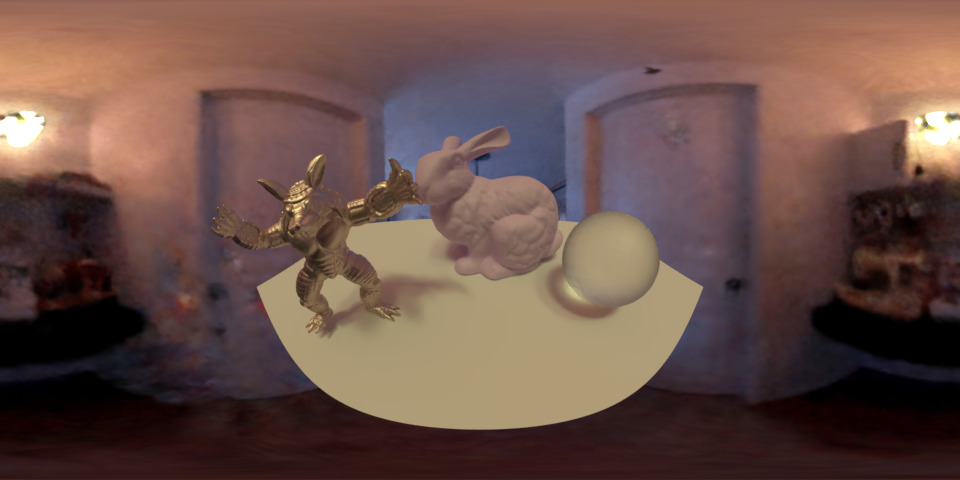} &
     \includegraphics[width=\mywidth]{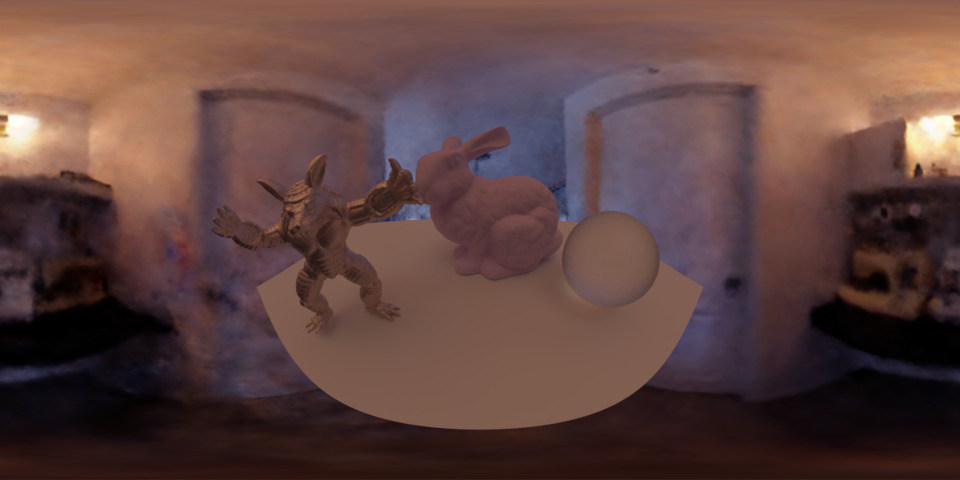} &
     \includegraphics[width=\mywidth]{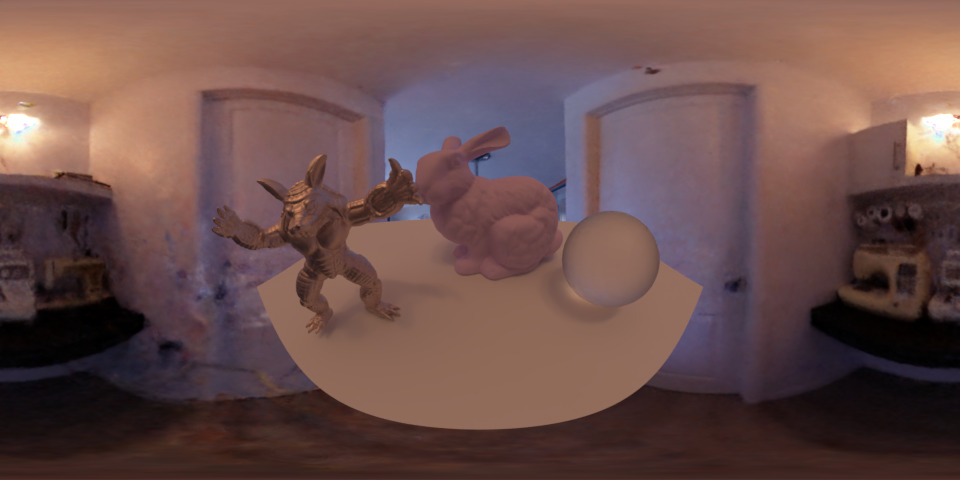} &
     \includegraphics[width=\mywidth]{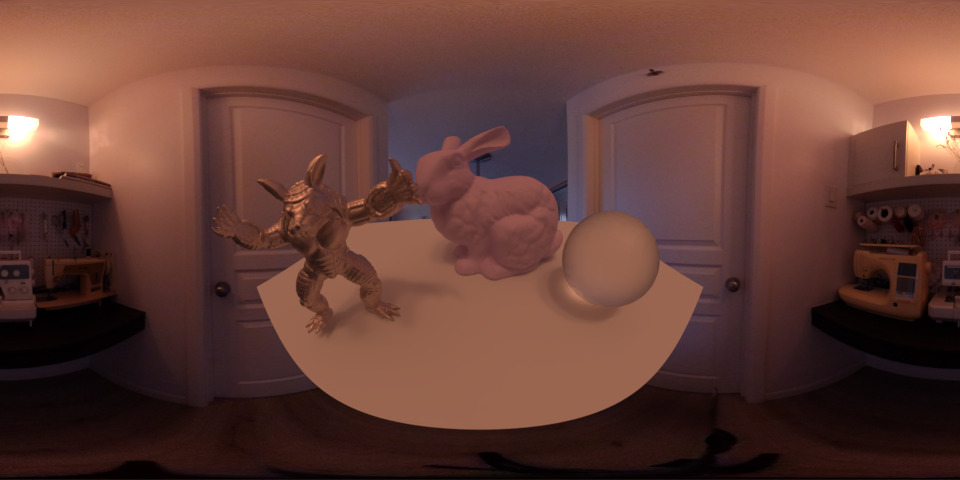} \\

     \includegraphics[width=\mywidth]{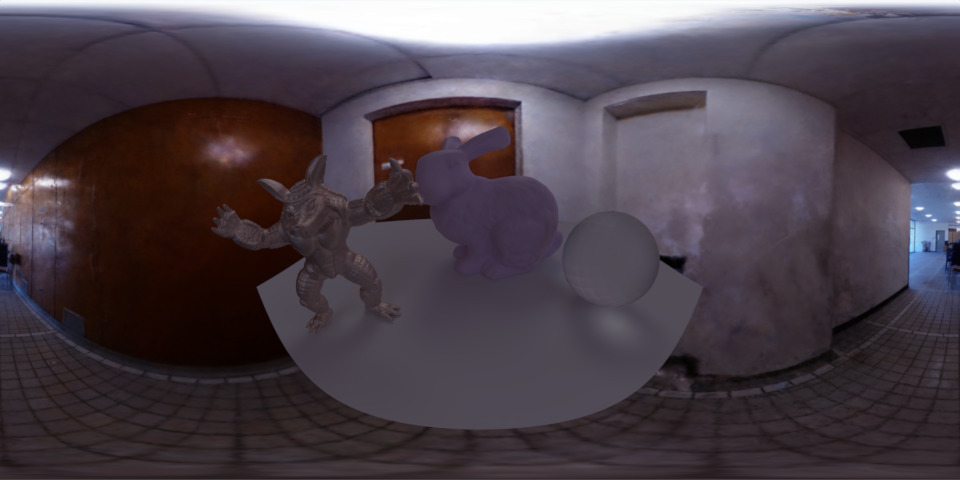} &
     \includegraphics[width=\mywidth]{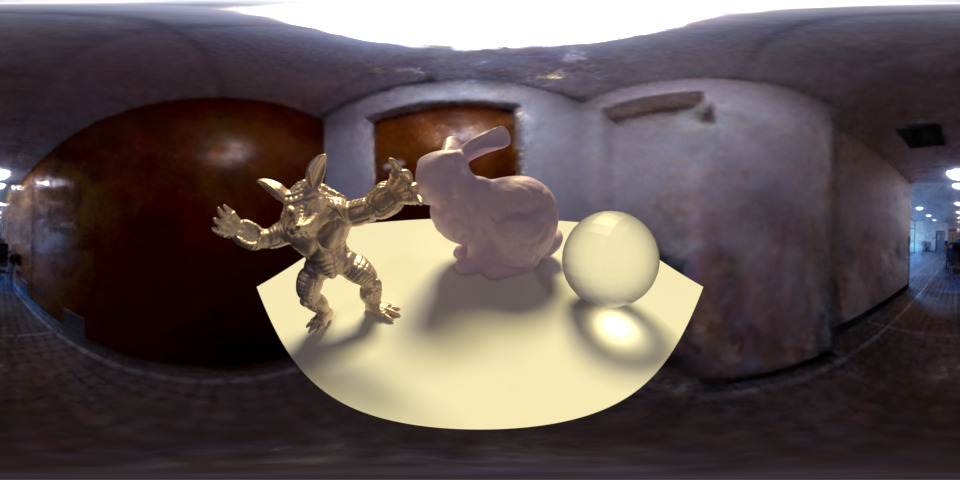} &
     \includegraphics[width=\mywidth]{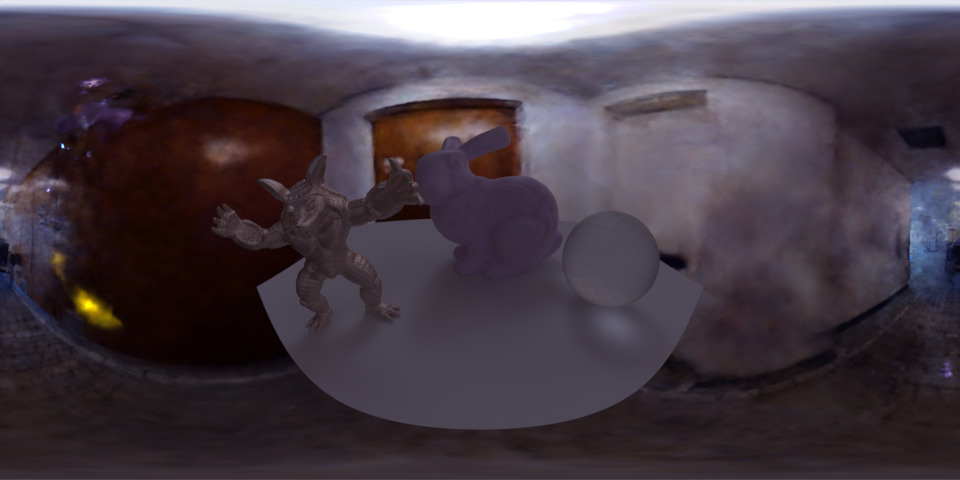} &
     \includegraphics[width=\mywidth]{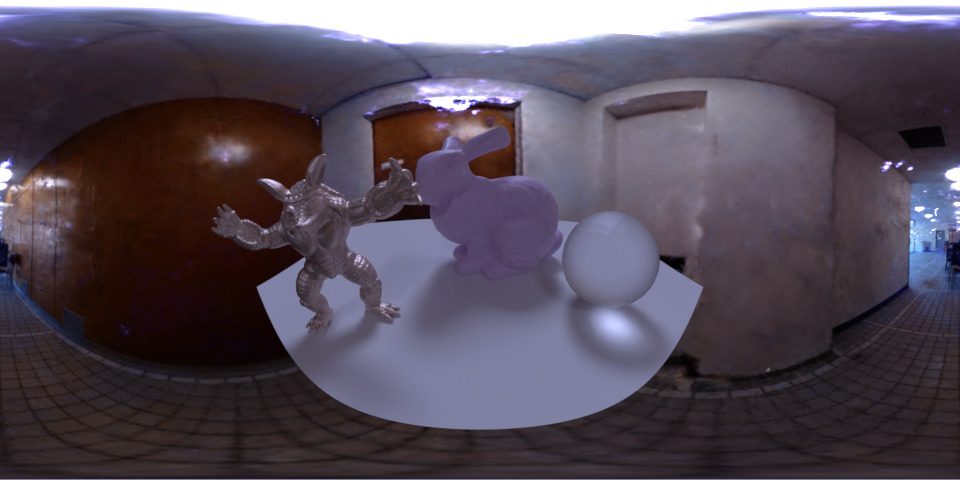} &
     \includegraphics[width=\mywidth]{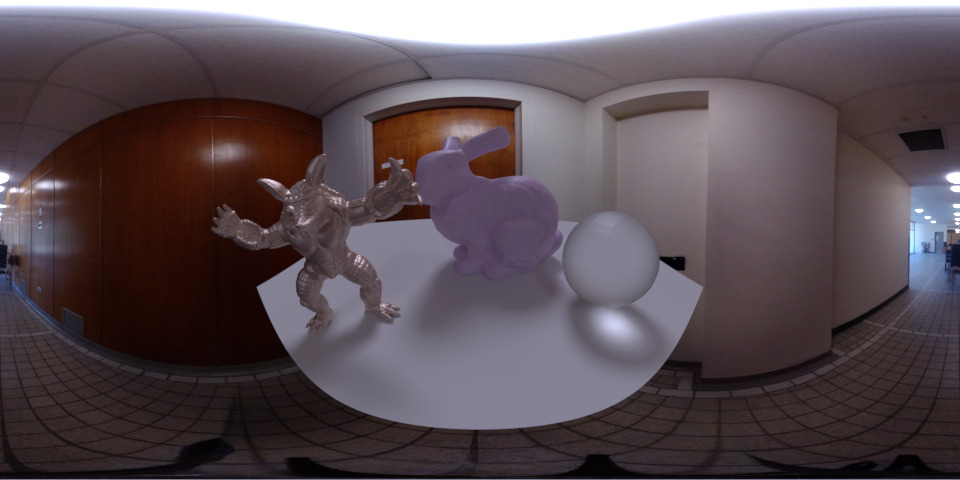} \\

     \includegraphics[width=\mywidth]{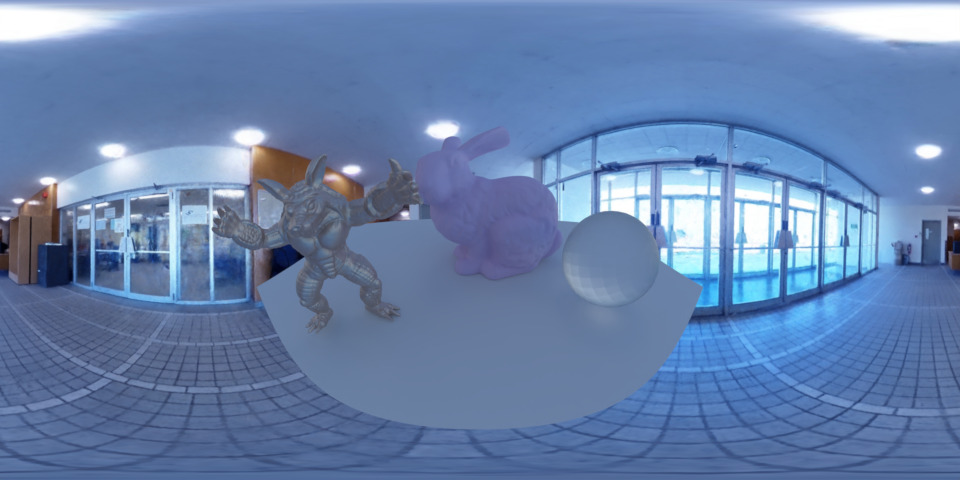} &
     \includegraphics[width=\mywidth]{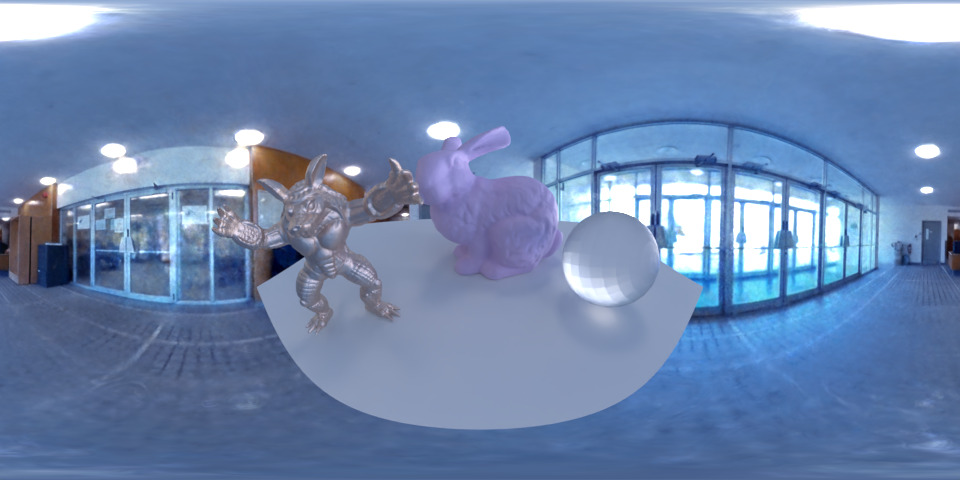} &
     \includegraphics[width=\mywidth]{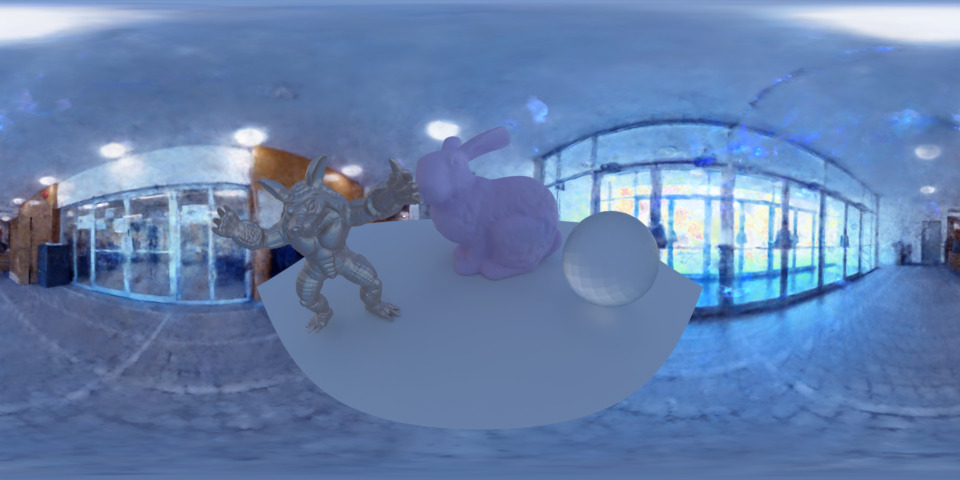} &
     \includegraphics[width=\mywidth]{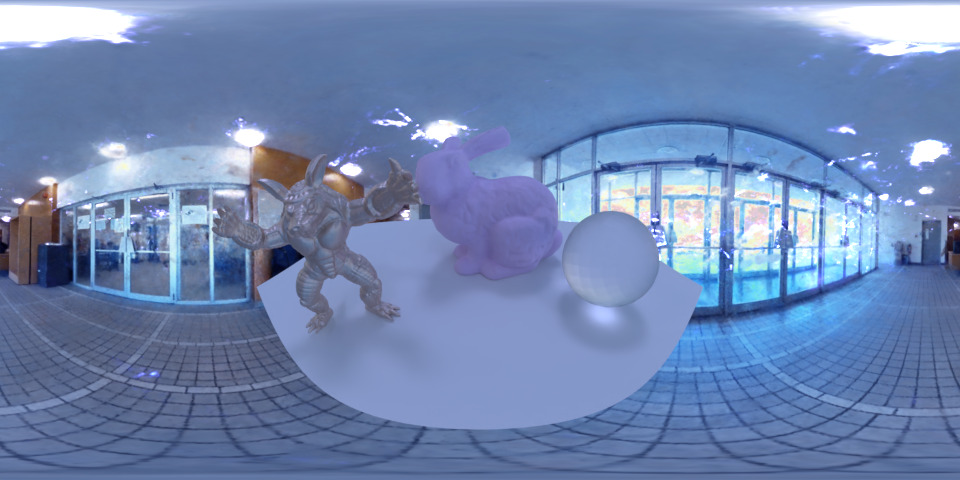} &
     \includegraphics[width=\mywidth]{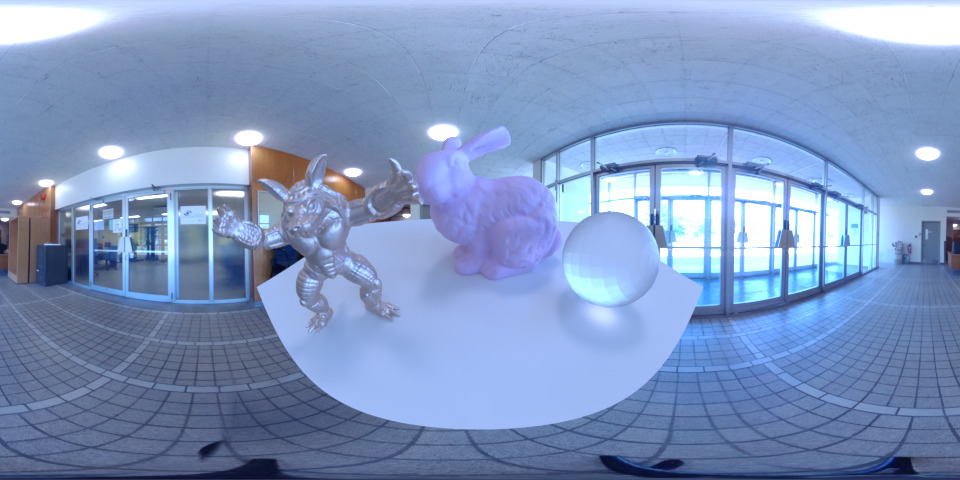} \\

     \includegraphics[width=\mywidth]{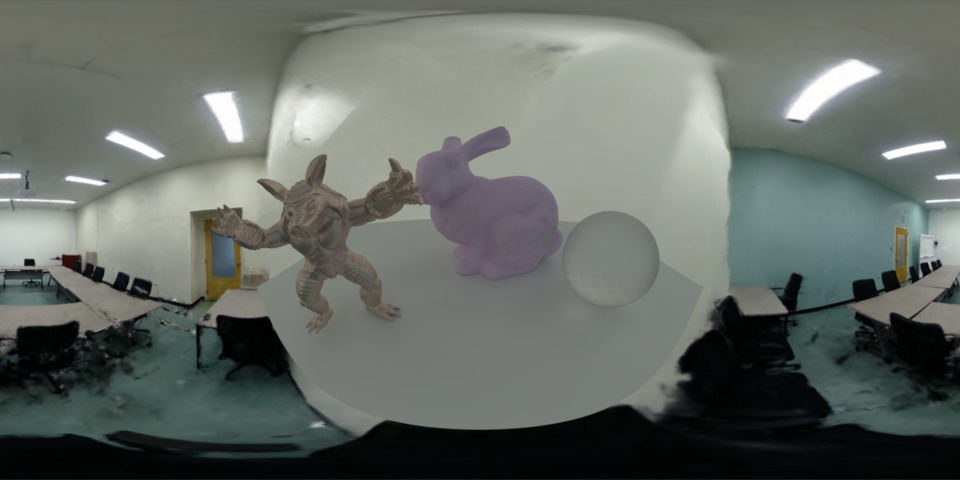} &
     \includegraphics[width=\mywidth]{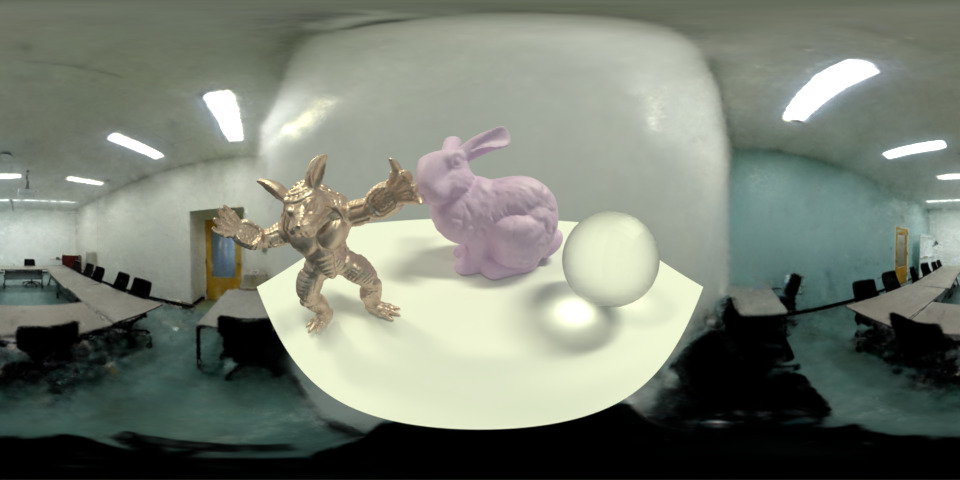} &
     \includegraphics[width=\mywidth]{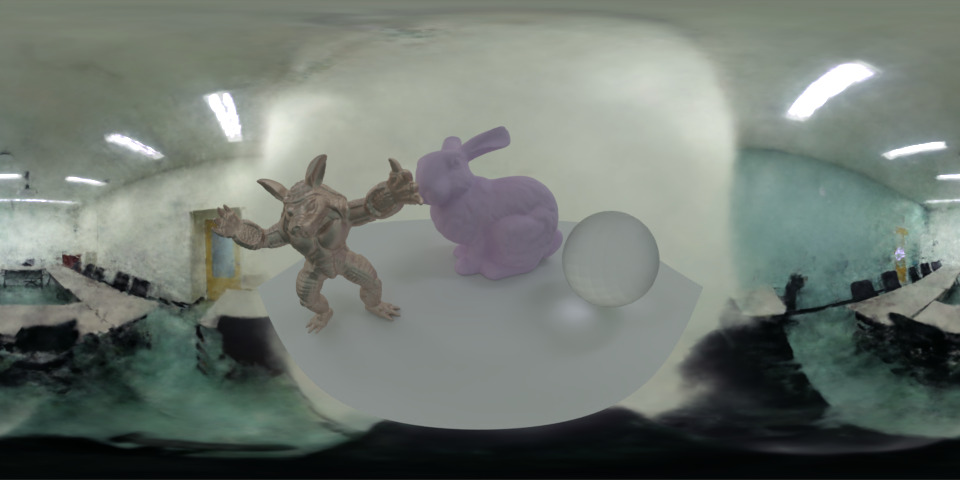} &
     \includegraphics[width=\mywidth]{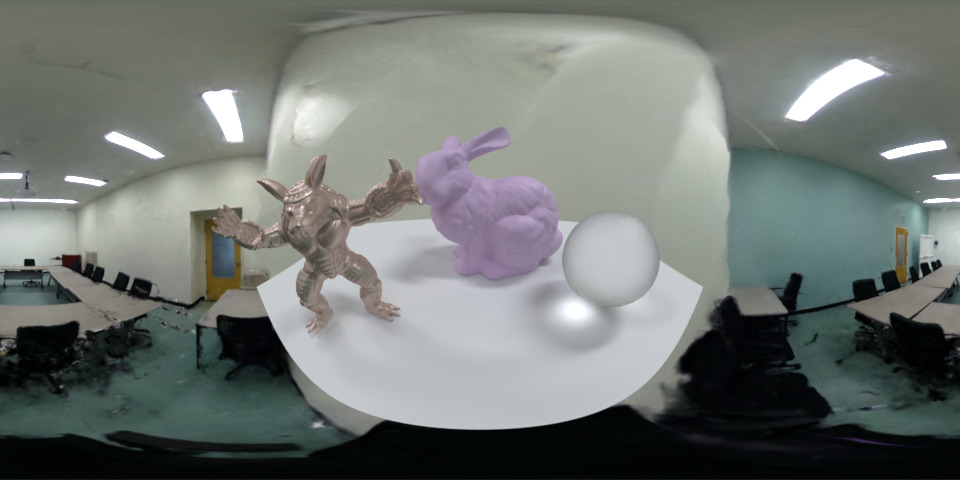} &
     \includegraphics[width=\mywidth]{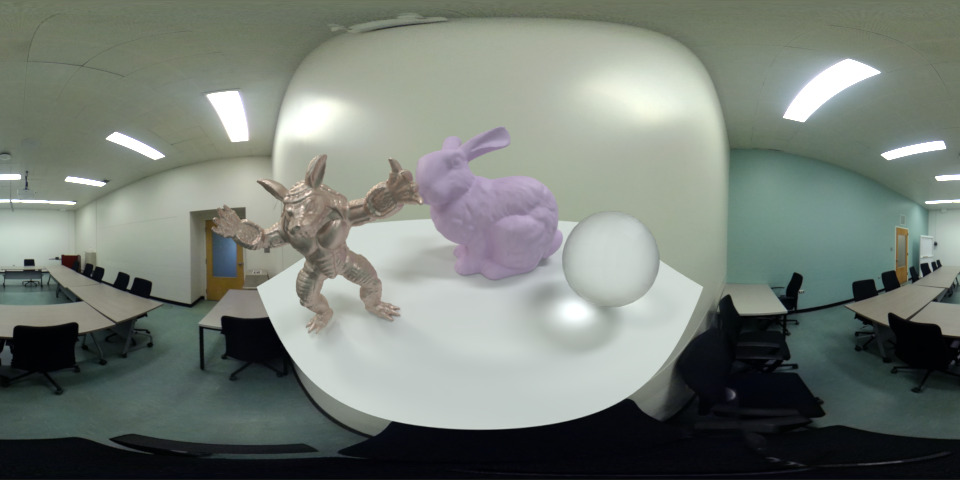} \\

     \includegraphics[width=\mywidth]{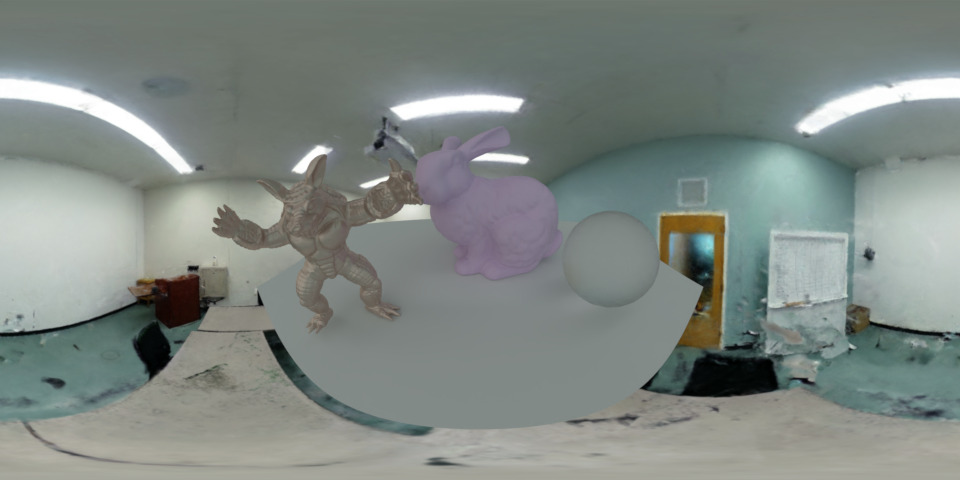} &
     \includegraphics[width=\mywidth]{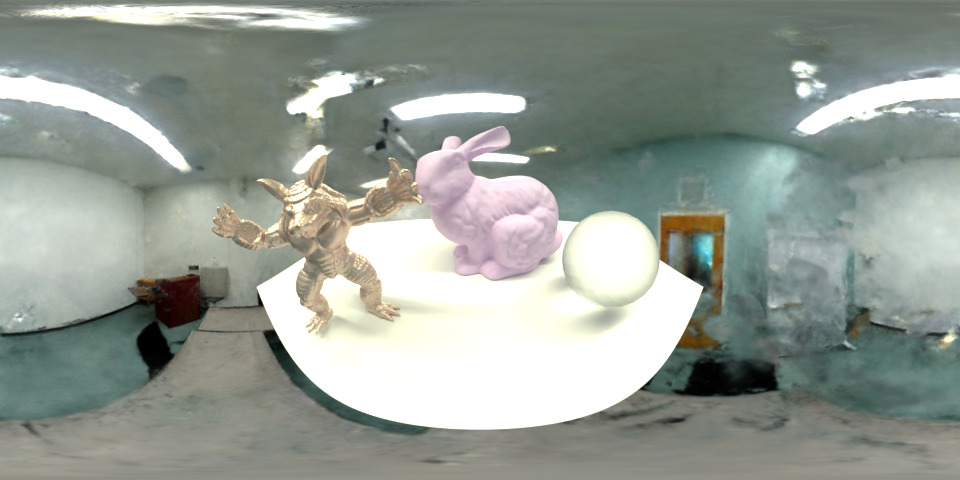} &
     \includegraphics[width=\mywidth]{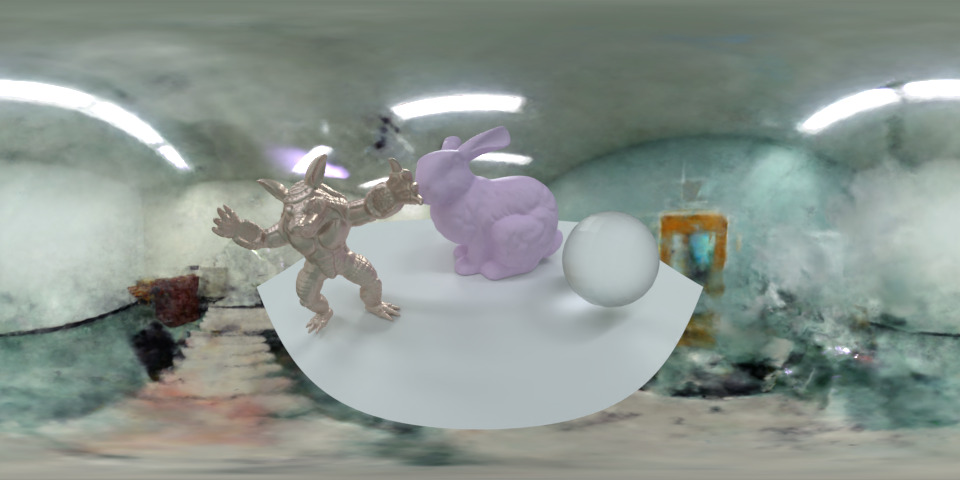} &
     \includegraphics[width=\mywidth]{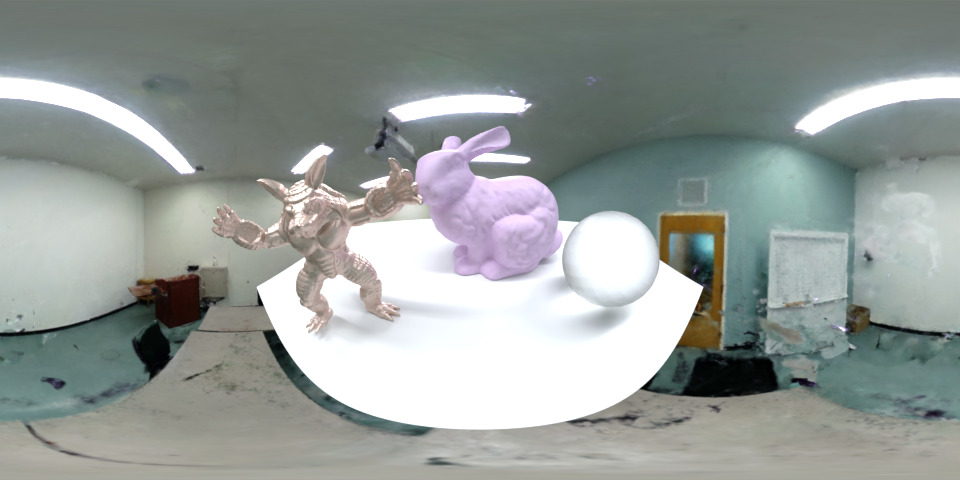} & 
     \includegraphics[width=\mywidth]{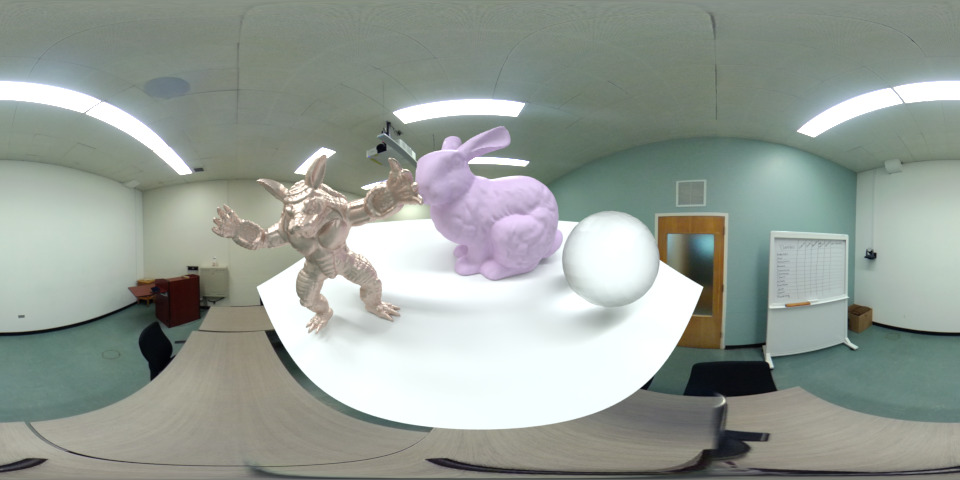} \\

     \includegraphics[width=\mywidth]{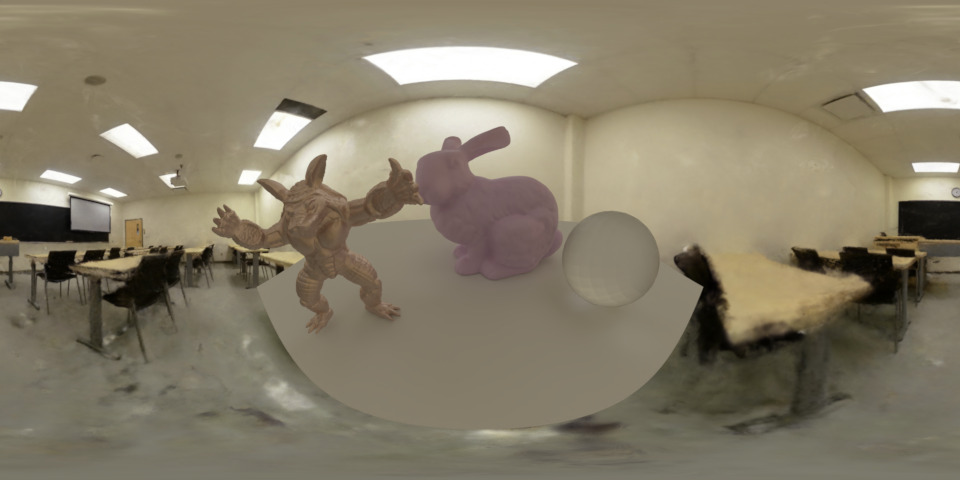} &
     \includegraphics[width=\mywidth]{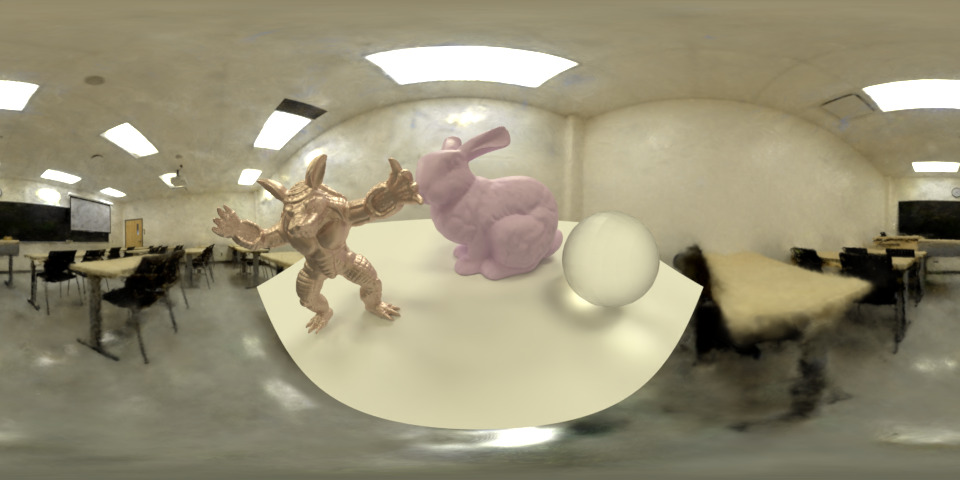} &
     \includegraphics[width=\mywidth]{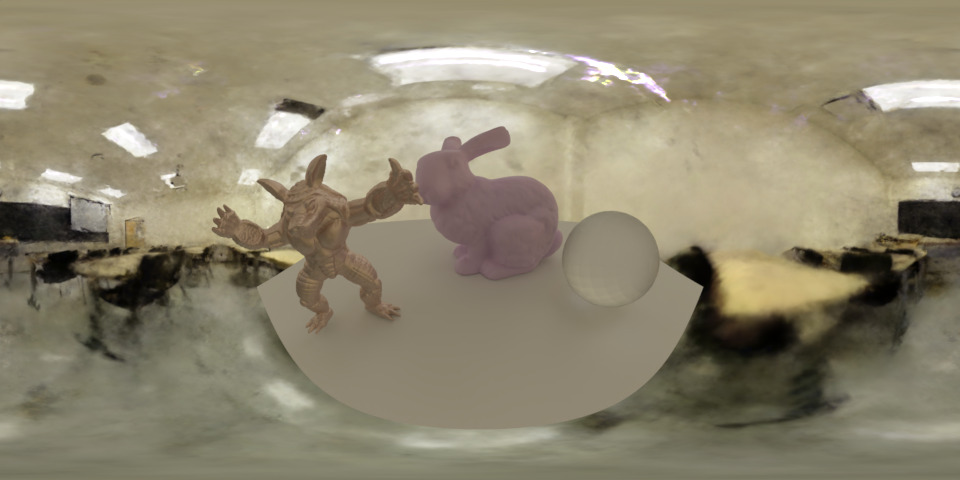} &
     \includegraphics[width=\mywidth]{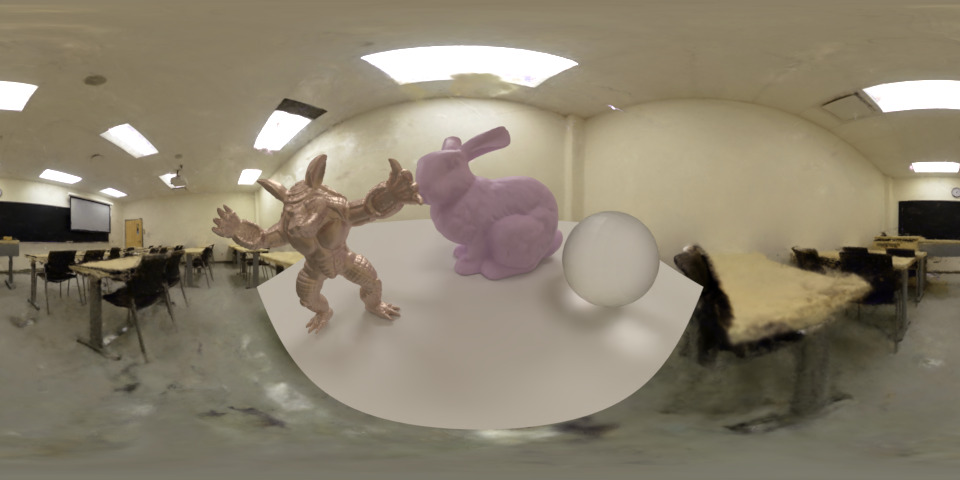} & 
     \includegraphics[width=\mywidth]{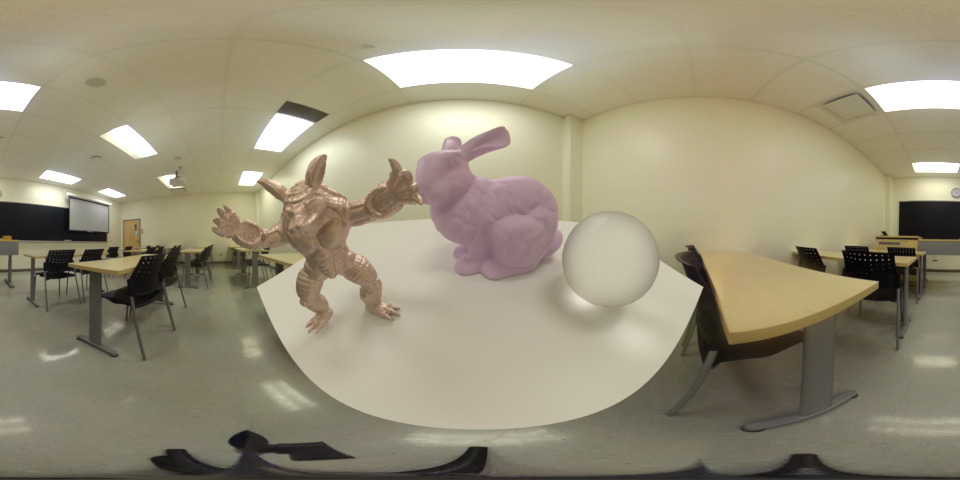} \\

     \includegraphics[width=\mywidth]{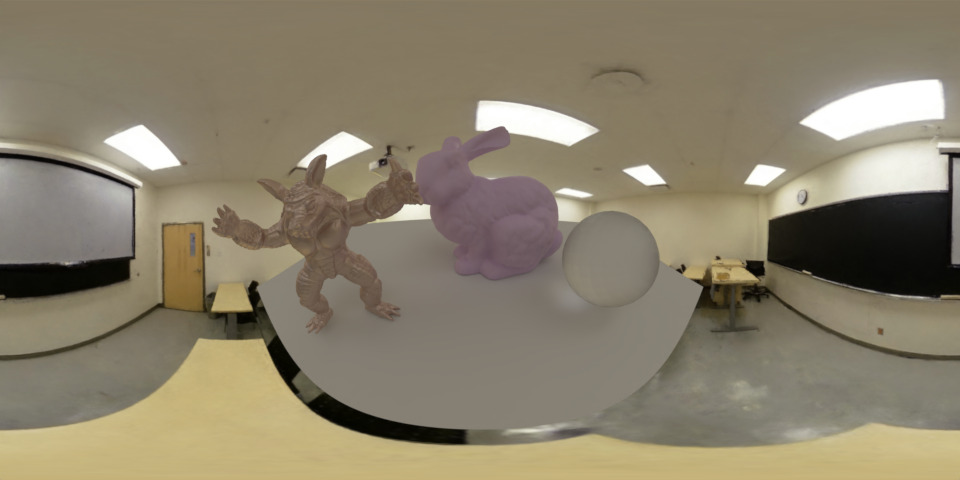} &
     \includegraphics[width=\mywidth]{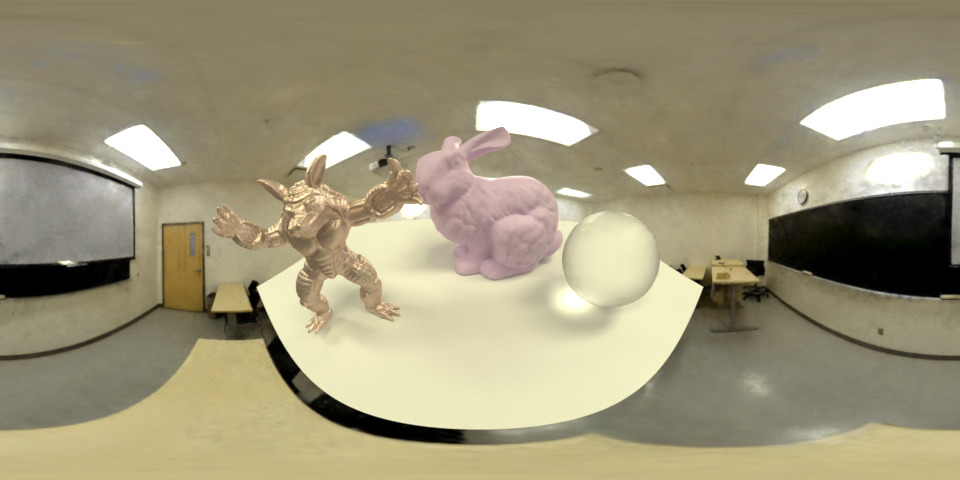} &
     \includegraphics[width=\mywidth]{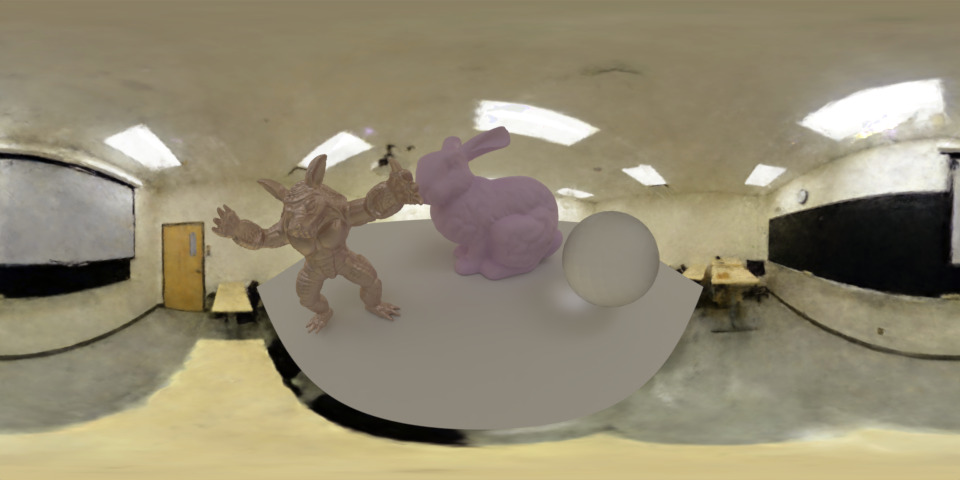} &
     \includegraphics[width=\mywidth]{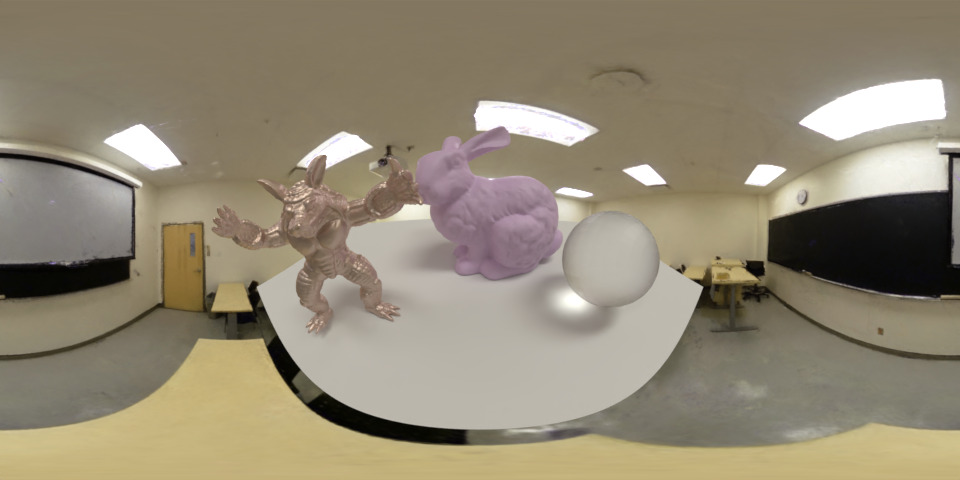} & 
     \includegraphics[width=\mywidth]{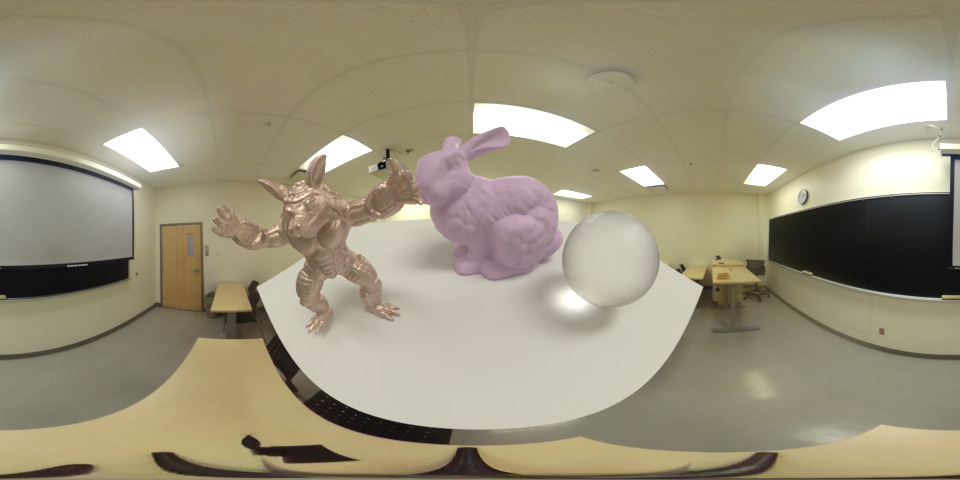} \\

     \includegraphics[width=\mywidth]{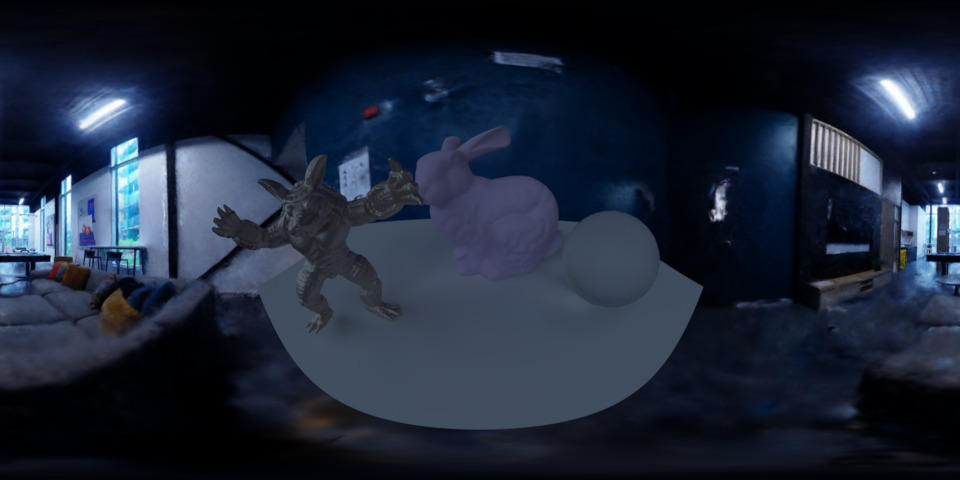} &
     \includegraphics[width=\mywidth]{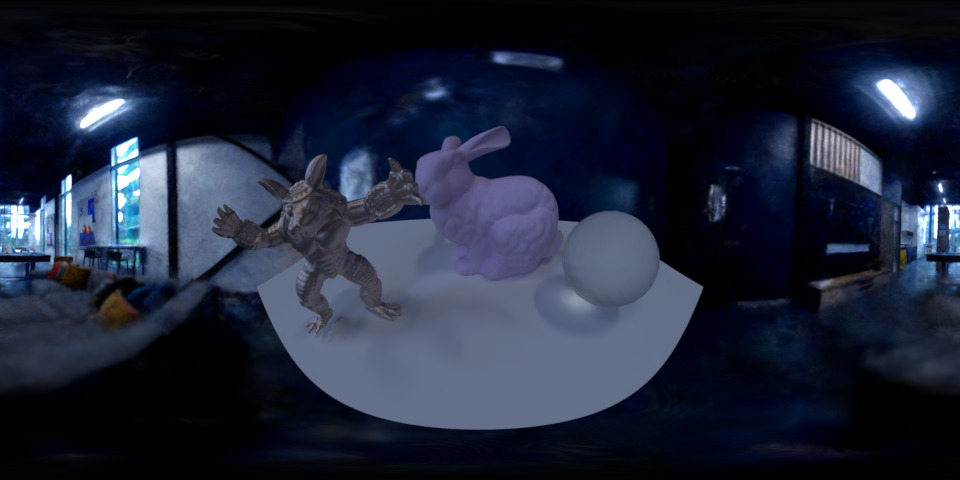} &
     \includegraphics[width=\mywidth]{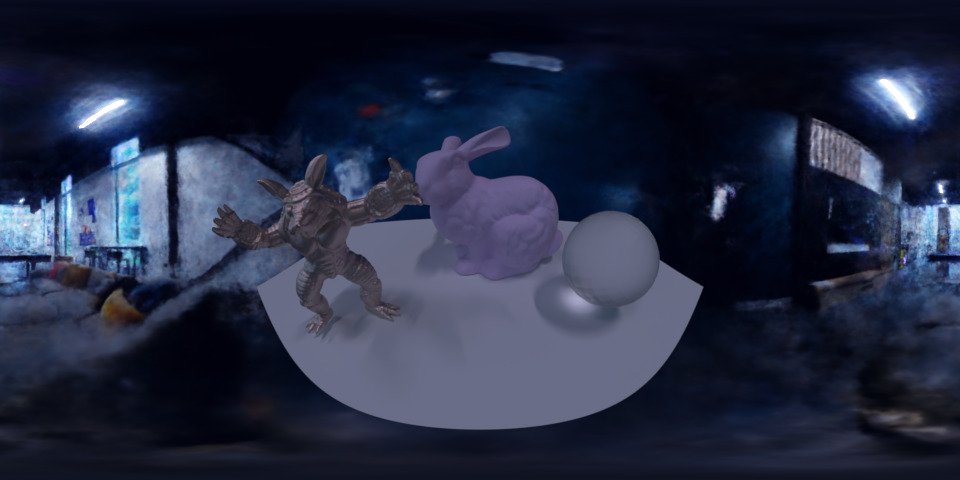} &
     \includegraphics[width=\mywidth]{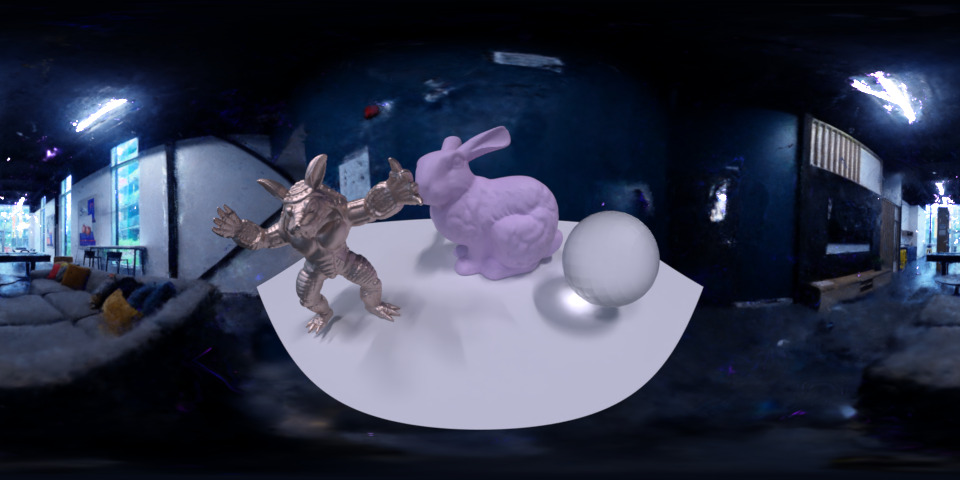} & 
     \includegraphics[width=\mywidth]{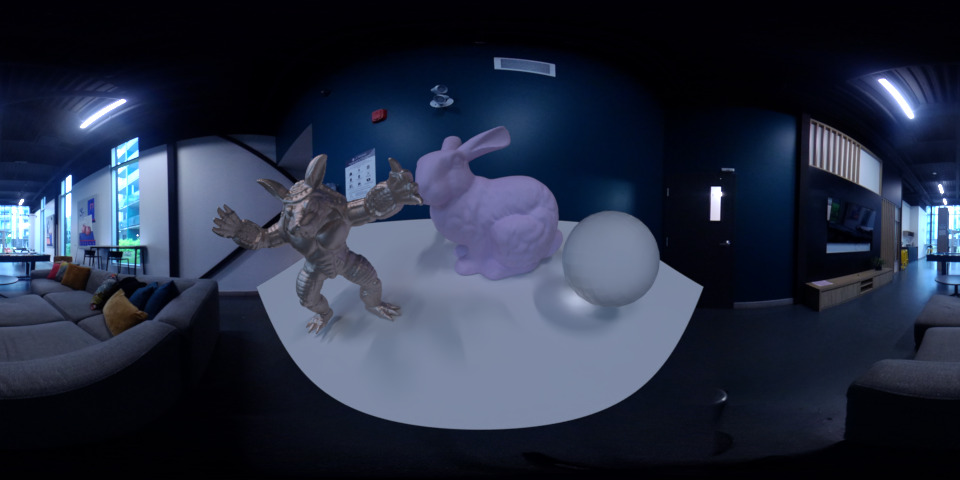} \\

     \includegraphics[width=\mywidth]{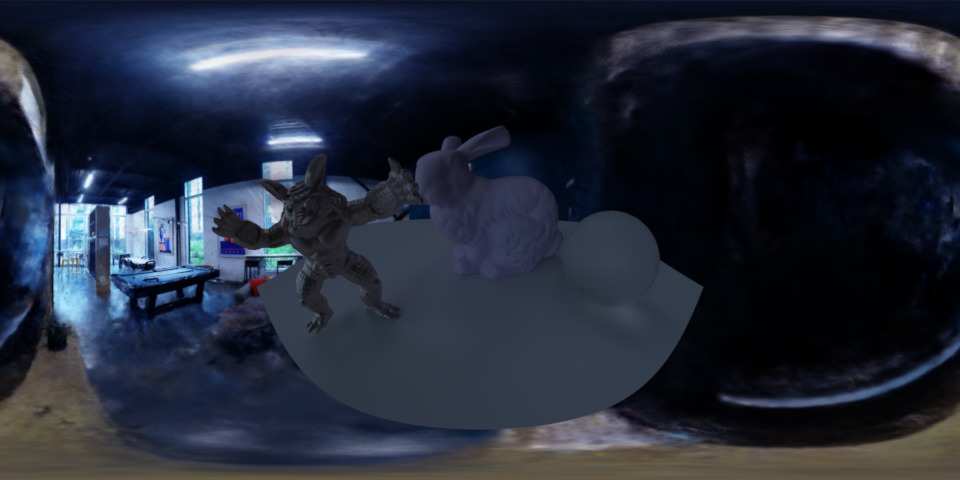} &
     \includegraphics[width=\mywidth]{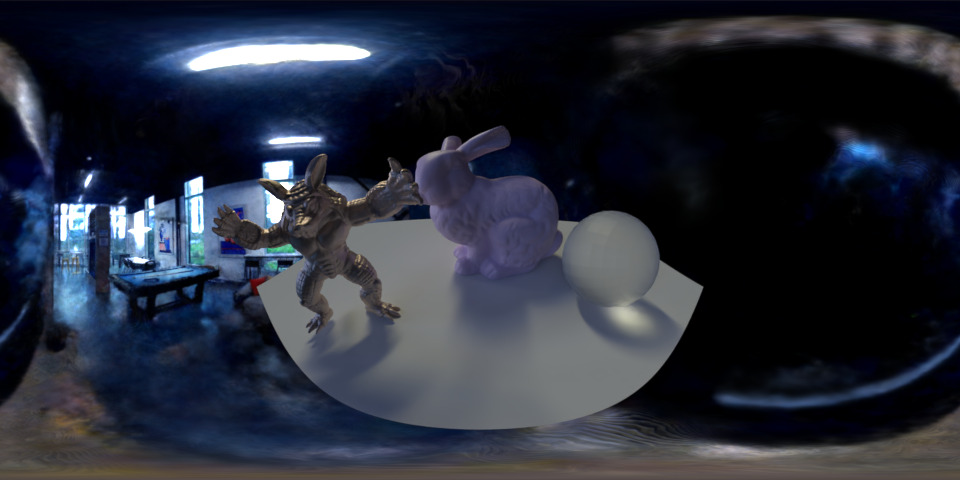} &
     \includegraphics[width=\mywidth]{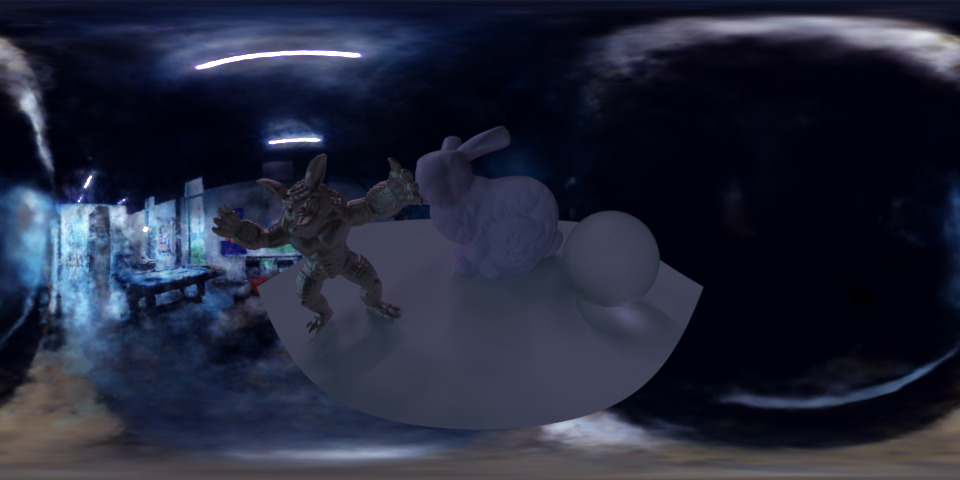} &
     \includegraphics[width=\mywidth]{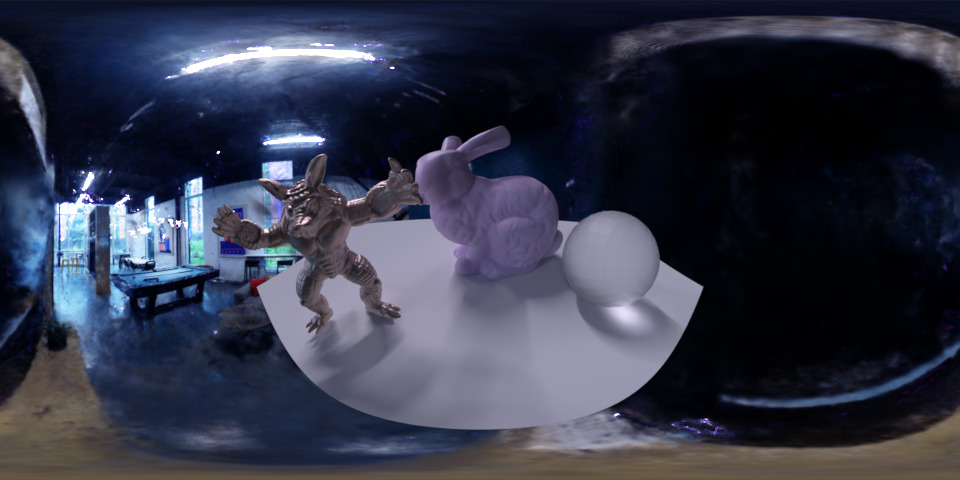} & 
     \includegraphics[width=\mywidth]{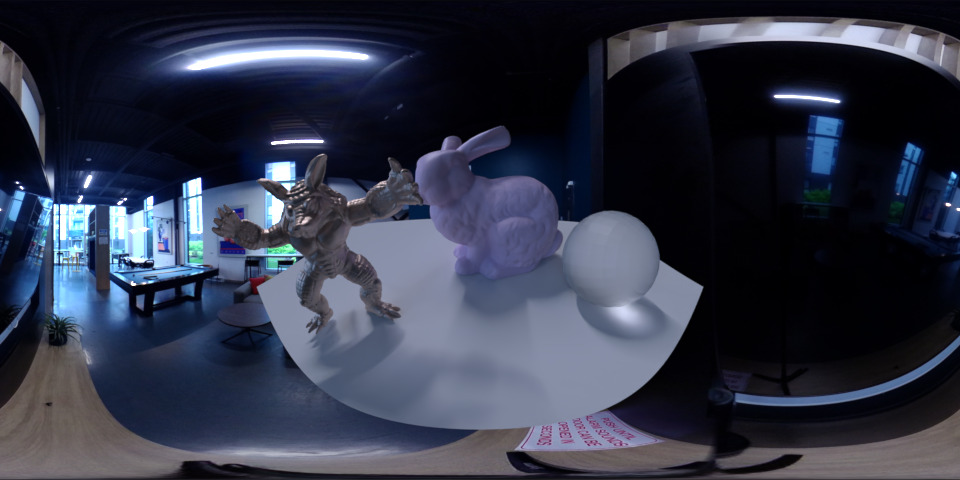}
    \end{tabular}
    \caption{Qualitative results on our captured scenes. Our method produces high-quality environment maps with a dynamic range closer to the ground truth, resulting in more realistic rendering results. From left to right, we compare: LDR-Nerfacto~\cite{nerfstudio}, PanoHDR-Nerfacto~\cite{gera2022casual}, HDR-Nerfacto~\cite{huang2022hdrnerf}, \thename (ours) and the ground truth.}
    \label{fig:qual}
\end{figure*}

%% file: figures/fast_exposed_results.tex
\begin{figure*} [t!]
\centering
\small
\setlength{\tabcolsep}{1pt}
\setlength{\alignwidth}{0.2\linewidth}
\begin{tabular}{cccc}
GT (well-exposed) & GT (fast-exposed) & Ours (\thename) & HDR-Nerfacto\\
\includegraphics[width=\alignwidth]{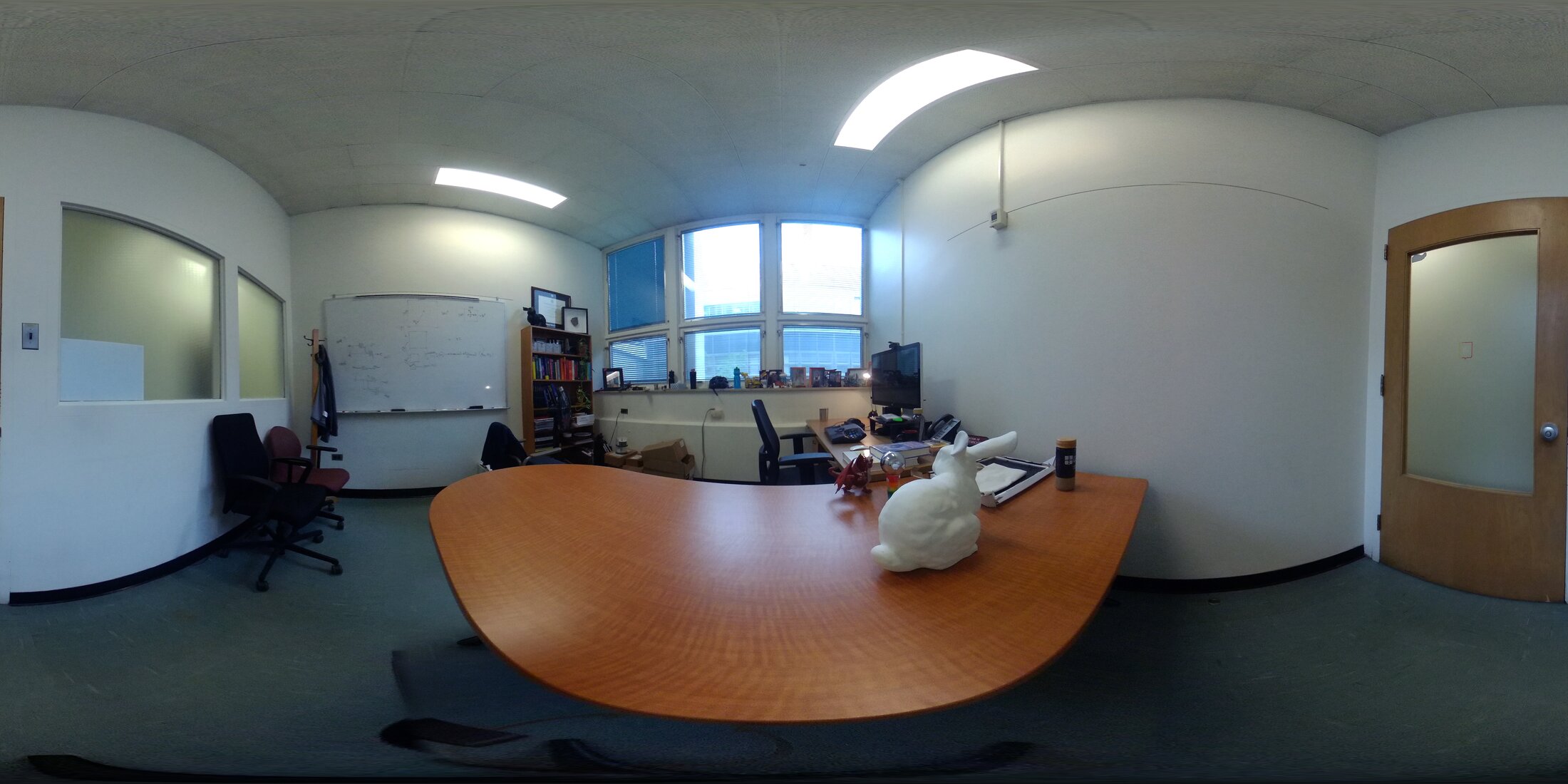} & 
\includegraphics[width=\alignwidth]{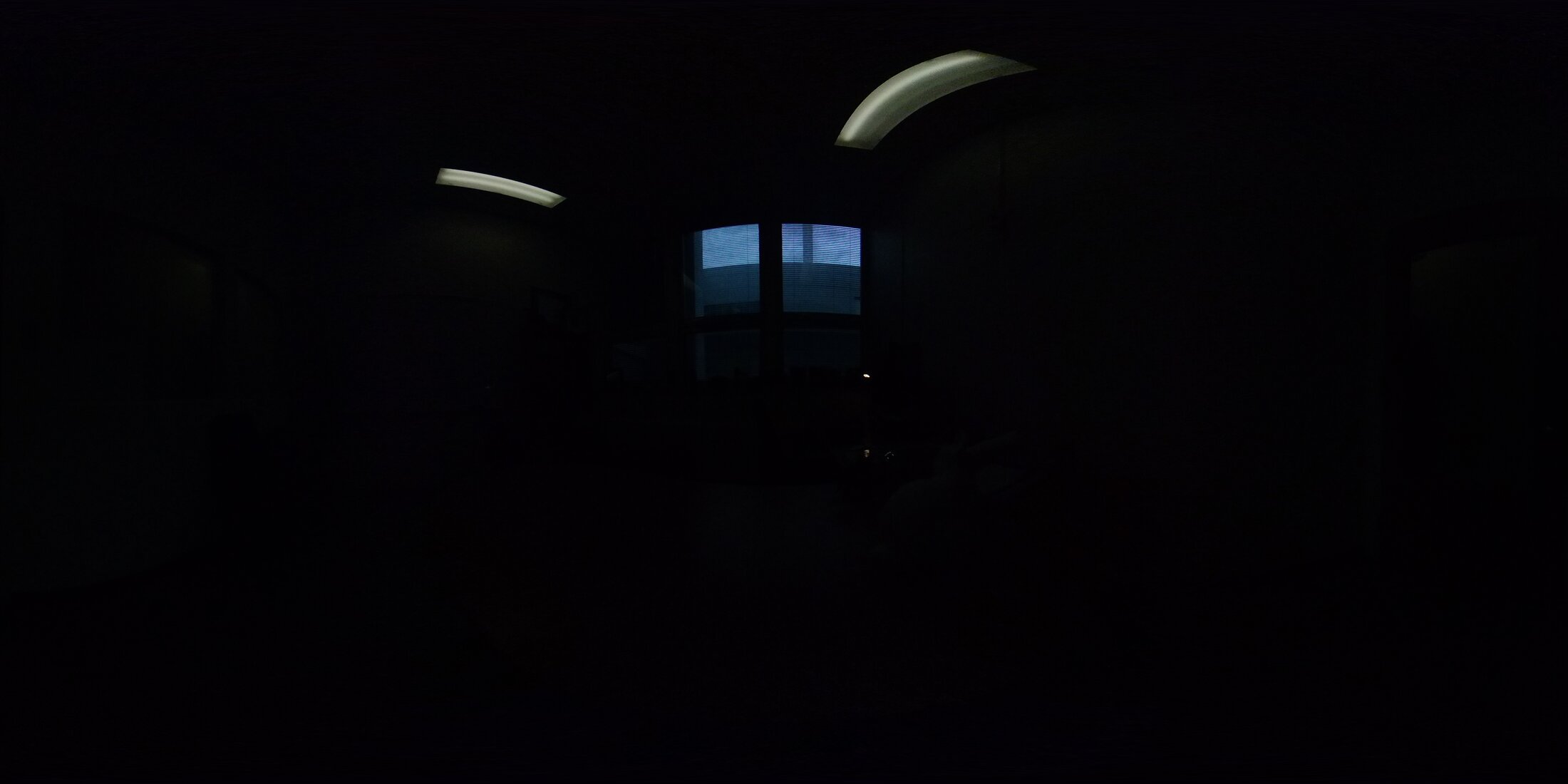} & 
\includegraphics[width=\alignwidth]{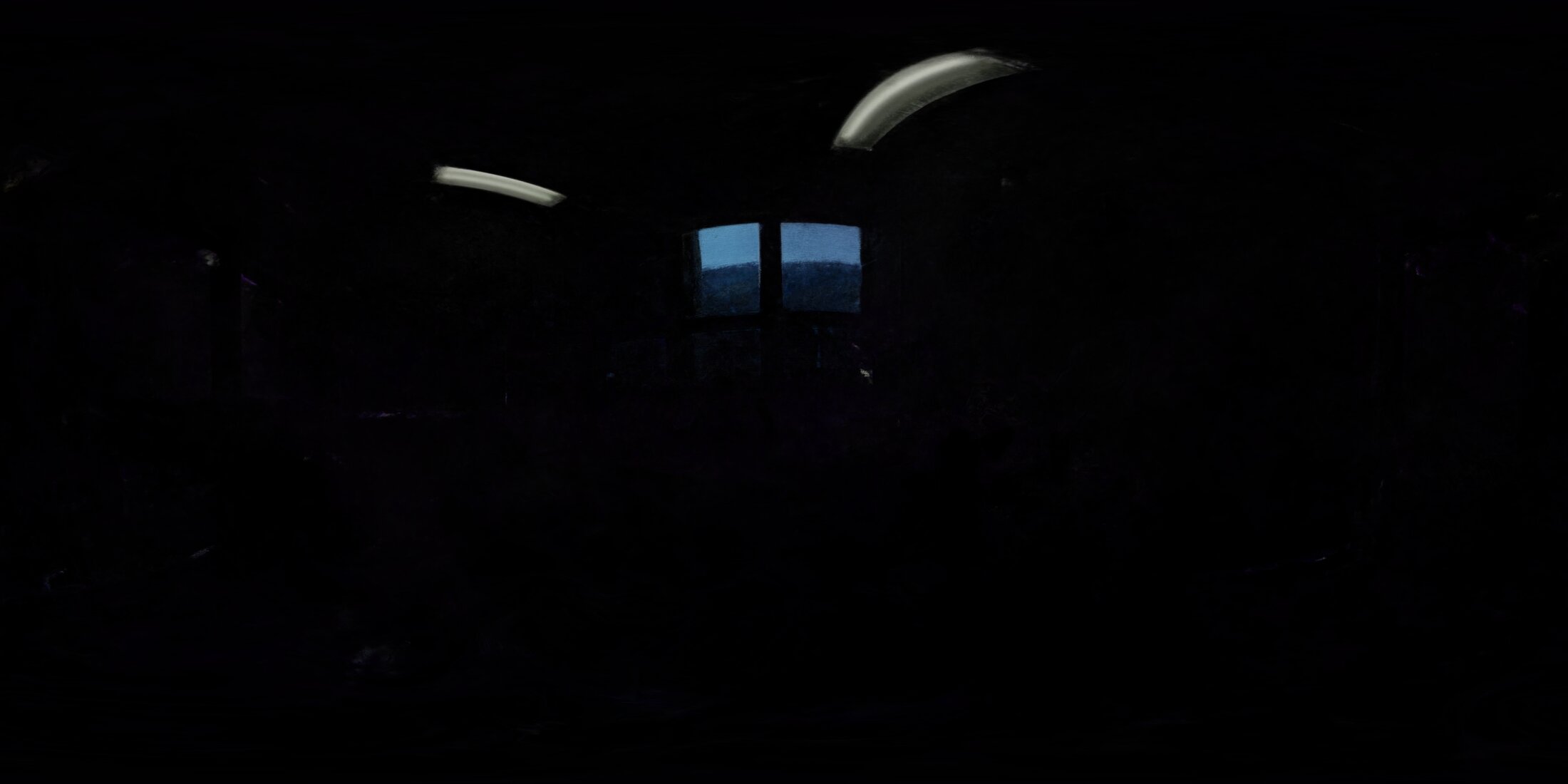}&
\includegraphics[width=\alignwidth]{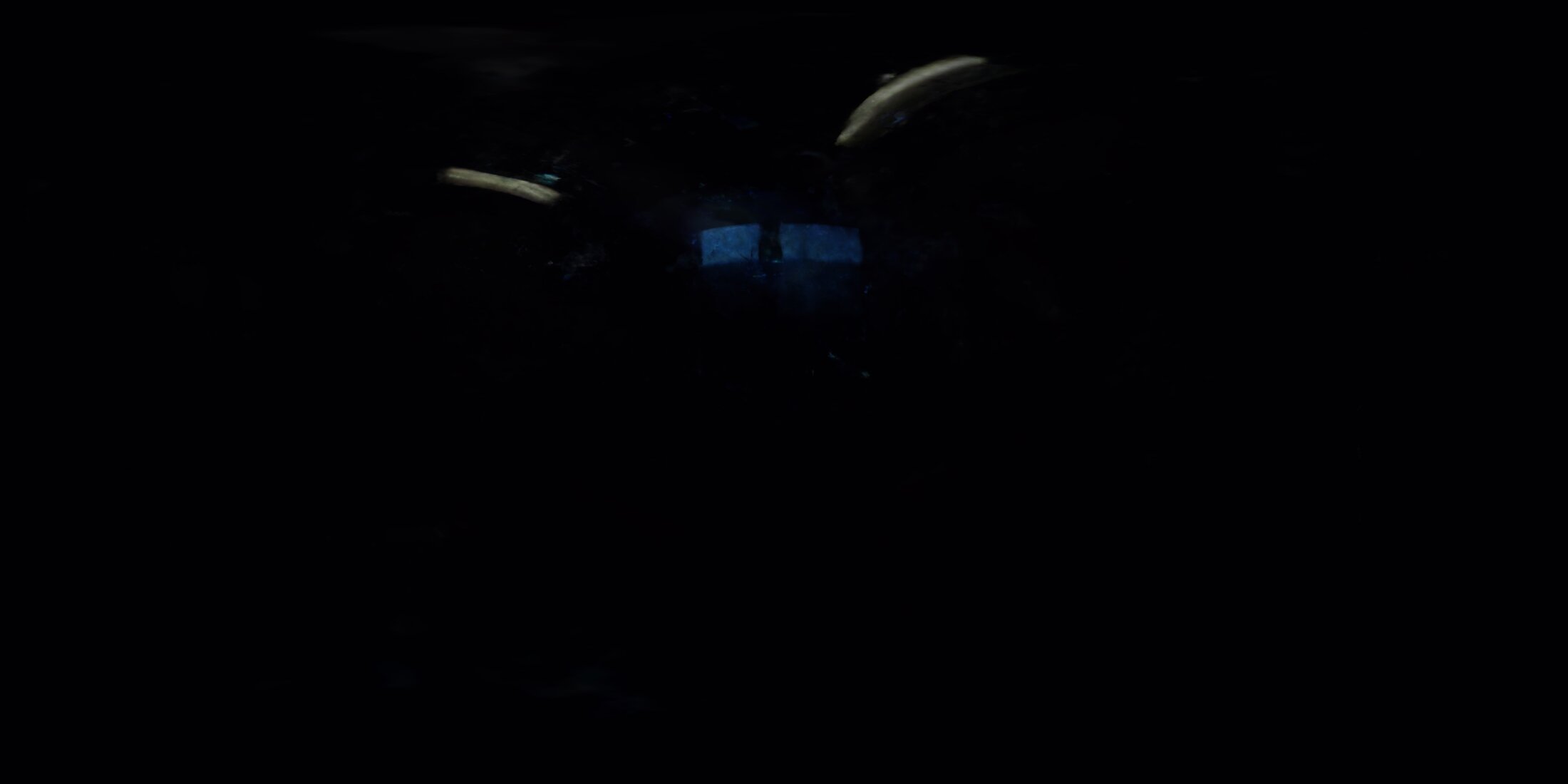} \\
\includegraphics[width=\alignwidth]{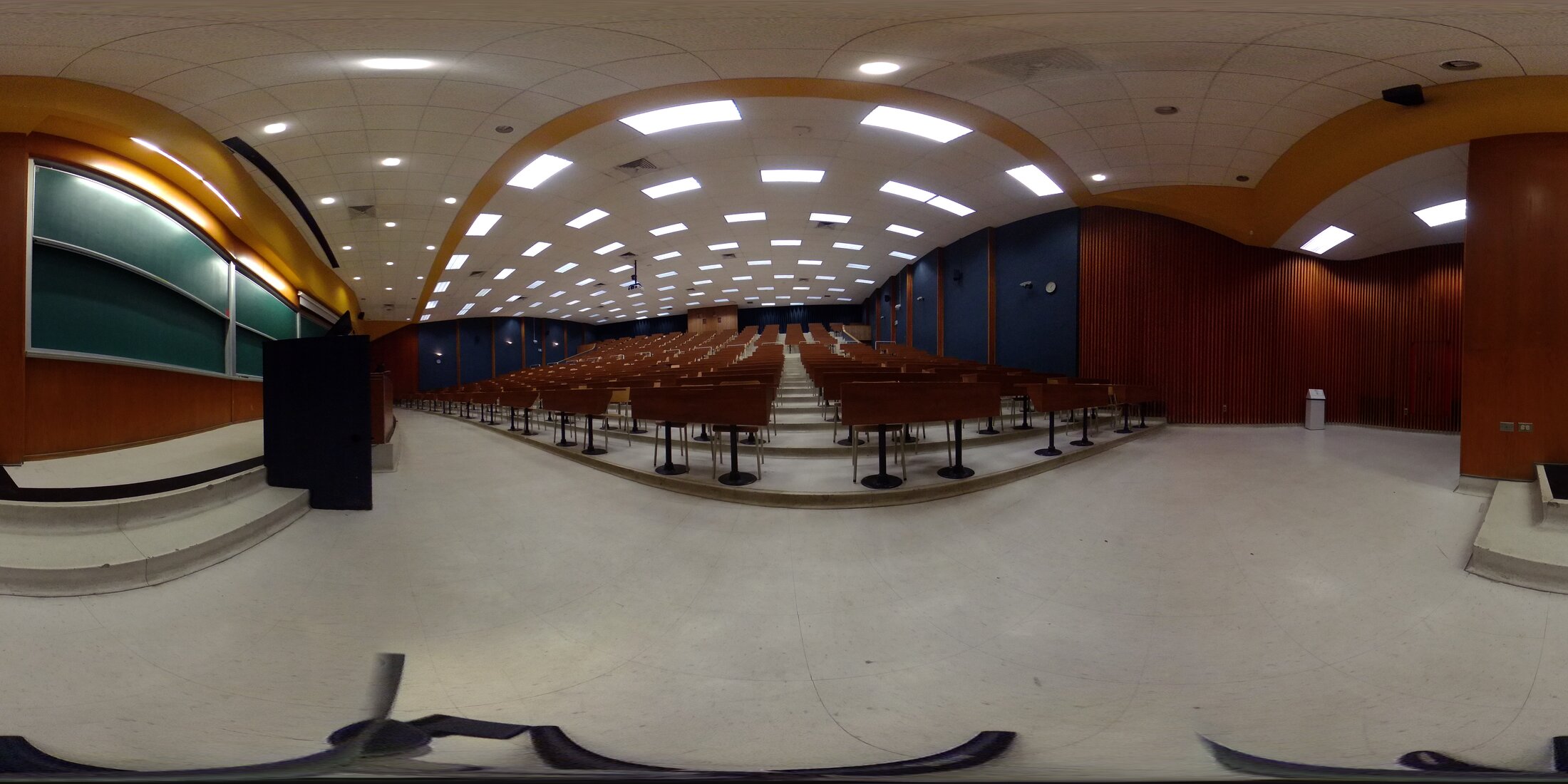} & \includegraphics[width=\alignwidth]{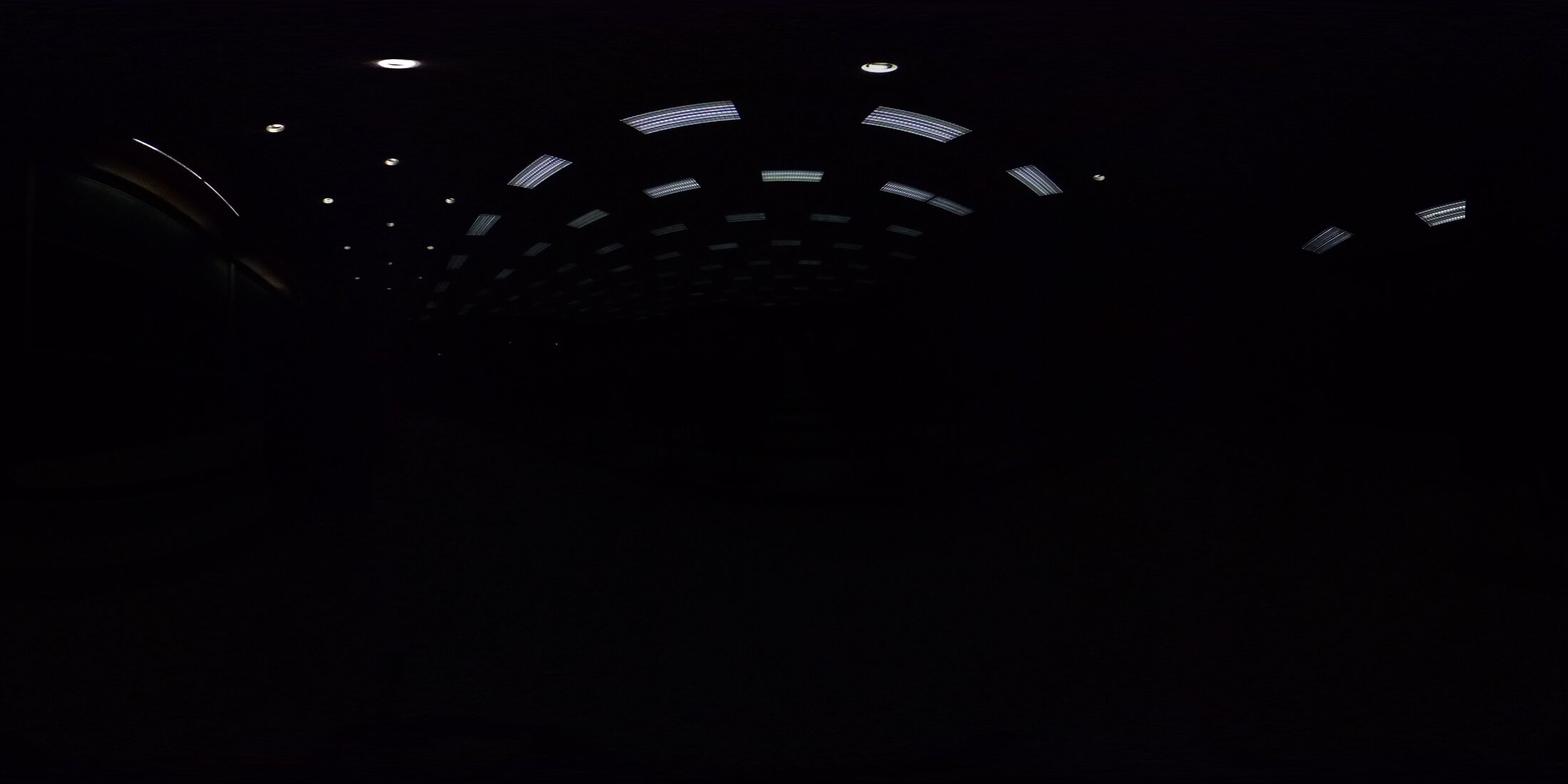} & 
\includegraphics[width=\alignwidth]{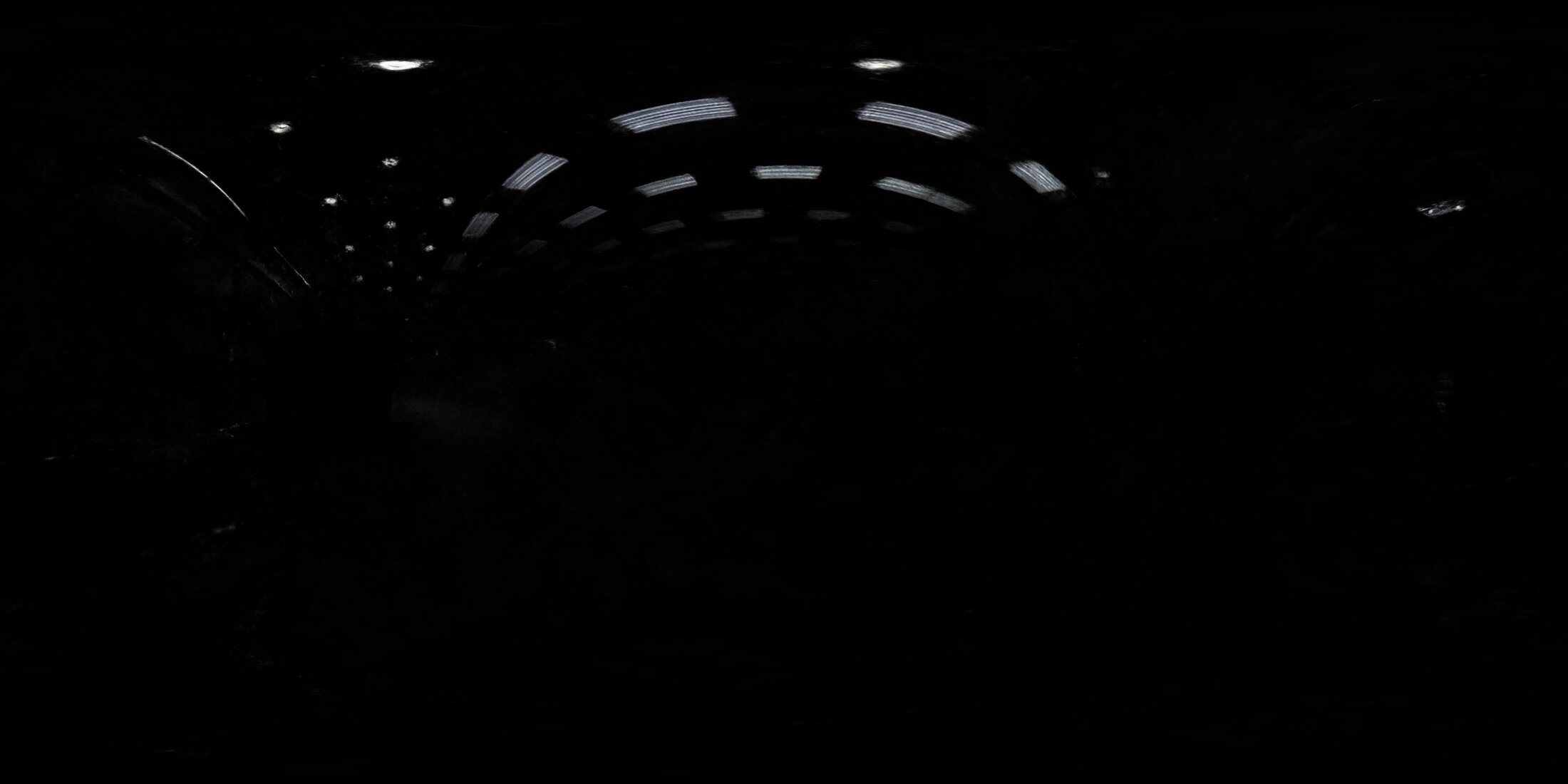} &
\includegraphics[width=\alignwidth]{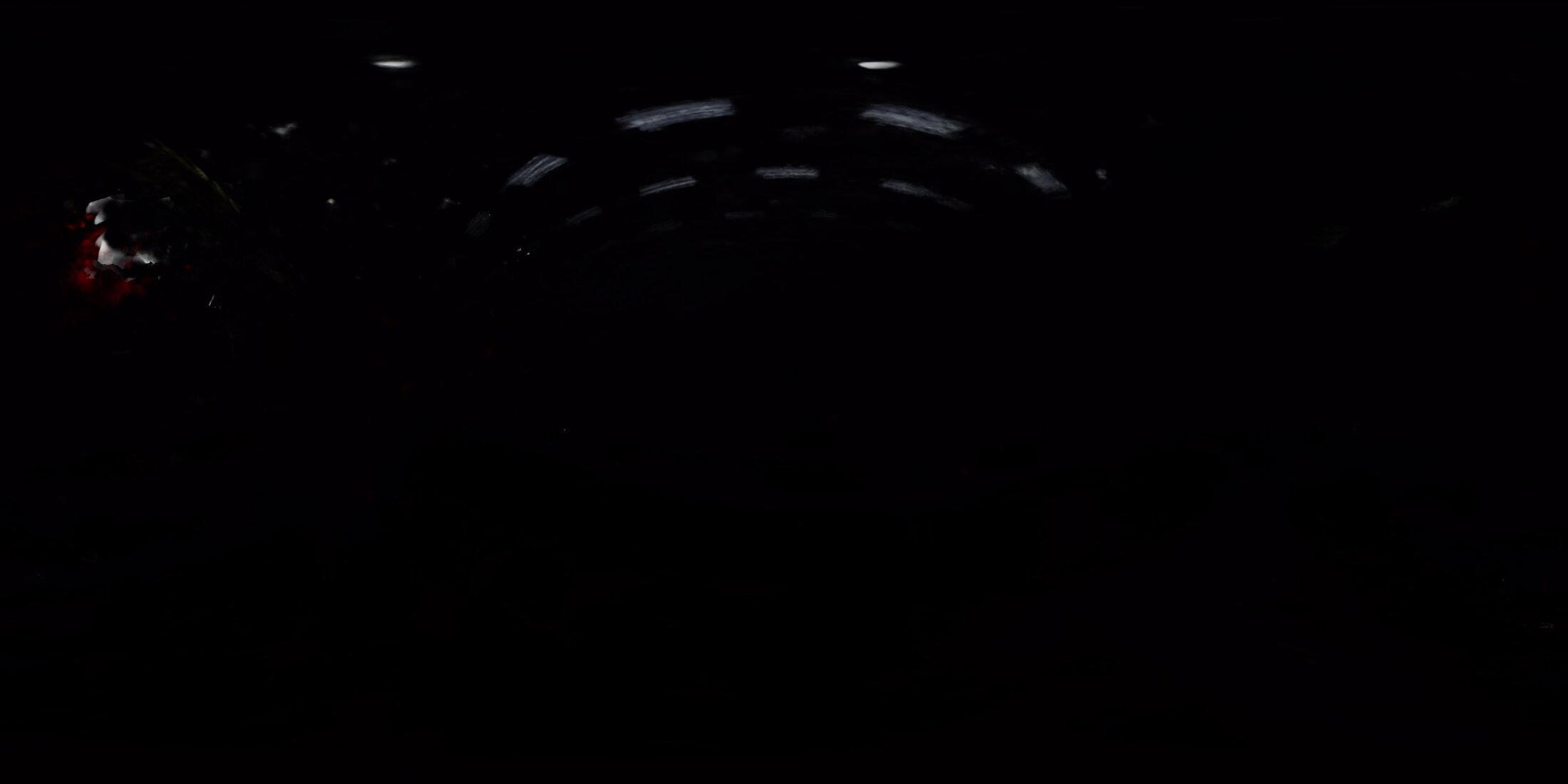}
\\
\includegraphics[width=\alignwidth]{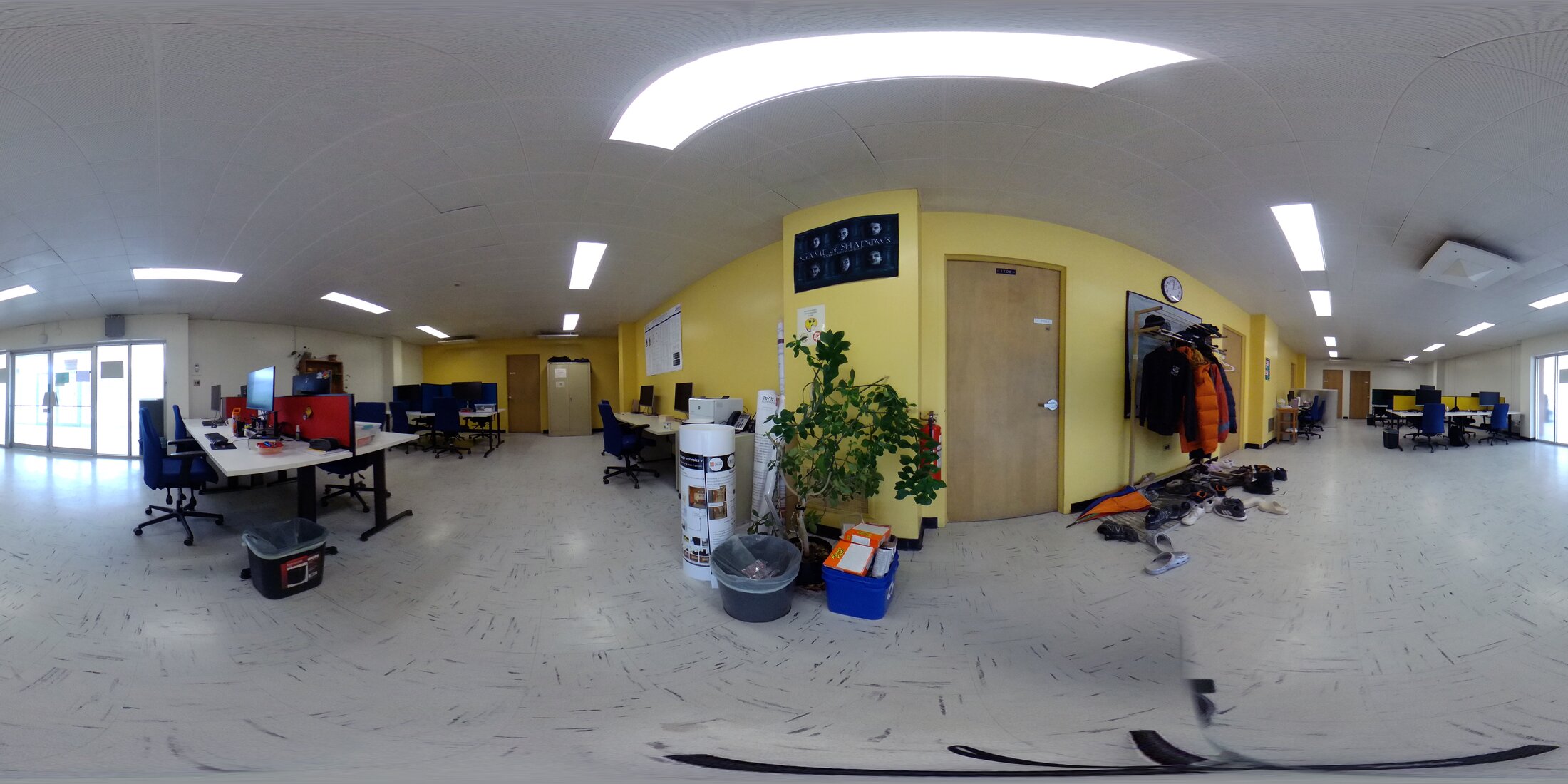} & 
\includegraphics[width=\alignwidth]{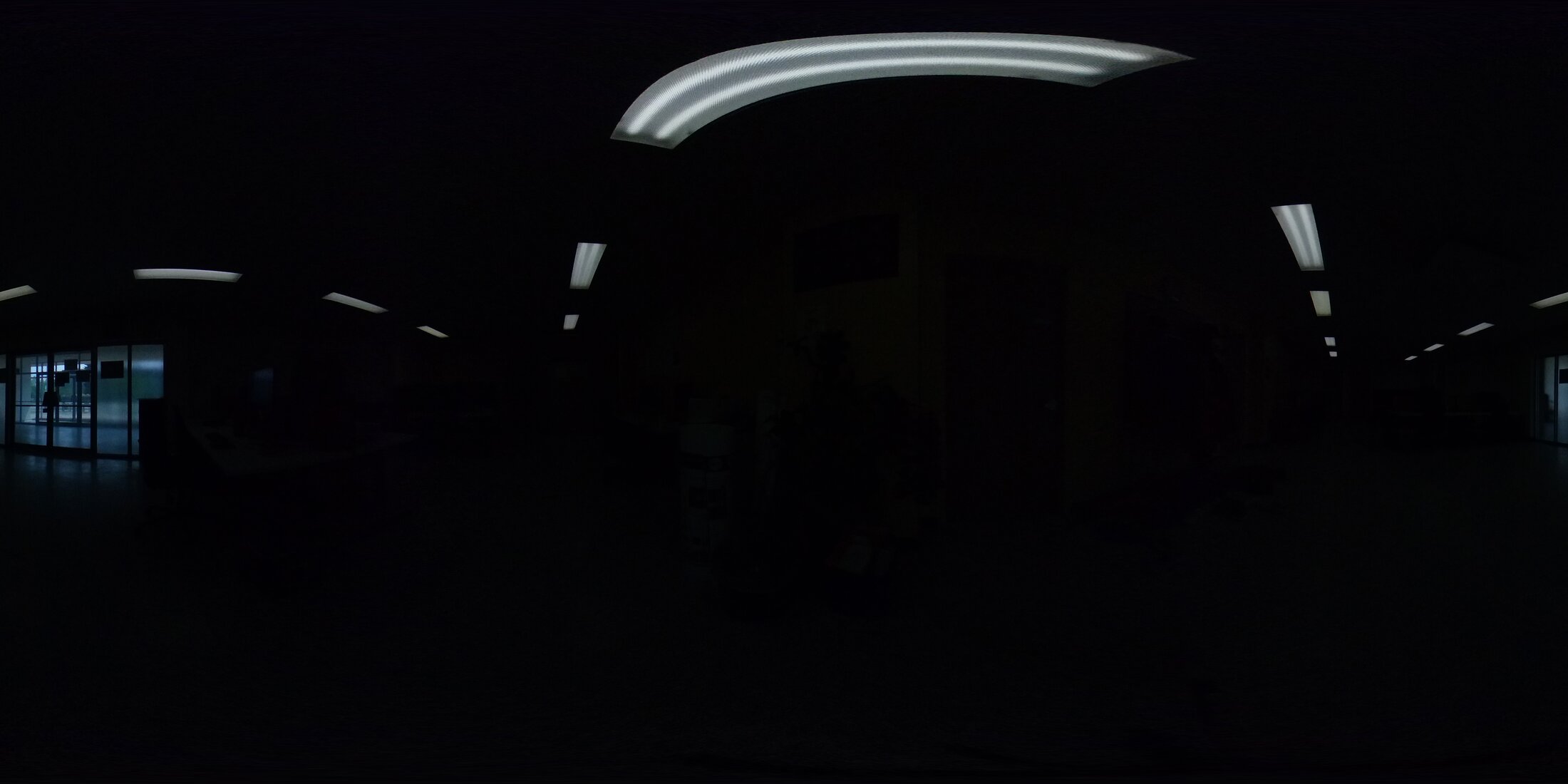} &
\includegraphics[width=\alignwidth]{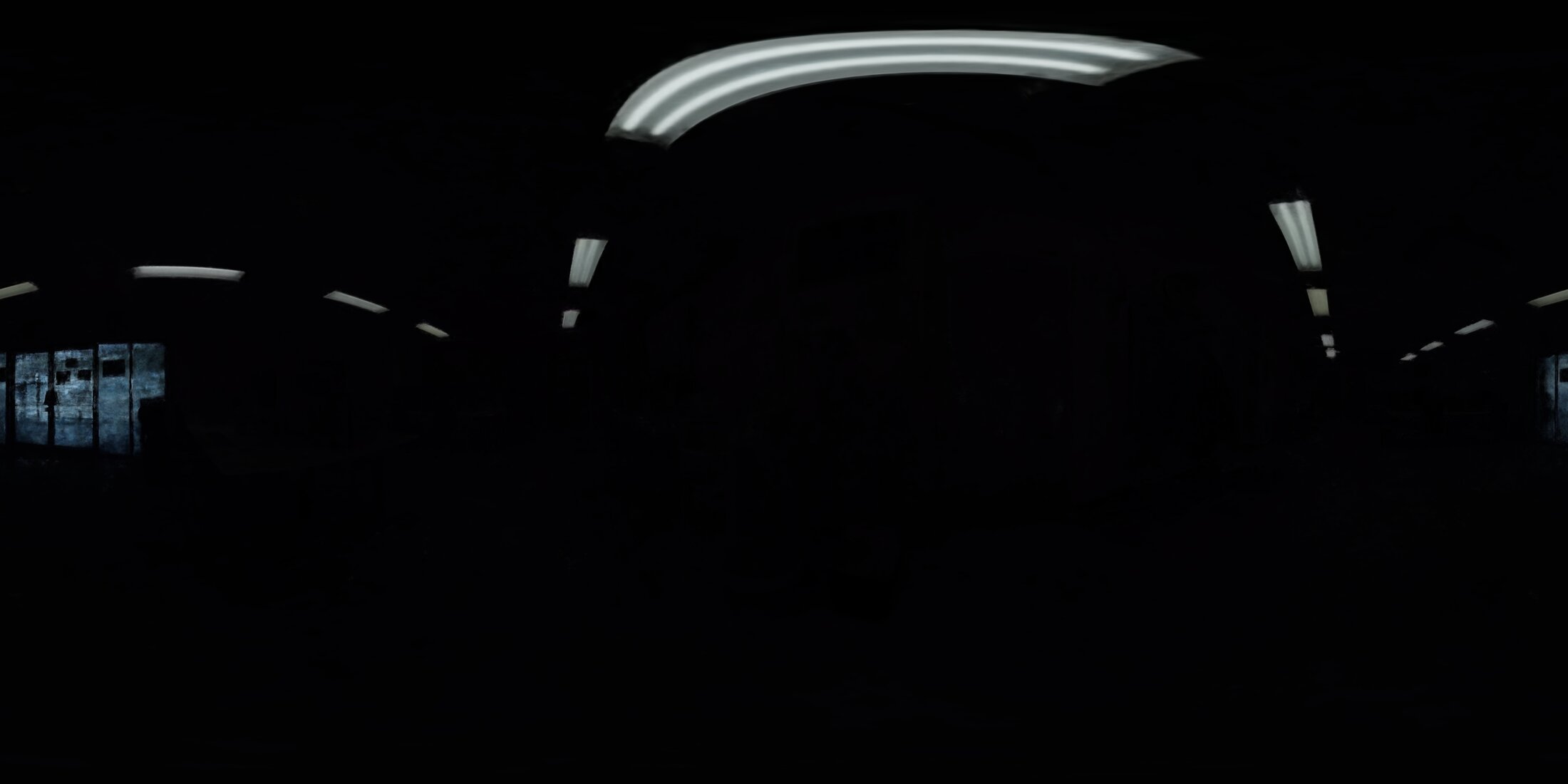} &
\includegraphics[width=\alignwidth]{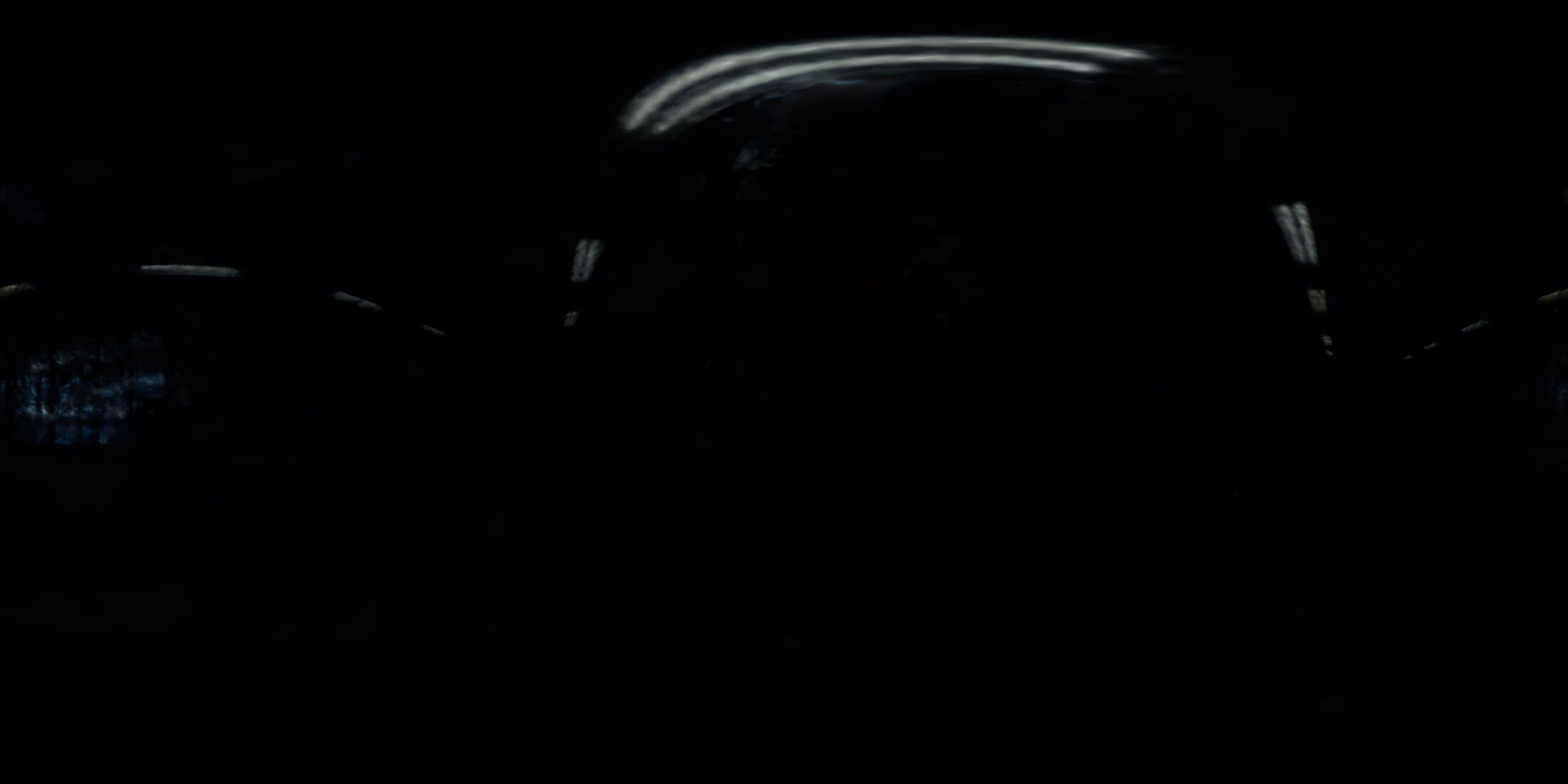} 
\\
\end{tabular}
\caption{\jf{Qualitative fast exposure reconstruction results. \thename produces results that more accurately capture the high dynamic range lighting~(3rd column) compared with HDR-Nerfacto~(4th column). Corresponding well-exposed images are shown as reference (1st column).}}
\label{fig:fast_exposed}
\end{figure*}

%% file: figures/histograms/histogram.tex
\begin{figure}
  \centering
  \includegraphics[width=0.8\linewidth]{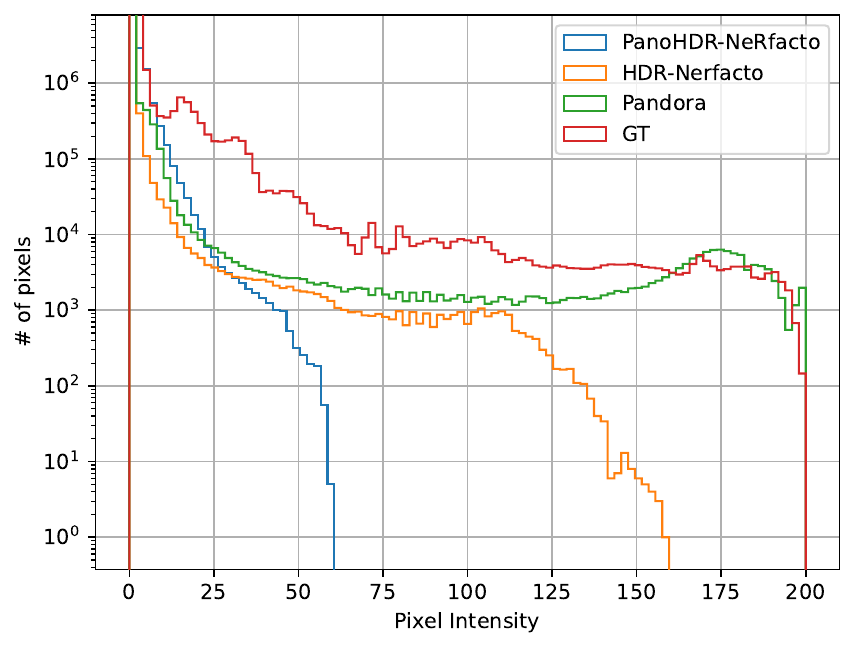}
  \caption{Histogram of pixel intensities after reconstructing the \sceneclubhouse scene. Compared to the ground truth, \thename more accurately recovers the scene's dynamic range than previous work: both \cite{gera2022casual} and \cite{huang2022hdrnerf} fail to reconstruct a large portion of the dynamic range.}
  \label{fig:histograms}
\end{figure}

%% file: tables/ablation.tex
\begin{table} [t]
\footnotesize
\setlength{\tabcolsep}{1pt}
\caption{Comparison of capture strategies. Capturing a single exposure (``Well-exposed'') and lifting it to HDR using an inverse tonemapping network (``Well+LDR2HDR~\cite{yu2021lanet}'') yields underperforming metrics. Capturing two simultaneous exposures helps, but our combination of well- and fast-exposures (``Well+fast'') yields the best performance, when compared to choosing another exposure (``Well+mid'').
Results are computed on the full HDR GT captures on 10 scenes from our dataset, the best metrics are highlighted in \colorbox{red!25}{red}.}
\vspace*{-4mm} 
\centering
\label{tab:ablation-two-exposure}
\begin{tabular}{clccccc}
\toprule
& & \multicolumn{3}{c}{HDR render}
& & LDR r. \\
\# exp.
& Strategy
& si-RMSE$_\downarrow$
& RMSE$_\downarrow$
& RGB ang.$_\downarrow$ 
& \,
& PSNR$_\uparrow$
\\
\midrule
1 & Well-exposed only
& 0.22 & 0.33 & 4.73 && 28.21
\\
1 & Well+LDR2HDR~\cite{yu2021lanet}  
& 0.17 & 0.29 & 5.84 && 28.61
\\
2 & Well+mid
& 0.11 & 0.17 & 2.48 && 31.07
\\
2 & Well+fast (ours) 
& \cellcolor{red!25}0.05 & \cellcolor{red!25}0.07 &\cellcolor{red!25} 1.75
& & \cellcolor{red!25}33.56
\\
\bottomrule
\end{tabular}
\end{table}

\begin{table} [t]
\footnotesize
\centering
\setlength{\tabcolsep}{1pt}
\caption{Effect of the fine alignment on quantitative metrics. Metrics are shown in 2 groups (left to right): LDR panos and HDR renders. The numbers are reported on the same scene as \cref{fig:alignment}, \sceneauditoriumbright, which contain many small, very bright light sources. The best metrics are highlighted in \colorbox{red!25}{red}.}
\vspace*{-4mm} 
\label{tab:ablation-pandora}
\begin{tabular}{lccccccc}
\toprule
& \multicolumn{3}{c}{LDR panos} 
& \;
& \multicolumn{3}{c}{HDR render} \\
Method
& SSIM$_\uparrow$
& PSNR$_\uparrow$
& LPIPS$_\downarrow$ 
& 
& si-RMSE$_\downarrow$ 
& RMSE$_\downarrow$ 
& RGB ang.$_\downarrow$ 
\\
\midrule
No fine align.  
& 0.57 & 17.53 & 0.41 
& & 0.050 & 0.066 & 4.72
\\
Fine align. 
& \cellcolor{red!25} 0.61 &  \cellcolor{red!25}18.34 &  \cellcolor{red!25}0.37 
& &  \cellcolor{red!25}0.042 &  \cellcolor{red!25}0.053 & \cellcolor{red!25}4.60
\\
\bottomrule
\end{tabular}
\end{table}


%% file: figures/spatially_varying_results/spatially_varying.tex
\begin{figure*}[!t]
    \centering
    \footnotesize
    \setlength{\tabcolsep}{1pt}
    \setlength{\mywidth}{0.19\linewidth}
    \begin{tabular}{ccccc}
    \includegraphics[width=\mywidth]{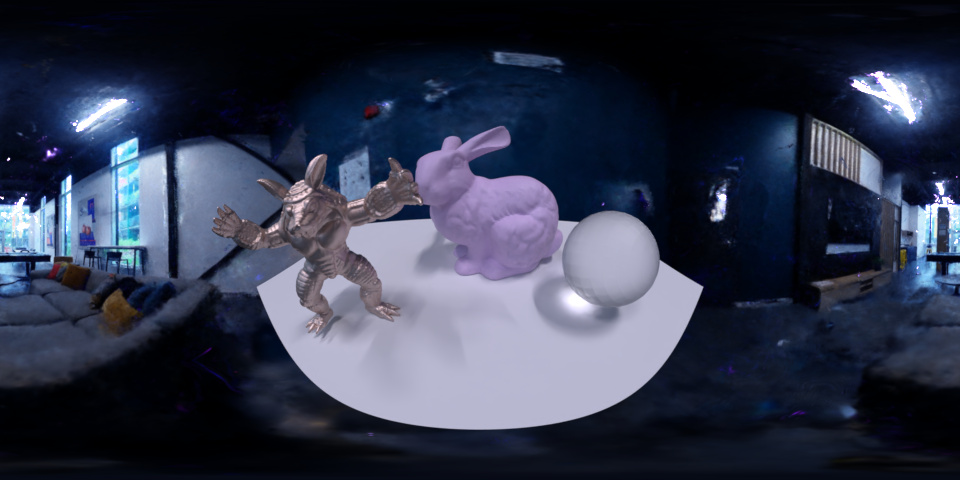} &
     \includegraphics[width=\mywidth]{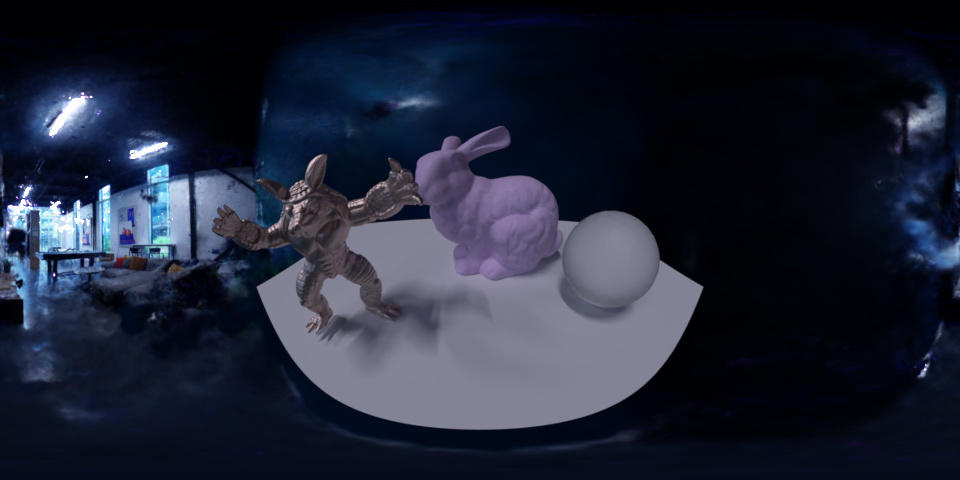} &
     \includegraphics[width=\mywidth]{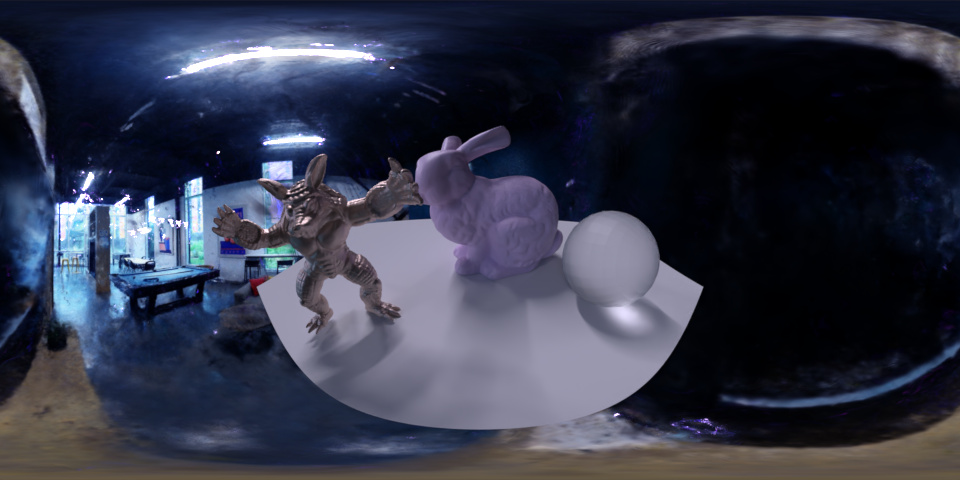} &
     \includegraphics[width=\mywidth]{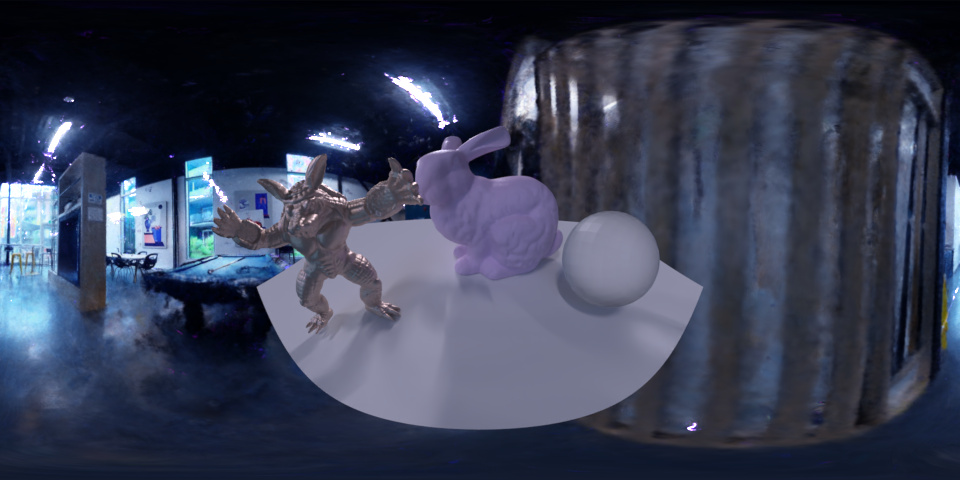} & 
     \includegraphics[width=\mywidth]{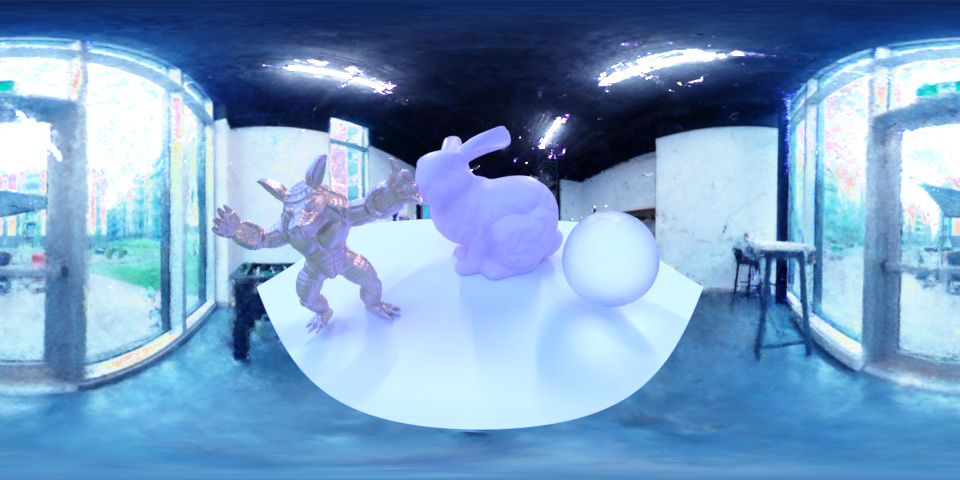} \\
     
     \includegraphics[width=\mywidth]{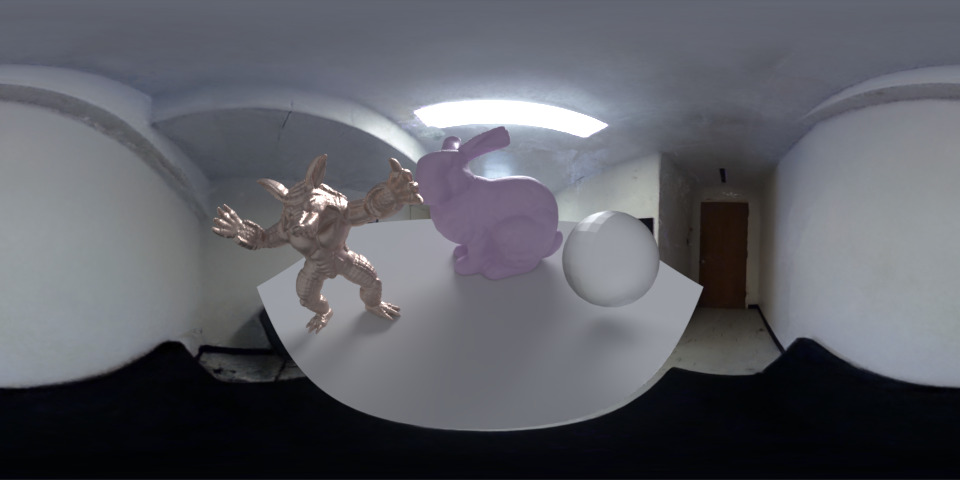} &
     \includegraphics[width=\mywidth]{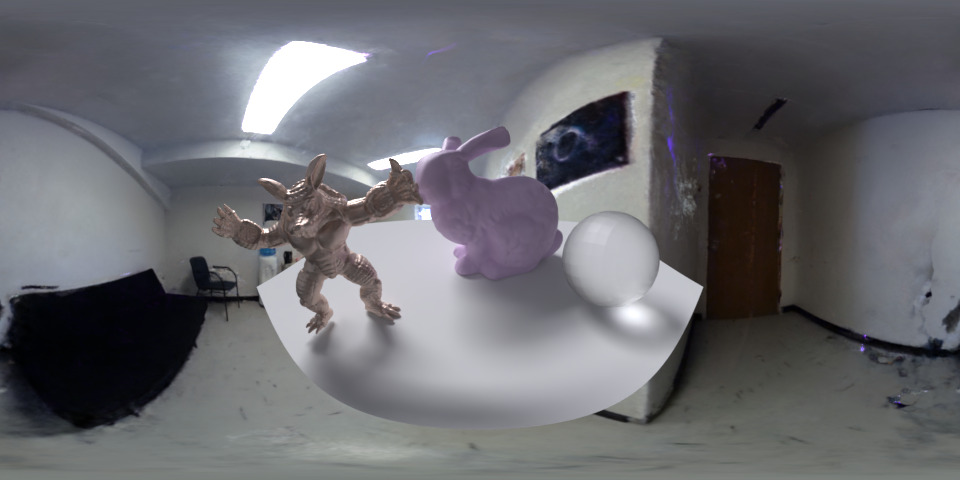} &
     \includegraphics[width=\mywidth]{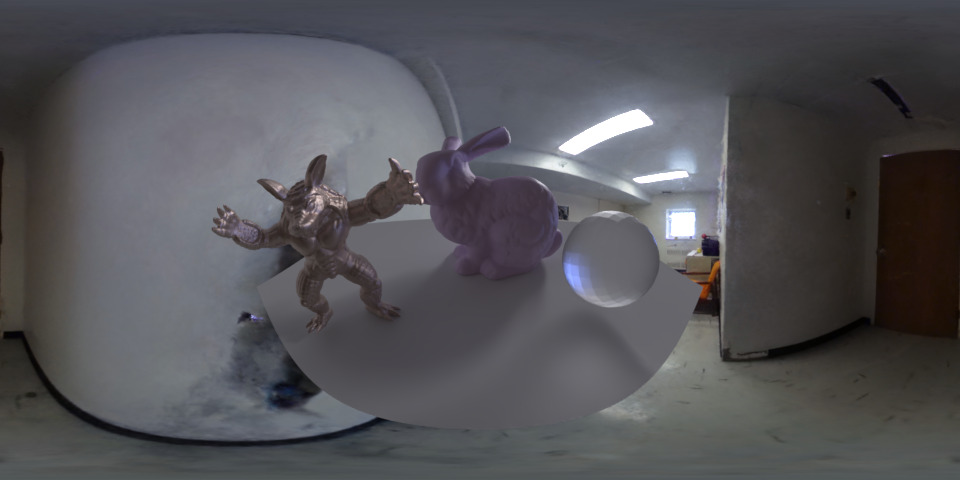} &
     \includegraphics[width=\mywidth]{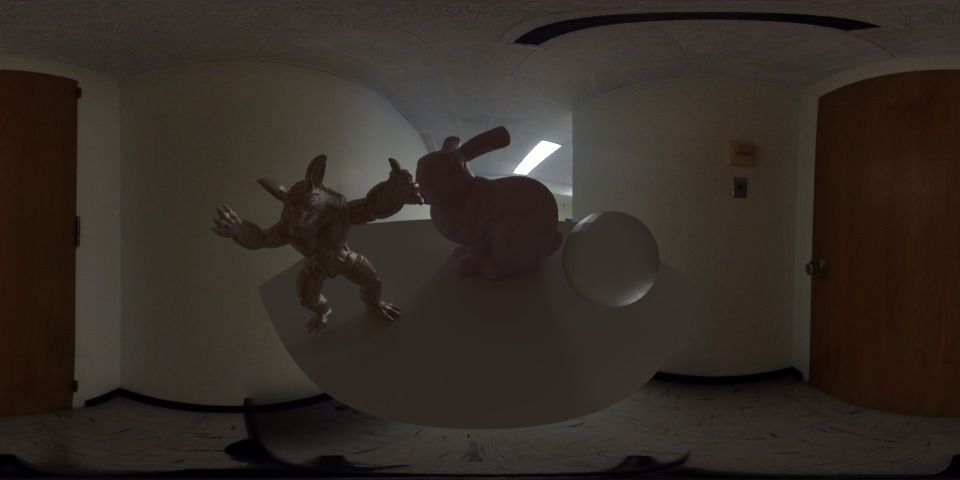} & 
     \includegraphics[width=\mywidth]{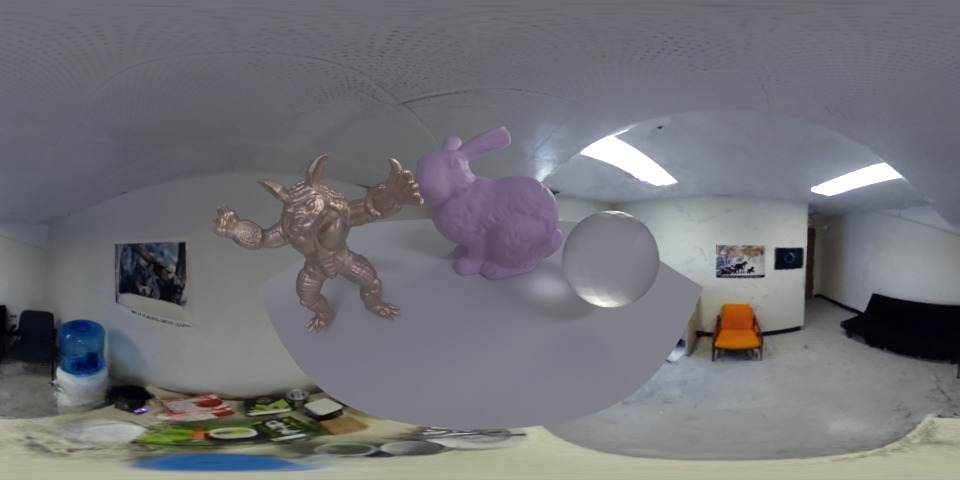} \\
     \includegraphics[width=\mywidth]{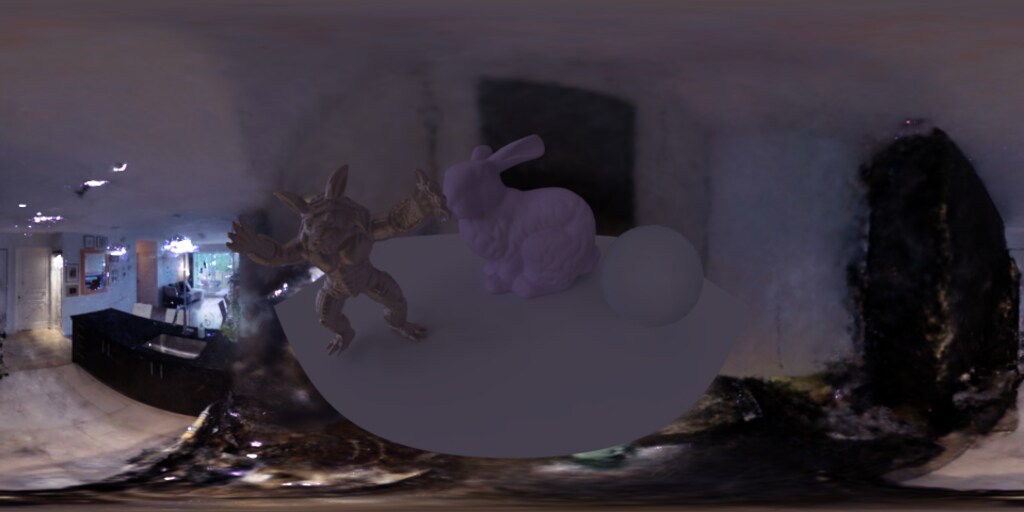} &
     \includegraphics[width=\mywidth]{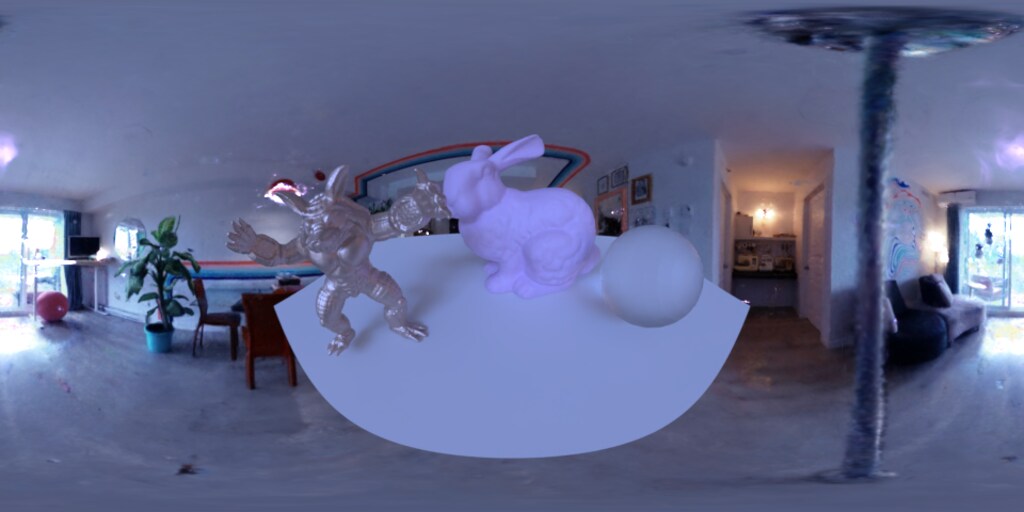} &
     \includegraphics[width=\mywidth]{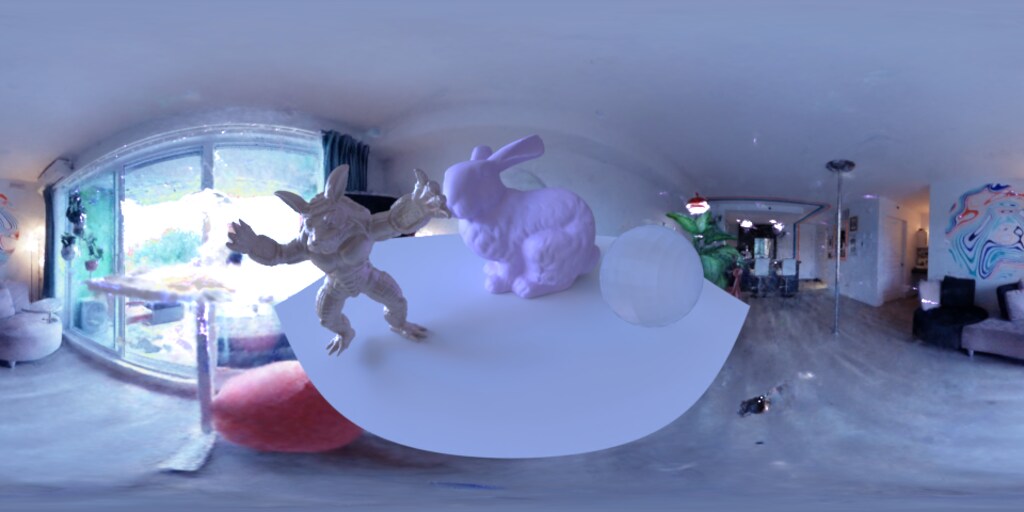} &
     \includegraphics[width=\mywidth]{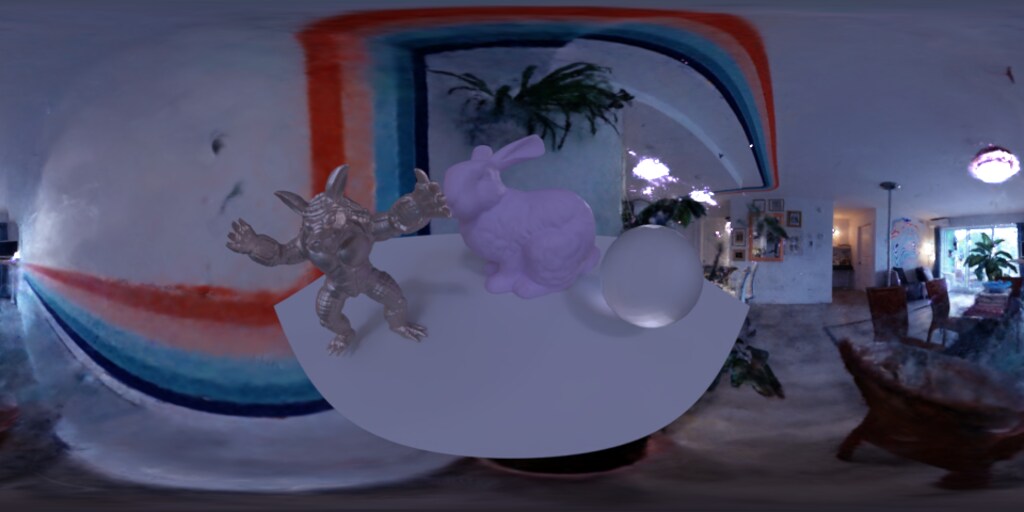} & 
     \includegraphics[width=\mywidth]{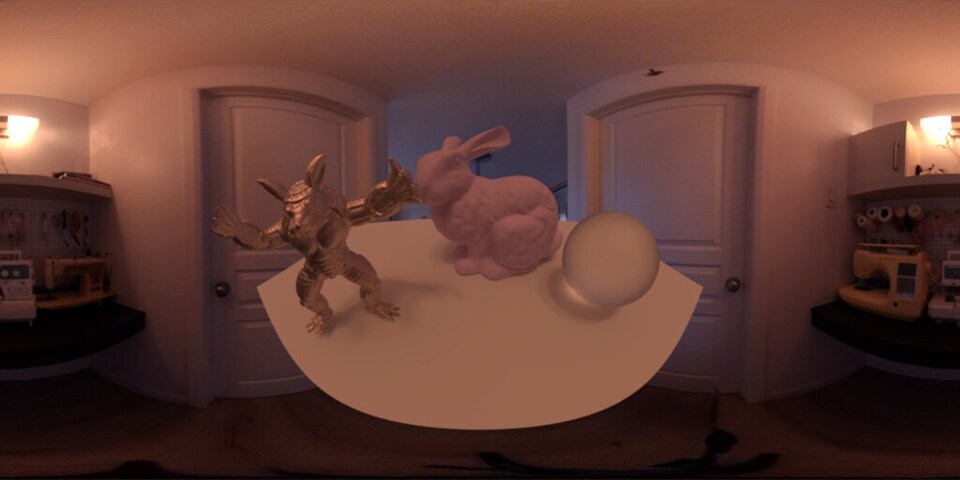}
    \end{tabular}
    \caption{\jf{Spatially-varying lighting reconstruction. Predicted lighting at multiple locations within three scenes with strong spatial variation (top to bottom: \sceneclubhouse,  \scenecoffeeroom, \scenerainbowlivingroom). Our method recovers coherent lighting across each scene's full dynamic range, from bright window regions to dark corners, capturing how illumination varies with position rather than predicting a single global estimate.}}
    \label{fig:spatially_varying}
\end{figure*}

%% file: tables/gsplats_vs_ours.tex
\begin{table} [t!]
\scriptsize
\centering
\setlength{\tabcolsep}{2pt}
\caption{Quantitative comparison of the \thename pipeline using NeRF vs. 3DGS across ten scenes. The best metrics are highlighted in \colorbox{red!25}{red}.}
\vspace*{-4mm} 
\label{tab:nerf-vs-gsplats-results}
\begin{tabular}{lcccccccccc}
\toprule
& \multicolumn{3}{c}{LDR panos}
& & \multicolumn{3}{c}{HDR render}
& & LDR r. \\
Method
& PSNR$_\uparrow$
& SSIM$_\uparrow$
& LPIPS$_\downarrow$
& \,
& si-RMSE$_\downarrow$
& RMSE$_\downarrow$
& RGB ang.$_\downarrow$ 
& \,
& PSNR$_\uparrow$ \\
\midrule
3DGS	
&   19.14 & 0.64 & 0.46	
& & 0.13 & 0.21	& \cellcolor{red!25}2.89	
& & 28.70 \\
NeRF	
&   \cellcolor{red!25}20.32	& \cellcolor{red!25}0.66	& \cellcolor{red!25}0.39
& & 0.13	& \cellcolor{red!25}0.18	& 2.93	
& & \cellcolor{red!25}29.60 \\
\bottomrule
\end{tabular}
\end{table}

\begin{figure}[t]
  \footnotesize
  \centering
  \setlength{\tabcolsep}{1pt} 
  \begin{tabular}{ccc} 
  GT & \thename (NeRF) &\thename (3DGS) \\
  
  \includegraphics[width=0.325\linewidth]{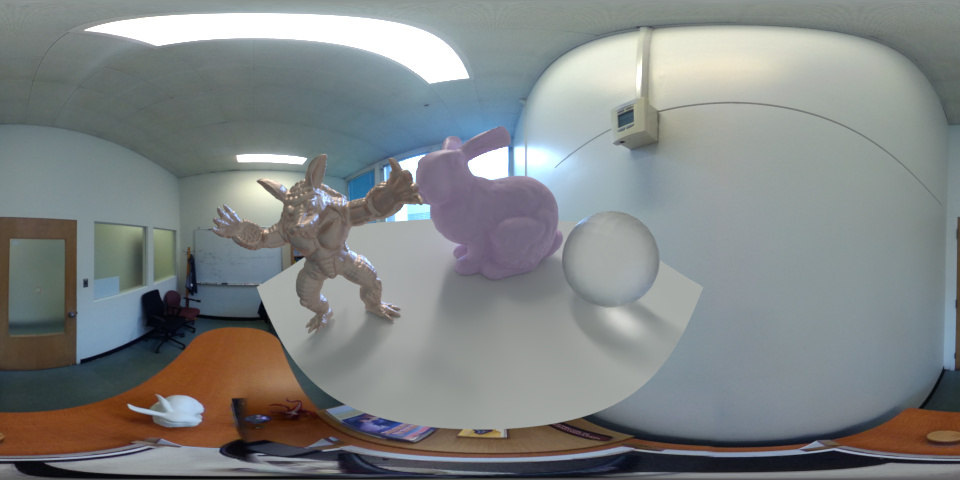}  & 
  \includegraphics[width=0.325\linewidth]{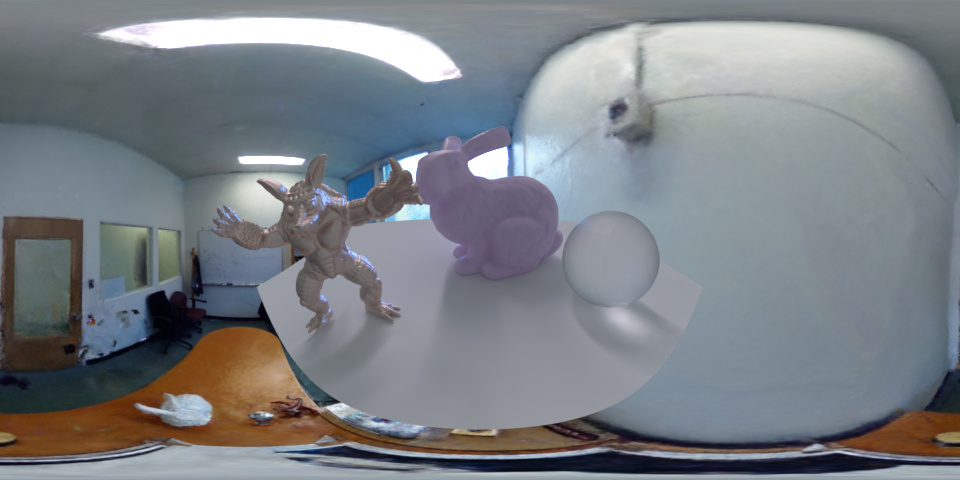} & \includegraphics[width=0.325\linewidth]{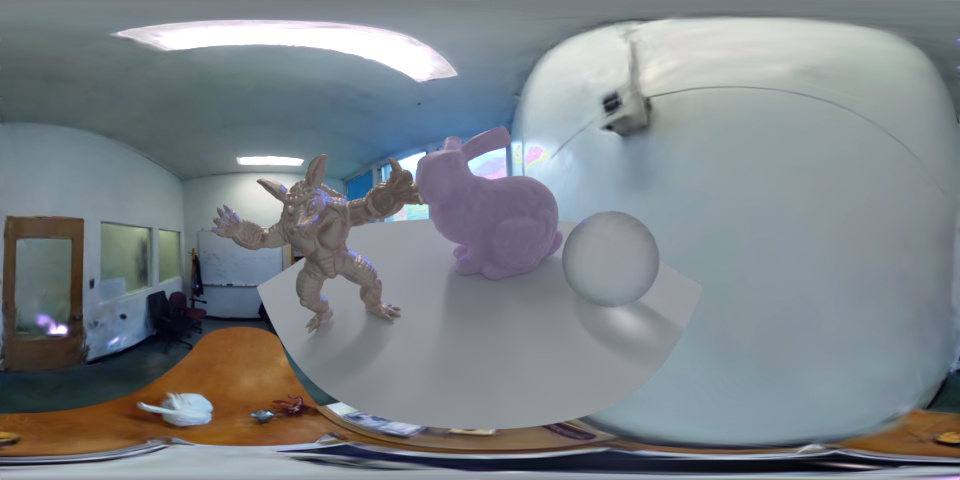} \\

  \includegraphics[width=0.325\linewidth]{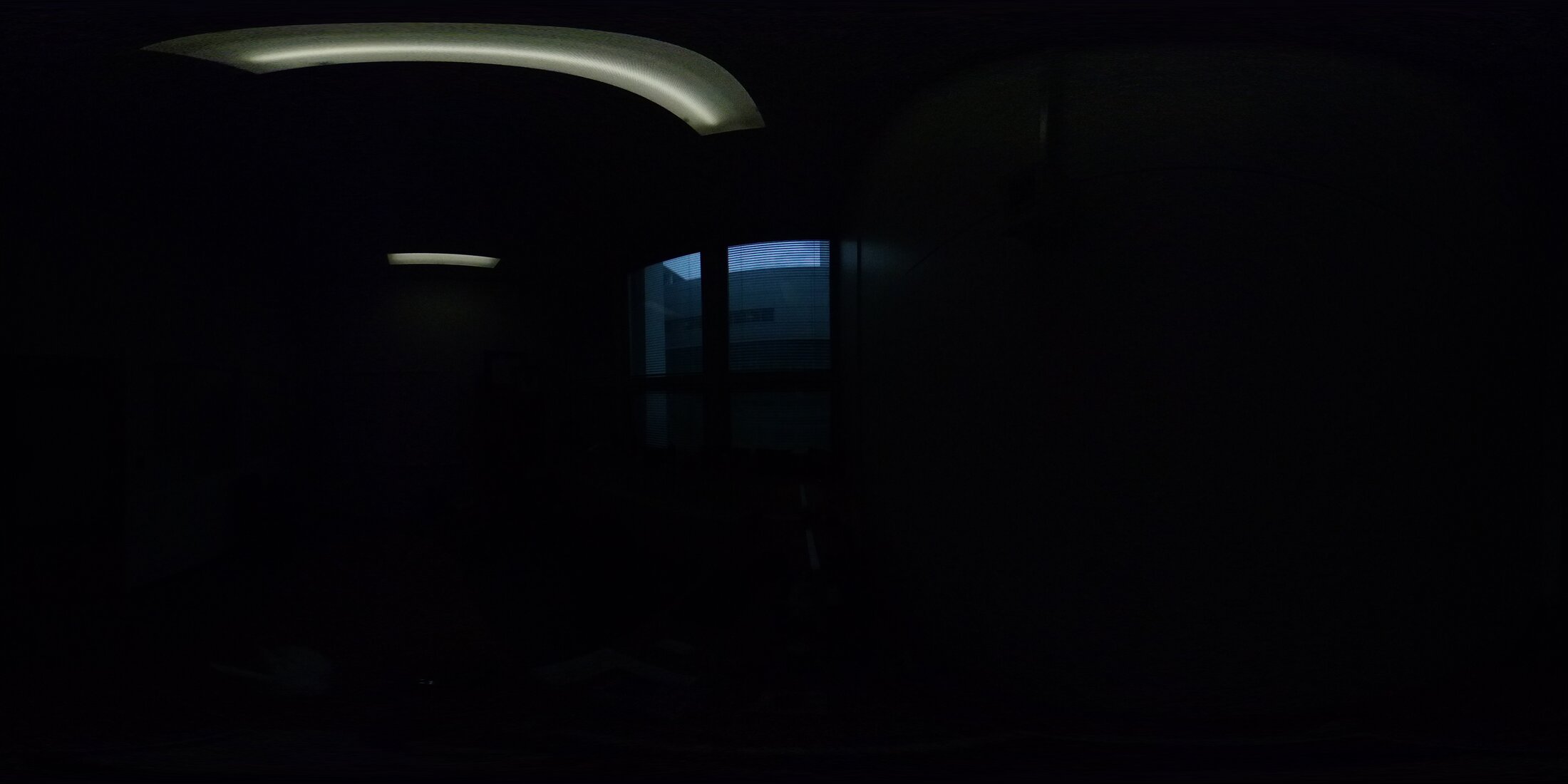} & 
  \includegraphics[width=0.325\linewidth]{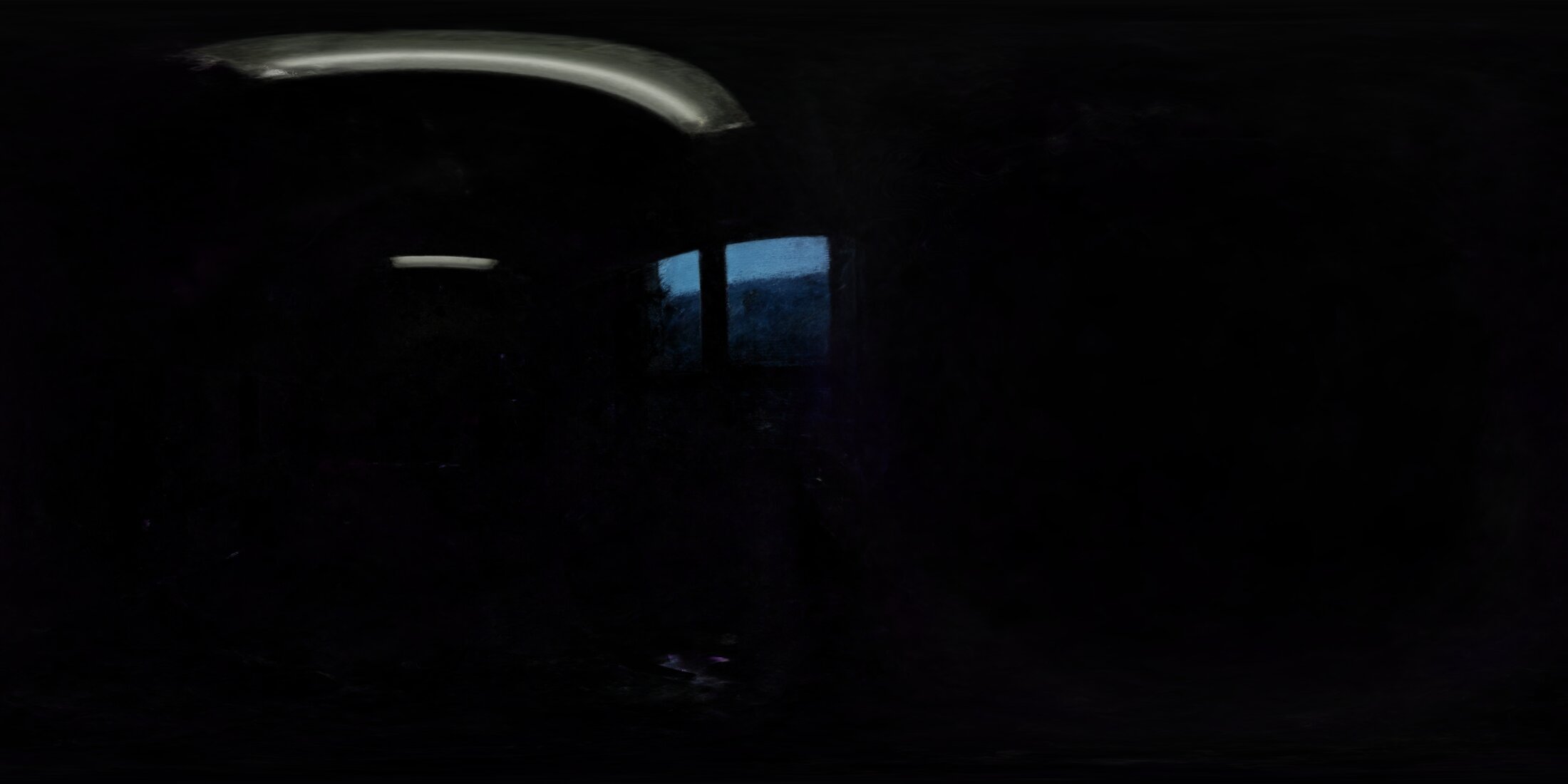} & \includegraphics[width=0.325\linewidth]{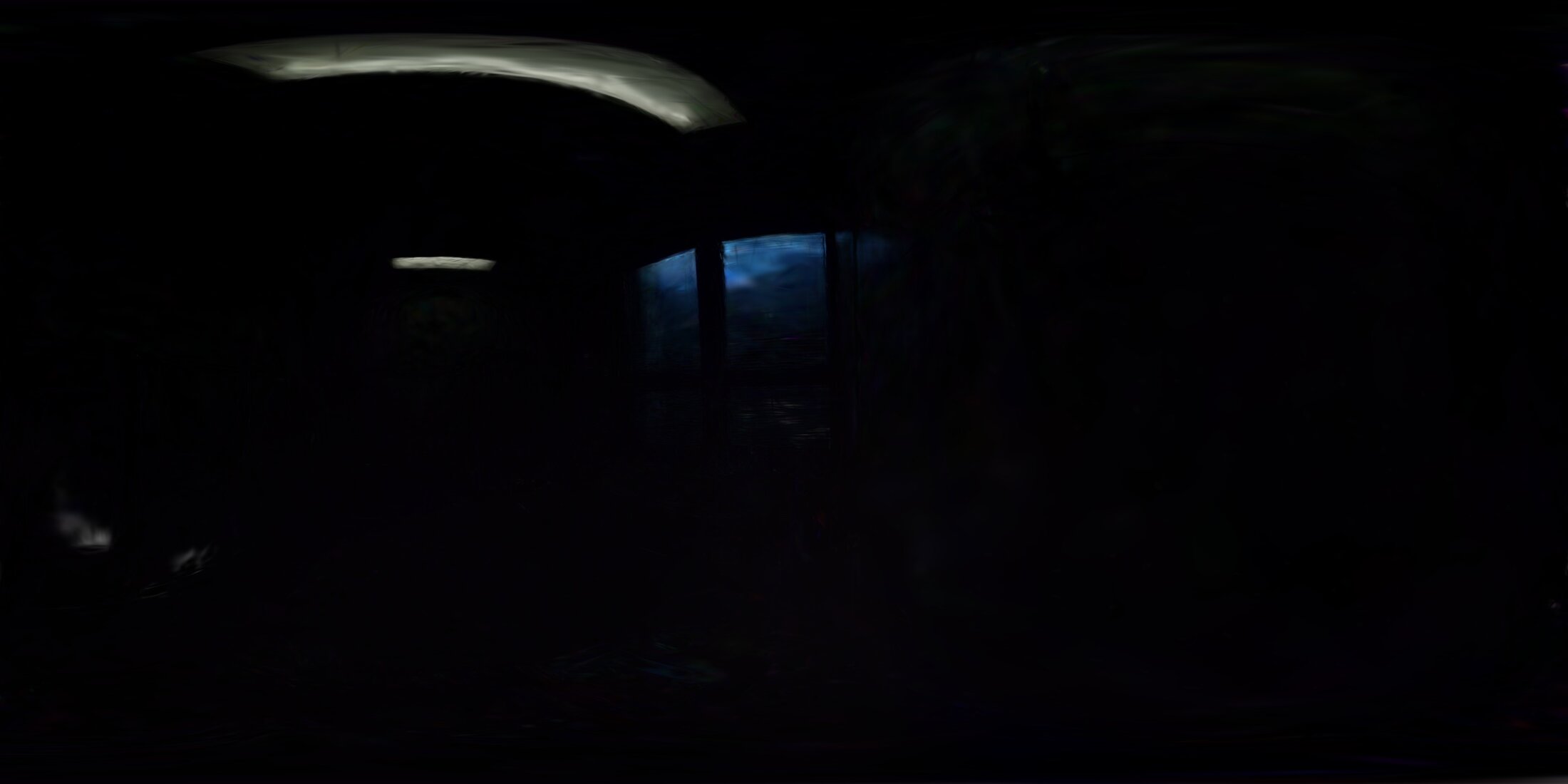} \\
  
  \includegraphics[width=0.325\linewidth]{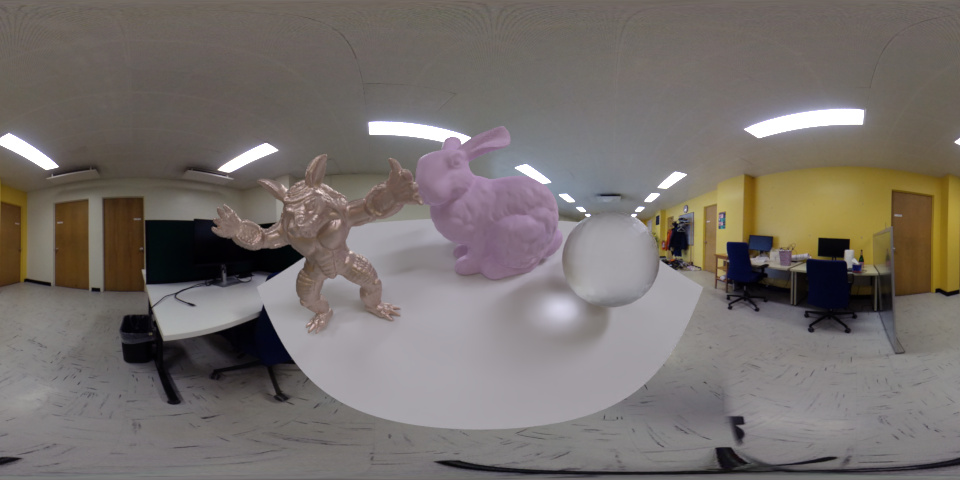} &
  \includegraphics[width=0.325\linewidth]{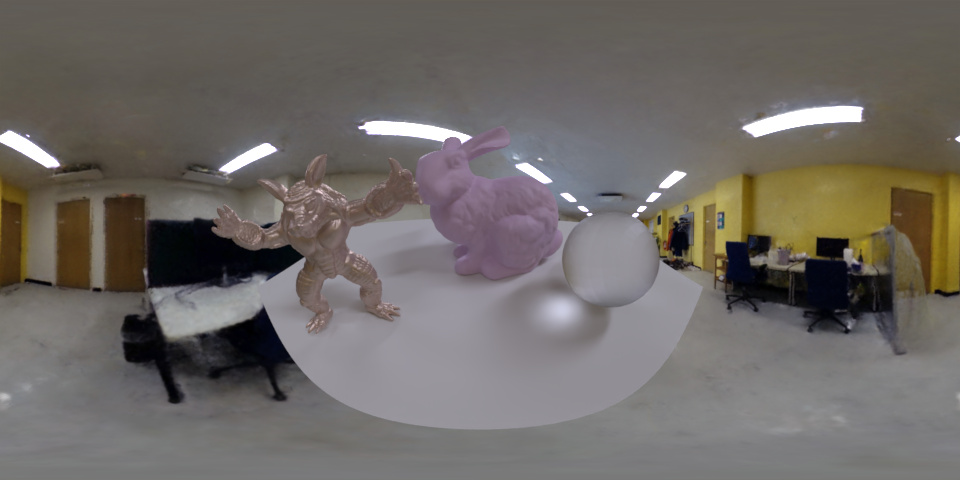} &  \includegraphics[width=0.325\linewidth]{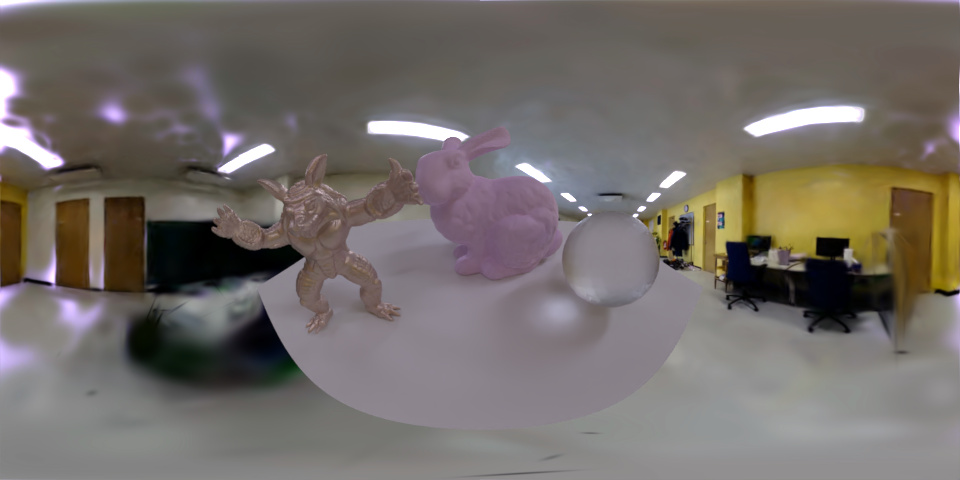} \\

\includegraphics[width=0.325\linewidth]{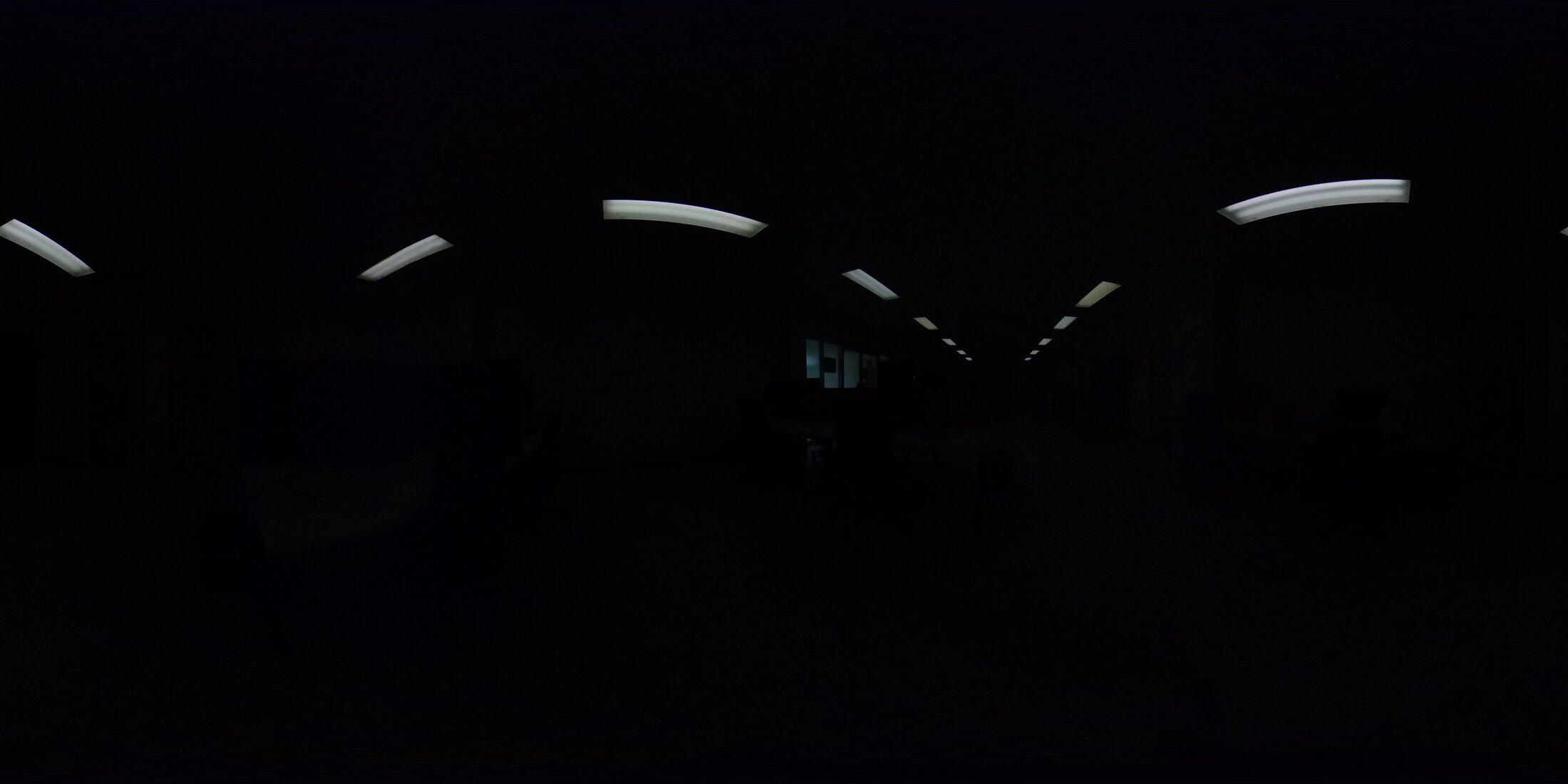} &
  \includegraphics[width=0.325\linewidth]{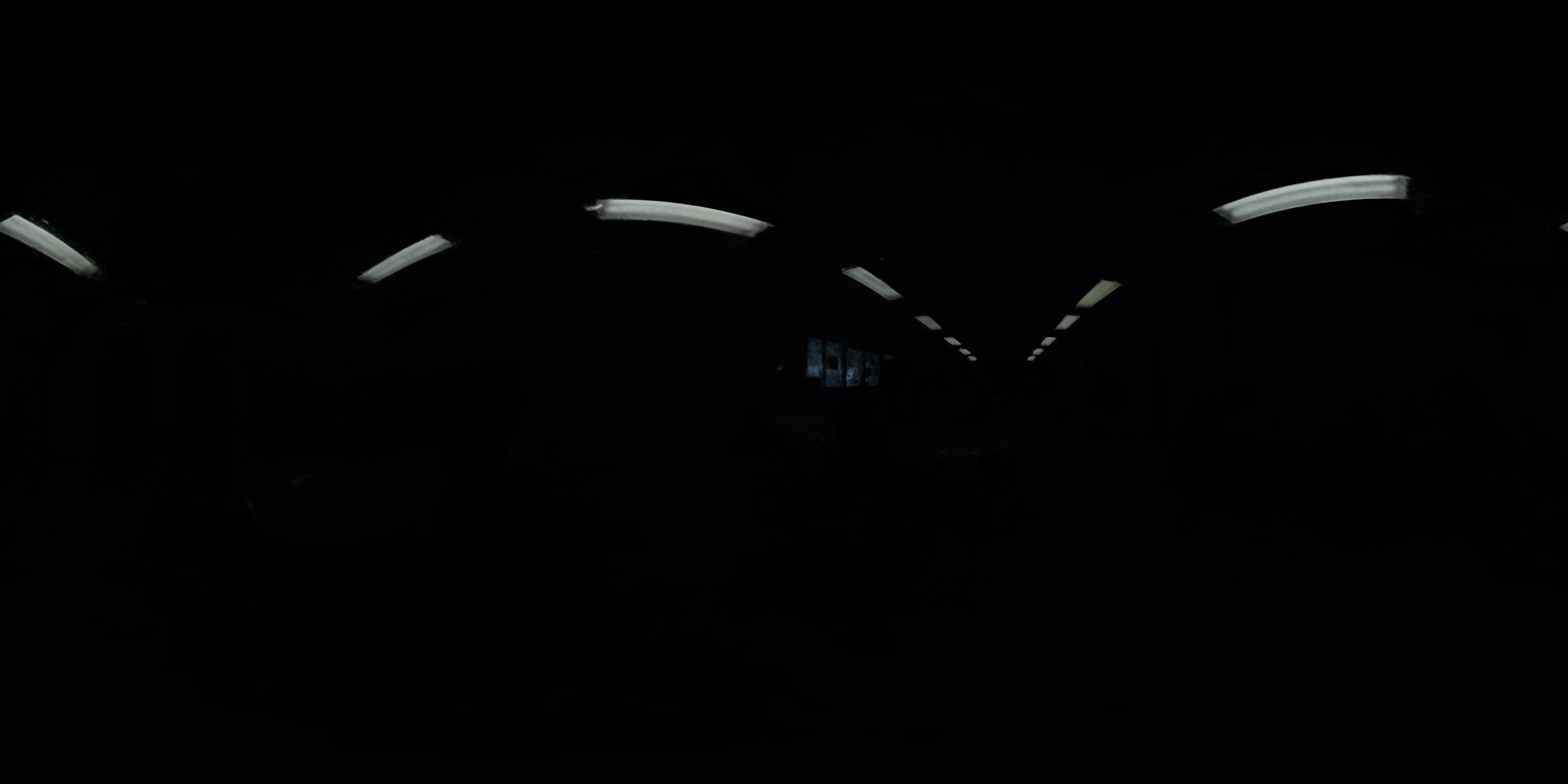} &  \includegraphics[width=0.325\linewidth]{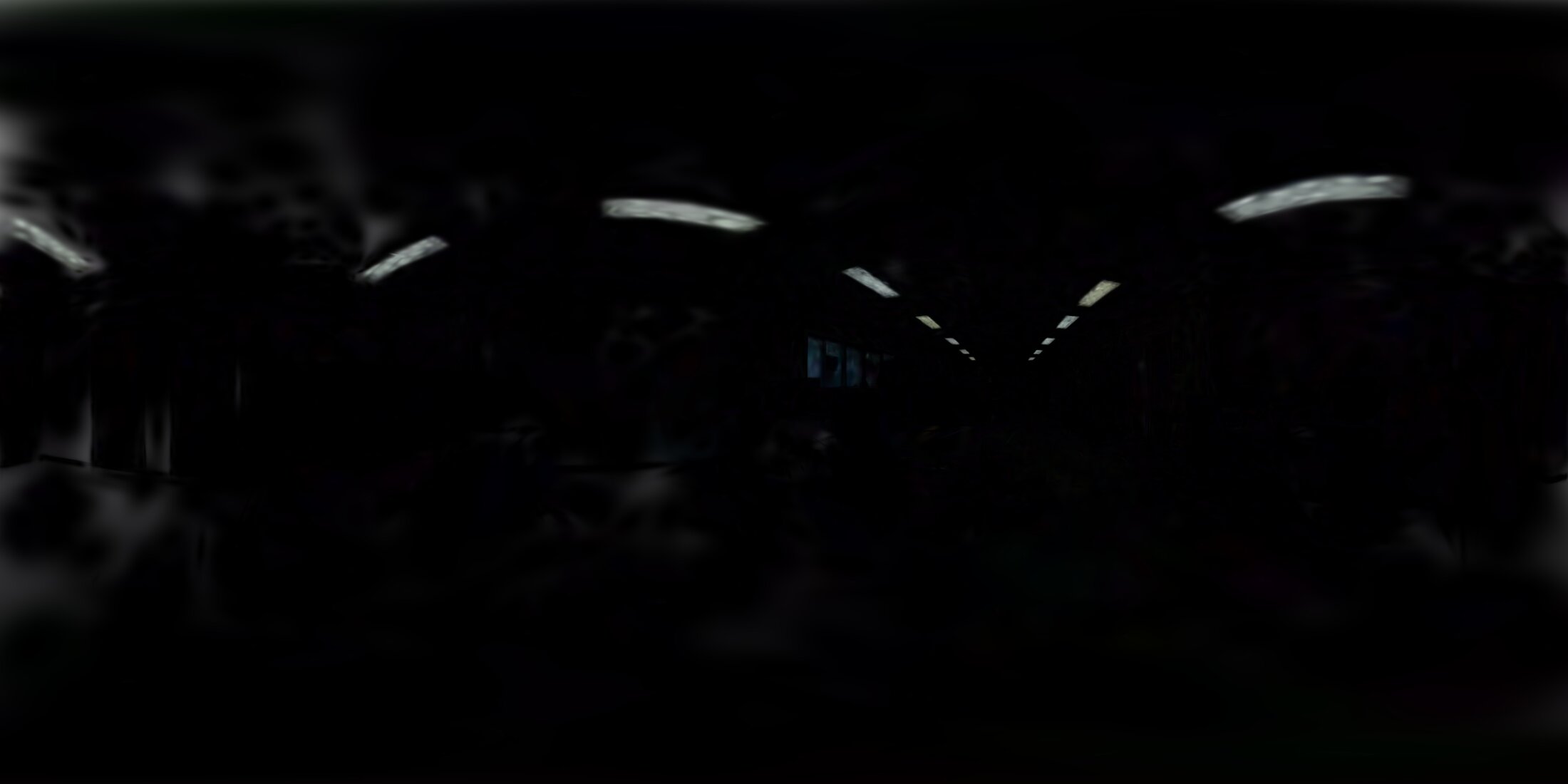}
  \end{tabular}
  \caption{Qualitative comparison of the \thename pipeline reveals a clear trade-off between NeRF and 3DGS. In our experiments, NeRF provides superior visual quality at the cost of speed, while 3DGS prioritizes real-time performance. \jf{Here, examples are shown on the \scenejfoffice and \scenelvsn scenes, respectively.}}
  \label{fig:gsplats}
\end{figure}

%% file: sections/5_discussion.tex

\input{figures/limitation/ablation_figure}

\section{Discussion}

We present \thename: a PANoramic Dual-Observer Radiance Acquisition system for easily capturing spatially-varying HDR radiance fields in indoor scenes. By utilizing our simple and inexpensive capture device, users can record a scene's dynamic range, including bright light sources, from multiple viewpoints. Our method generates HDR environment maps at any 3D position within the scene, ready for immediate use in standard rendering engines. As demonstrated in \cref{fig:teaser}(c) and \cref{fig:object-insertion}—which showcase virtual objects from different perspectives—this approach enables high-fidelity digital staging, augmented reality, and coherent object relighting with full spatial consistency. 

\myparagraph{Limitations and future work} Our system captures only two exposures, potentially missing the dynamic range ``in between''. Bright light sources are captured without saturation; however, some regions will be under-exposed in fast-exposed images and over-exposed (saturated) in well-exposed images, resulting in artifacts such as halos around lights (\cref{fig:limitation}, top). An improved HDR interpolation strategy may help address this issue. Additionally, our system is designed for indoor capture. Consumer cameras currently available cannot capture the sun's radiance with their fastest shutter speed, resulting in saturated observations and erroneous relighting. \jf{Moreover, our two-exposure capture sometimes cannot reconstruct the darkest parts of a scene. Since these regions contribute negligibly to light emission, this omission does not meaningfully affect image-based lighting. Like many NeRF-based approaches, our method is prone to motion blur if the user moves the apparatus too fast. The presence of significant motion blur could either make the self-calibration method fail (thereby naturally filtering out blurry images) or result in oversmooth NeRF reconstructions.}

\noindent~Although our fine alignment procedure (c.f. \cref{sec:camera_calibration}) is robust, it may still not be perfect. Even a slight misalignment left can cause reconstruction errors in small, bright light sources such as spotlights (\cref{fig:limitation}, bottom). Fine-tuning camera poses through backpropagation~\cite{lin2021barf} was attempted, but it did not significantly improve alignment. \jf{In addition, the fine alignment technique struggles with regions that fall between the two captured exposures.} Developing reconstruction methods adapted to vast gaps in dynamic ranges is a key area for future work. Additionally, the necessity for rapid exposures inherently introduces low-light noise into the captured dataset. Leveraging methods for radiance field reconstruction under low-light~(e.g., \cite{qu2024lush, cui_luminance_gs, wang2023lighting}) offers a compelling avenue for future work. We hope our approach and dataset will help spur this research effort forward.



%% file: figures/limitation/ablation_figure.tex
\begin{figure}[t]
  \footnotesize
  \centering
  \setlength{\tabcolsep}{1pt} 
  \begin{tabular}{cc} 
  GT & \thename (ours) \\
  \includegraphics[width=0.45\linewidth]{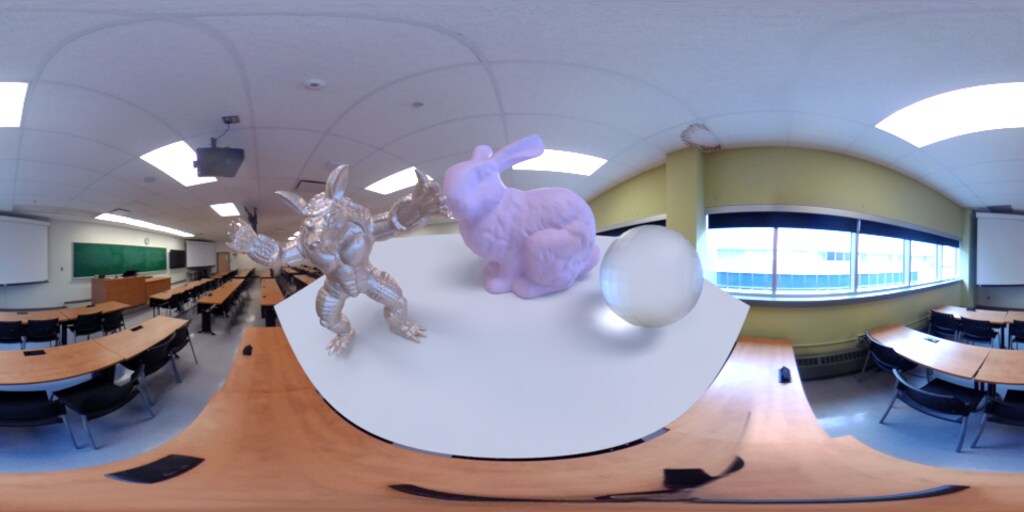}  &
  \includegraphics[width=0.45\linewidth]{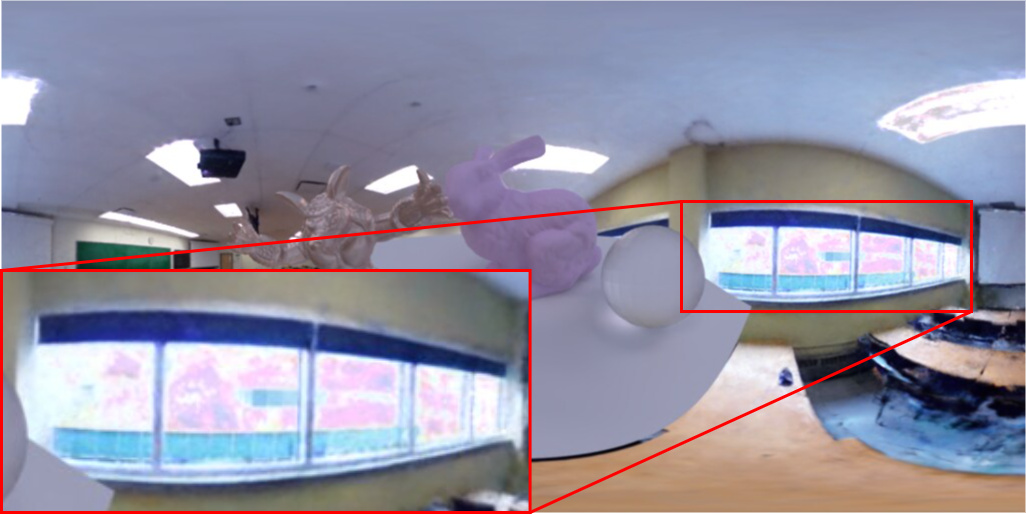} \\
  \includegraphics[width=0.45\linewidth]{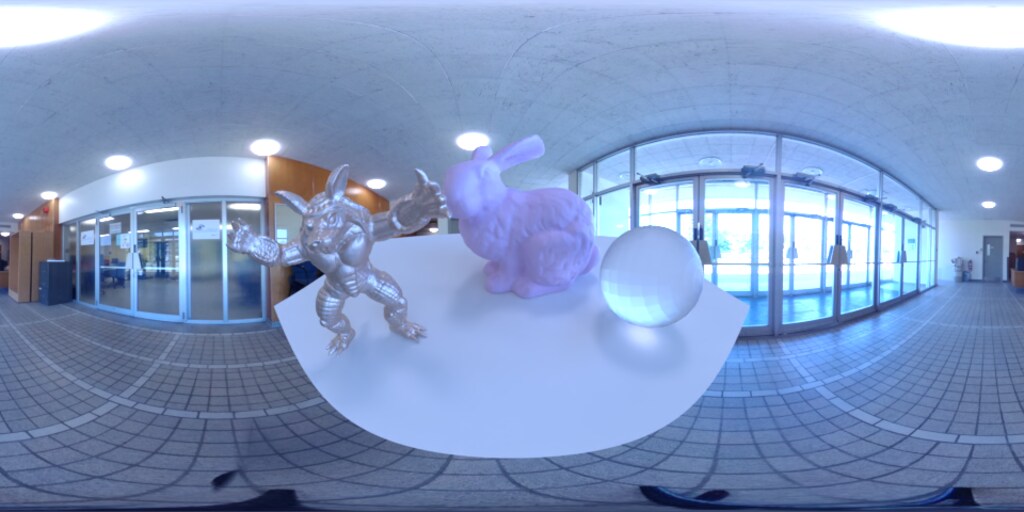} &
  \includegraphics[width=0.45\linewidth]{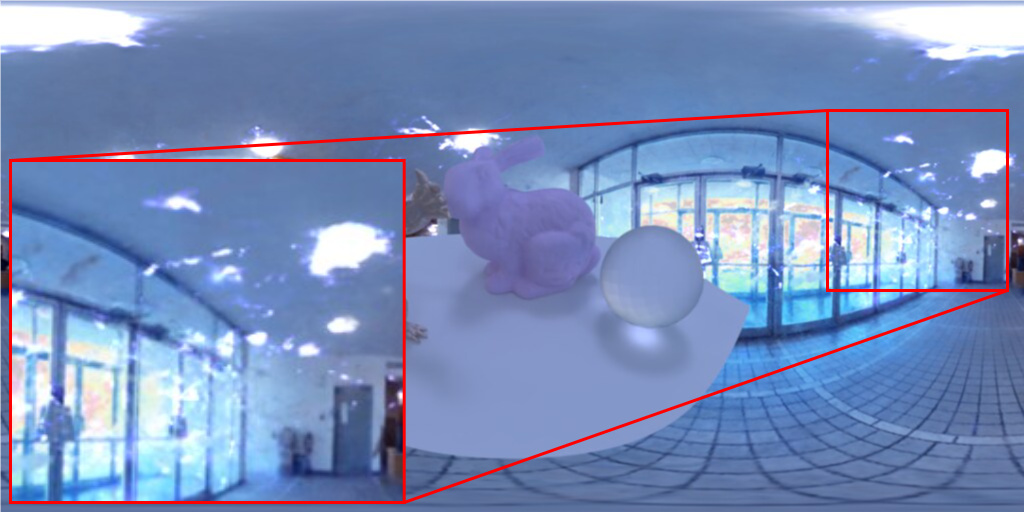}
  \end{tabular}
  \caption{Limitations. Our system captures only two exposures, which can result in missing parts of the dynamic range in some regions (top-right). Moreover, when scenes contain small light sources, precise alignment between the well-exposed and fast-exposed sequences becomes critical. Despite improving results, our fine alignment strategy is not always perfect and slight misalignment between exposures can lead to inaccurate reconstructions (bottom-right).}
  \label{fig:limitation}
\end{figure}

%% file: sec/X_suppl.tex

\section{Quantitative results per scene}
\Cref{tab:quantitative-extended-results} and \Cref{tab:quantitative-extended-results_2} show all quantitative results for each scene in our evaluation dataset individually. The same metrics as in tab.~1 from the main paper are used.


\section{More details on HDR-Nerfacto}





\subsection{Implementation details}

Our implementation of HDR-Nerfacto matches as closely as possible the original HDR-NeRF method while taking advantage of the strengths of the Nerfacto framework. Following the original HDR-NeRF implementation, we keep the HDR radiance field MLP output in the log domain and add it to the log-exposure before feeding it to the RGB CRF MLPs. We swap the HDR radiance field to the Nerfacto backbone. We remove the ReLU activation function from the output layer of this backbone, as it would prevent the model from outputting necessary negative log-exposures. The three CRF MLPs from HDR-NeRF are implemented the same way as in the original implementation, using two-layer MLPs with a width of 128.
We note the importance of disabling Nerfacto's appearance embedding, which essentially makes the network invariant to different exposures. As in HDR-NeRF, we include the unit exposure loss during training. 

\subsection{Obtaining HDR radiance}

\begin{figure}
    \centering
    \tiny	
    \setlength{\tabcolsep}{1pt}
    \begin{tabular}{ccccc}\\
      EV & -11.0 & -3.0 & 0.0 & 3.0 \\
       \rotatebox{90}{\shortstack{Radiance \\ Output}} &
       \includegraphics[width=0.23\linewidth]{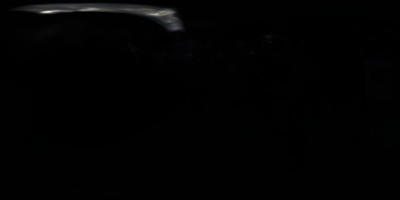} & 
       \includegraphics[width=0.23\linewidth]{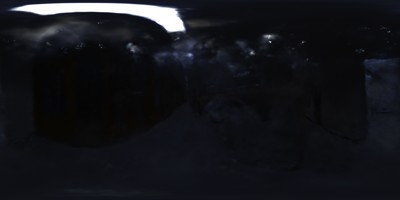} & 
       \includegraphics[width=0.23\linewidth]{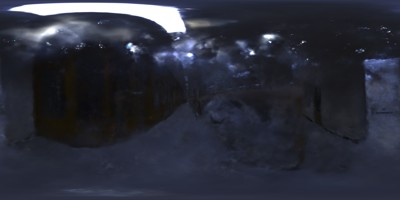} & 
       \includegraphics[width=0.23\linewidth]{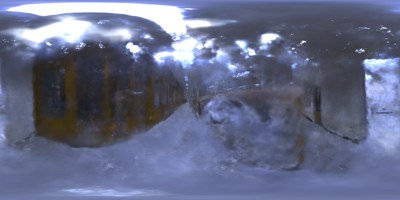} \\
       \rotatebox{90}{Combined} &
       \includegraphics[width=0.23\linewidth]{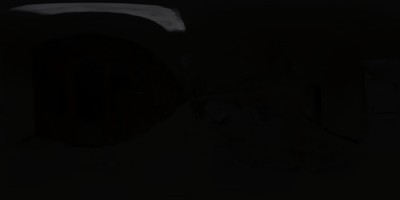} & 
       \includegraphics[width=0.23\linewidth]{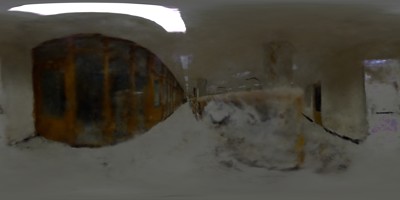} & 
       \includegraphics[width=0.23\linewidth]{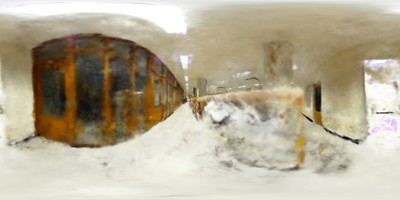} & 
       \includegraphics[width=0.23\linewidth]{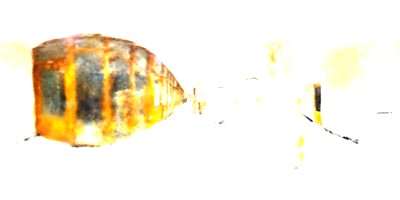} \\
        \rotatebox{90}{\shortstack{Ground \\ Truth}} &
       \includegraphics[width=0.23\linewidth]{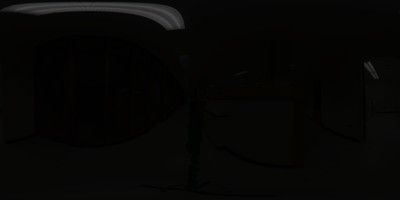} & 
       \includegraphics[width=0.23\linewidth]{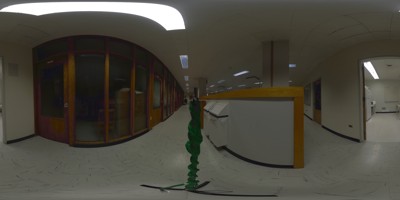} & 
       \includegraphics[width=0.23\linewidth]{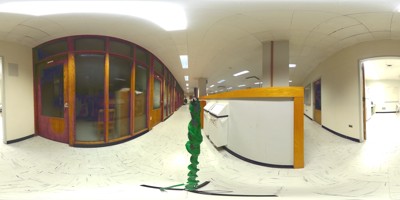} & 
       \includegraphics[width=0.23\linewidth]{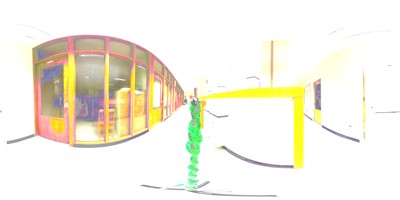} \\
       \rotatebox{90}{\shortstack{Radiance \\ Output}} &
       \includegraphics[width=0.23\linewidth]{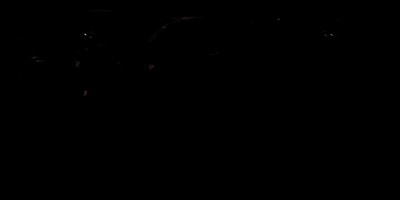} & 
       \includegraphics[width=0.23\linewidth]{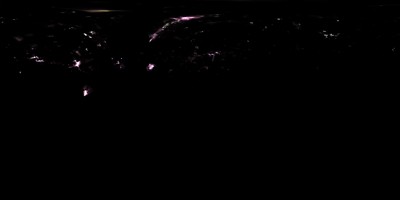} & 
       \includegraphics[width=0.23\linewidth]{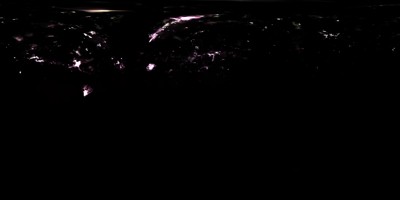} & 
       \includegraphics[width=0.23\linewidth]{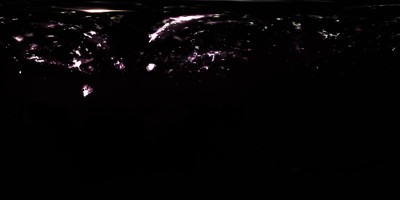} \\
       \rotatebox{90}{Combined} &
       \includegraphics[width=0.23\linewidth]{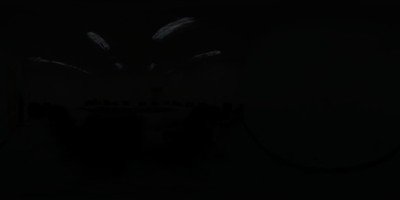} & 
       \includegraphics[width=0.23\linewidth]{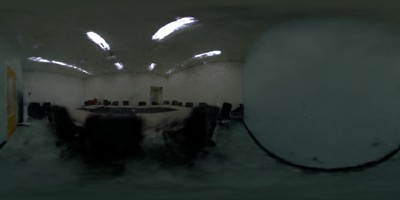} & 
       \includegraphics[width=0.23\linewidth]{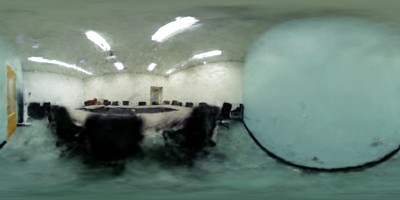} & 
       \includegraphics[width=0.23\linewidth]{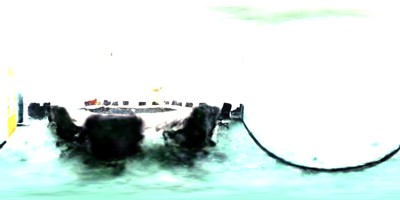} \\
        \rotatebox{90}{\shortstack{Ground \\ Truth}} &
       \includegraphics[width=0.23\linewidth]{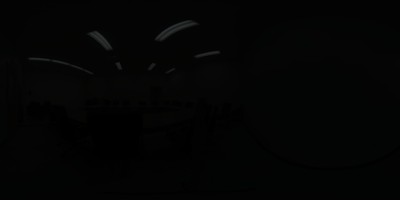} & 
       \includegraphics[width=0.23\linewidth]{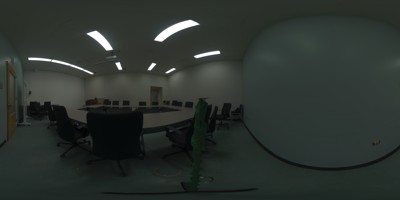} & 
       \includegraphics[width=0.23\linewidth]{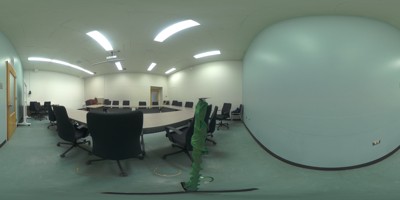} & 
       \includegraphics[width=0.23\linewidth]{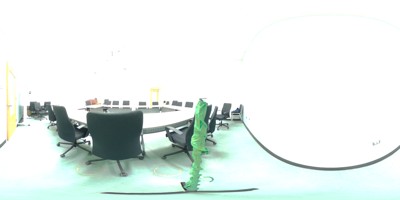} \\
    \end{tabular}
    \caption{Comparison of the ways to evaluate a trained HDR-Nerfacto model. The radiance output of HDR-Nerfacto needs to be scaled by an unknown color balance term (which was obtained from the ground truth here) and suffers from poor image quality, generating various artifacts. Instead, we combine the outputs of the tonemapper module at the well- and fast-exposed exposure to provide a fairer baseline.}
    \label{fig:hdr_nerfacto_output_compare}
\end{figure}

The radiance output of HDR-NeRF (and our implementation HDR-Nerfacto) is supposed to provide us with a linear HDR radiance output suitable for reconstructing high-quality images at any exposure. However, we find two significant problems with this output. First, the radiance output is known only up to scale; an unknown per-channel scaling factor must be obtained for each scene as a post-processing step. Even though HDR-NeRF is trained with the unit exposure loss, which regularizes the range of radiances, the scaling factors vary widely between scenes, as shown in \cref{tab:hdr_nerfacto_scale_factors}. Second, even after applying this scaling, the images at all the exposures exhibit extreme artifacts and are far from the ground truth, as shown in the ``Radiance output'' rows of \cref{fig:hdr_nerfacto_output_compare}. We attribute this issue to the fact that the radiance output is underconstrained when trained with only two non-overlapping exposures, which is always the case in our setup. Instead, we choose to combine the outputs of the tonemapper module at the well- and fast-exposed exposure, which drastically improves the reconstruction quality and provides a fairer baseline, as can be seen in the ``Combined'' rows of \cref{fig:hdr_nerfacto_output_compare}.


\begin{table}[t!]
    \centering
    \small	
    \caption{HDR-Nerfacto's radiance output has different scale factors per scene and color channel, which require additional computation to obtain. Here, we divide the per-channel mean value of the GT radiance with HDR-Nerfacto's to infer the scale factor.}
    \begin{tabular}{lccc}
    \toprule
        Scene & \multicolumn{3}{c}{Radiance scale factor} \\
         & Red & Green & Blue \\
         \midrule
         \scenedownstairlab & $3.22\!\times\!10^{-3}$ & $3.71\!\times\!10^{-3}$ & $9.18\!\times\!10^{-3}$ \\ 
         \scenemeetingroom & $3.14\!\times\!10^{-6}$ & $5.57\!\times\!10^{-7}$ & $1.73\!\times\!10^{-6}$ \\
         \bottomrule
    \end{tabular}
    \label{tab:hdr_nerfacto_scale_factors}
\end{table}

\include{tables/quantitative_extended}
\include{tables/quantitative_extended_2}

%% file: tables/quantitative_extended.tex
\begin{table*}
\small
\centering
\setlength{\tabcolsep}{2pt}
\caption{Quantitative results on the first seven scenes of our dataset of 14 real scenes. Metrics are shown in 4 groups (left to right): LDR panoramas, HDR panoramas, HDR and LDR renders (``LDR r.''). For ``renders'', we use the HDR panoramas to render a virtual scene (see fig.~7 from the paper), the metrics are computed on the result.}
\label{tab:quantitative-extended-results}
\begin{tabular}{llccccccccccccc}
\toprule
& & \multicolumn{3}{c}{LDR panos}
& & \multicolumn{3}{c}{HDR panos}
& & \multicolumn{3}{c}{HDR render}
& & LDR r. \\
Scene & Method
& PSNR$_\uparrow$
& SSIM$_\uparrow$
& LPIPS$_\downarrow$
& \,
& PU-PSNR$_\uparrow$ 
& HDR-VDP$_\uparrow$
& PU-SSIM$_\uparrow$
& \,
& si-RMSE$_\downarrow$
& RMSE$_\downarrow$
& RGB ang.$_\downarrow$ 
& \,
& PSNR$_\uparrow$ \\
\midrule
\multirow{4}{*}{\rotatebox{90}{\tiny \textsc{\scenelablobby}}}
& LDR-Nerfacto     
& 20.09 & 0.629 & 0.338
& & 24.57 & 6.520 & 0.885 
& & 0.457 & 0.605 & 3.76 
& & 27.37 \\ 
& PanoHDR-Nerfacto     
& 19.36 & 0.577 & 0.414
& & 27.44 & 6.793 & 0.878 
& & 0.390 & 0.494 & 9.95 
& & 28.11 \\
& HDR-Nerfacto     
& 19.41 & 0.537 & 0.481
& & 24.35 & 6.076 & 0.874 
& & 0.462 & 0.600 & 3.46 
& & 27.37 \\
& \thename (ours)      
& 19.32 & 0.576 & 0.393 
& & 25.59 & 6.177 & 0.877 
& & 0.467 & 0.551 & 4.26 
& & 28.04 \\
\midrule
\multirow{4}{*}{\rotatebox{90}{\tiny \textsc{\sceneclubhouse}}}
& LDR-Nerfacto     
& 17.53 & 0.588 & 0.421
& & 23.82 & 5.745 & 0.848 
& & 0.616 & 1.315 & 10.70
& & 27.97 \\ 
& PanoHDR-Nerfacto     
& 16.71 & 0.509 & 0.465
& & 25.22 & 6.373 & 0.827 
& & 0.580 & 1.025 & 10.92 
& & 28.12 \\
& HDR-Nerfacto     
& 17.03 & 0.536 & 0.521
& & 24.90 & 6.110 & 0.831 
& & 0.322 & 1.114 & 5.46 
& & 27.86 \\
& \thename (ours)      
& 17.26 & 0.559 & 0.442 
& & 24.44 & 5.731 & 0.844 
& & 0.420 & 0.951 & 7.28
& & 28.98 \\
\midrule
\multirow{4}{*}{\rotatebox{90}{\tiny \textsc{\scenerainbowlivingroom}}}
& LDR-Nerfacto     
& 16.93 & 0.596 & 0.424
& & 30.68 & 6.609 & 0.767 
& & 0.141 & 0.156 & 7.13
& & 30.12 \\ 
& PanoHDR-Nerfacto     
& 17.57 & 0.576 & 0.476
& & 28.57 & 6.515 & 0.796 
& & 0.143 & 0.366 & 8.72 
& & 29.09 \\
& HDR-Nerfacto     
& 17.17 & 0.571 & 0.506
& & 31.39 & 6.446 & 0.781 
& & 0.117 & 0.158 & 8.11 
& & 29.28 \\
& \thename (ours)      
& 16.59 & 0.580 & 0.431 
& & 30.48 & 6.438 & 0.762 
& & 0.107 & 0.219 & 7.57
& & 29.00 \\
\midrule
\multirow{4}{*}{\rotatebox{90}{\tiny \textsc{\scenebluebed}}}
& LDR-Nerfacto     
& 21.46 & 0.748 & 0.396
& & 28.44 & 6.762 & 0.922 
& & 0.476 & 0.525 & 3.05
& & 27.43 \\ 
& PanoHDR-Nerfacto     
& 19.32 & 0.663 & 0.461
& & 28.31 & 7.372 & 0.873 
& & 0.0418 & 0.134 & 5.46 
& & 27.52 \\
& HDR-Nerfacto     
& 20.09 & 0.685 & 0.494
& & 27.44 & 6.604 & 0.904 
& & 0.494 & 0.529 & 4.90
& & 28.00 \\
& \thename (ours)      
& 21.49 & 0.729 & 0.398 
& & 28.48 & 7.358 & 0.921
& & 0.475 & 0.52 & 3.66
& & 27.99 \\
\midrule
\multirow{4}{*}{\rotatebox{90}{\tiny \textsc{\scenemeetingroom}}}
& LDR-Nerfacto     
& 22.17 & 0.734 & 0.404
& & 25.93 & 6.462 & 0.940 
& & 0.183 & 0.356 & 1.48
& & 28.20 \\ 
& PanoHDR-Nerfacto     
& 21.76 & 0.708 & 0.433
& & 29.77 & 6.810 & 0.940 
& & 0.129 & 0.158 & 5.00 
& & 29.75 \\
& HDR-Nerfacto     
& 21.12 & 0.684 & 0.480
& & 28.11 & 6.541 &  0.932 
& & 0.146 & 0.274 & 1.61
& & 28.59 \\ 
& \thename (ours)      
& 21.96 & 0.728 & 0.402
& & 29.56 & 6.947 & 0.941 
& & 0.109 & 0.126 & 2.19
& & 32.17 \\ 
\midrule
\multirow{4}{*}{\rotatebox{90}{\tiny \textsc{\sceneauditoriumbright}}}
& LDR-Nerfacto     
& 18.41 & 0.619 & 0.384
& & 26.67 & 5.979 & 0.899 
& & 0.093 & 0.159 & 5.49
& & 27.59 \\ 
& PanoHDR-Nerfacto     
& 18.22 & 0.570 & 0.404
& & 25.77 & 6.019 & 0.890 
& & 0.076 & 0.108 & 7.05
& & 29.31 \\
& HDR-Nerfacto     
& 17.92 & 0.588 & 0.445
& & 26.47 & 5.984 & 0.884 
& & 0.069 & 0.167 & 2.85
& & 27.57 \\
& \thename (ours)      
& 18.34 & 0.614 & 0.372
& & 26.58 & 6.437 & 0.896 
& & 0.042 & 0.053 & 4.60
& & 30.19 \\
\midrule
\multirow{4}{*}{\rotatebox{90}{\tiny \textsc{Class no. Win.}}}
& LDR-Nerfacto     
& 20.03 & 0.739 & 0.362
& & 26.20 & 5.820 & 0.920 
& & 0.151 & 0.288 & 1.98
& & 27.59 \\ 
& PanoHDR-Nerfacto     
& 20.32 & 0.717  & 0.374
& & 27.77 & 6.619 & 0.931 
& & 0.057 & 0.186 & 1.19 
& & 28.23 \\
& HDR-Nerfacto     
& 19.65 & 0.699 & 0.445
& & 26.24 &  6.267 & 0.920 
& & 0.151 & 0.265 & 1.83
& & 27.69 \\ 
& \thename (ours)      
& 19.80 & 0.733 & 0.367
& & 29.99 & 6.698 & 0.923 
& & 0.084 & 0.123 & 1.68
& & 28.07 \\ 
\bottomrule
\end{tabular}

\end{table*}

%% file: tables/quantitative_extended_2.tex
\begin{table*}
\small
\centering
\setlength{\tabcolsep}{2pt}
\caption{Quantitative results on the second seven scenes of our dataset of 14 real scenes. Metrics are shown in 4 groups (left to right): LDR panoramas, HDR panoramas, HDR and LDR renders (``LDR r.''). For ``renders'', we use the HDR panoramas to render a virtual scene (see fig.~7 from the paper), the metrics are computed on the result.}
\label{tab:quantitative-extended-results_2}
\begin{tabular}{llccccccccccccc}
\toprule
& & \multicolumn{3}{c}{LDR panos}
& & \multicolumn{3}{c}{HDR panos}
& & \multicolumn{3}{c}{HDR render}
& & LDR r. \\
Scene & Method
& PSNR$_\uparrow$
& SSIM$_\uparrow$
& LPIPS$_\downarrow$
& \,
& PU-PSNR$_\uparrow$ 
& HDR-VDP$_\uparrow$
& PU-SSIM$_\uparrow$
& \,
& si-RMSE$_\downarrow$
& RMSE$_\downarrow$
& RGB ang.$_\downarrow$ 
& \,
& PSNR$_\uparrow$ \\
\midrule
\multirow{4}{*}{\rotatebox{90}{\tiny \textsc{Class w. Win.}}}
& LDR-Nerfacto     
& 19.695 & 0.670 & 0.375
& & 25.51 & 5.603 & 0.907 
& & 0.163 & 0.377 & 2.76
& & 27.84 \\ 
& PanoHDR-Nerfacto     
& 19.45 & 0.654 & 0.406
& & 28.88 & 6.798 & 0.912 
& & 0.113 & 0.159 & 2.85 
& & 30.27 \\
& HDR-Nerfacto     
& 19.11 & 0.622 & 0.482
& & 25.34 & 6.093 & 0.898 
& & 0.168 & 0.368 & 3.01
& & 27.81 \\ 
& \thename (ours)      
& 19.58 & 0.662 & 0.382
& & 28.84 &  6.723 & 0.912
& & 0.112 & 0.149 & 2.54
& & 29.31 \\ 
\midrule
\multirow{4}{*}{\rotatebox{90}{\tiny \textsc{\scenedownstairlab}}}
& LDR-Nerfacto     
& 19.49 & 0.637 & 0.411
& & 26.61 & 6.063 & 0.916 
& & 0.454 & 0.205 & 6.18
& & 27.89 \\ 
& PanoHDR-Nerfacto     
& 18.85 & 0.582 & 0.462
& & 26.24 & 6.326 & 0.904 
& & 0.159 & 0.547 & 7.05 
& & 28.68 \\
& HDR-Nerfacto     
& 19.12 & 0.608 & 0.442
& & 29.22 & 6.497 & 0.913 
& & 0.063 & 0.118 & 2.46
& & 29.92 \\ 
& \thename (ours)      
& 19.18 & 0.612 & 0.434
& &  29.22 & 6.532 & 0.912 
& & 0.061 & 0.121 & 2.42
& & 29.96 \\ 
\midrule
\multirow{4}{*}{\rotatebox{90}{\tiny \textsc{\sceneg}}}
& LDR-Nerfacto     
& 20.61 & 0.670 & 0.389
& & 26.15 & 6.191 & 0.928 
& & 0.189 & 0.346 & 8.14
& & 28.09 \\ 
& PanoHDR-Nerfacto     
& 19.23 & 0.623 & 0.480
& & 31.55 & 6.691 & 0.926 
& & 0.093 & 0.194 & 6.79 
& & 29.09 \\
& HDR-Nerfacto     
& 19.23 & 0.480 & 0.623
& & 28.78 & 6.283 & 0.905 
& & 0.118 & 0.268 & 6.44
& & 27.66 \\ 
& \thename (ours)      
& 20.74 & 0.679 & 0.396
& & 32.52 & 6.722 & 0.934 
& & 0.043 & 0.085 & 1.71
& & 28.91 \\ 
\midrule
\multirow{4}{*}{\rotatebox{90}{\tiny \textsc{\scenecoffeeroom}}}
& LDR-Nerfacto     
& 21.69 & 0.706 & 0.441
& & 28.15 & 6.350 & 0.938 
& & 0.234 & 0.333 & 4.66
& & 28.48 \\ 
& PanoHDR-Nerfacto     
& 21.21 & 0.649 & 0.478
& & 31.10 & 6.617 & 0.927 
& & 0.151 & 0.214 & 2.95
& & 30.39 \\
& HDR-Nerfacto     
& 19.97 & 0.645 & 0.523
& & 28.83 & 6.233 & 0.914 
& & 0.523 & 0.314 & 4.27
& & 28.34 \\ 
& \thename (ours)      
& 21.60 & 0.696 & 0.436
& & 32.43 & 6.916 & 0.939 
& & 0.126 & 0.188 & 3.66
& & 30.07 \\ 
\midrule
\multirow{4}{*}{\rotatebox{90}{\tiny \textsc{\scenejfoffice}}}
& LDR-Nerfacto     
& 21.60 & 0.684 & 0.397
& & 28.02 & 6.740 & 0.927 
& & 0.367 & 0.400 & 3.81 
& & 27.71 \\ 
& PanoHDR-Nerfacto     
& 20.88 & 0.643 & 0.432
& & 29.25 & 7.047 & 0.921 
& & 0.314 & 0.410 & 2.51 
& & 32.46 \\ 
& HDR-Nerfacto     
& 20.82 & 0.640 & 0.481
& & 28.89 & 6.867 & 0.916 
& & 0.316 & 0.364 & 4.52 
& & 28.50 \\ 
& \thename (ours)      
& 21.44 & 0.677 & 0.392
& & 31.38 & 7.319 & 0.926 
& & 0.204 & 0.236 & 4.51 
& & 31.06 \\ 
\midrule
\multirow{4}{*}{\rotatebox{90}{\tiny \textsc{\sceneauditoriumdark}}}
& LDR-Nerfacto     
& 19.77 & 0.621 & 0.361
& & 28.99 & 6.334 & 0.921 
& & 0.132 & 0.197 & 11.44 
& & 27.57 \\ 
& PanoHDR-Nerfacto     
& 19.47 & 0.545 & 0.396
& & 29.28 & 6.361 & 0.913 
& & 0.084 & 0.145 & 11.32 
& & 28.10 \\ 
& HDR-Nerfacto     
& 19.50 & 0.590 & 0.435
& & 30.06 & 6.524 & 0.911 
& & 0.050 & 0.095 & 4.12 
& & 29.46 \\ 
& \thename (ours)      
& 19.66 & 0.604 & 0.354
& & 27.79 & 6.198 & 0.916 
& & 0.043 & 0.168 & 2.76 
& & 28.67 \\ 
\midrule
\multirow{4}{*}{\rotatebox{90}{\tiny \textsc{\scenelvsn}}}
& LDR-Nerfacto     
& 20.87 & 0.645 & 0.431
& & 25.62 & 6.349 & 0.925 
& & 0.199 & 0.377 & 5.08 
& & 27.66 \\ 
& PanoHDR-Nerfacto     
& 20.86 & 0.626 & 0.454
& & 28.85 & 6.687 & 0.928 
& & 0.163 & 0.196 & 5.68 
& & 29.44 \\ 
& HDR-Nerfacto     
& 19.69 & 0.522 & 0.585
& & 26.18 & 6.267 & 0.914 
& & 0.189 & 0.341 & 3.88 
& & 27.58 \\ 
& \thename (ours)      
& 20.72 & 0.639 & 0.426
& & 29.89 & 6.792 & 0.927 
& & 0.117 & 0.129 & 1.44 
& & 29.46 \\ 
\bottomrule
\end{tabular}
\end{table*}